\documentclass[11pt]{article}
\usepackage[margin=2cm]{geometry}
\usepackage{authblk}
\usepackage{enumitem}
\setlist{nosep} 

\usepackage[round]{natbib}

\usepackage{amsgen,amsmath,amstext,amsbsy,amsopn,amssymb,amsthm,bm}
\usepackage{float}
\usepackage{xcolor}
\usepackage{mathtools}
\usepackage{multirow,bigstrut,booktabs,makecell,diagbox}
\usepackage{caption}
\usepackage{subcaption}
\usepackage[linesnumbered,ruled,vlined]{algorithm2e}
\usepackage[hidelinks]{hyperref}

\newcommand{\alert}[1]{{\leavevmode\color{black}{#1}}}

\newtheorem{thm}{Theorem}[section]
\newtheorem{lem}[thm]{Lemma}

\theoremstyle{definition}
\newtheorem{dfn}{Definition}[section]

\newtheorem{asp}{Assumption}

\theoremstyle{remark}

\newcommand{\be}{\bm{e}}

\newcommand{\bg}{\bm{g}}
\newcommand{\bh}{\bm{h}}

\newcommand{\bv}{\bm{v}}
\newcommand{\bw}{\bm{w}}
\newcommand{\bx}{\bm{x}}

\newcommand{\bD}{\bm{D}}

\newcommand{\bH}{\bm{H}}
\newcommand{\bI}{\bm{I}}

\newcommand{\bQ}{\bm{Q}}

\newcommand{\bU}{\bm{U}}

\newcommand{\bX}{\bm{X}}

\newcommand{\Ib}{\mathbf{I}}

\newcommand{\bbB}{\mathbb{B}}

\newcommand{\bbE}{\mathbb{E}}

\newcommand{\bbG}{\mathbb{G}}

\newcommand{\bbN}{\mathbb{N}}

\newcommand{\bbP}{\mathbb{P}}

\newcommand{\bbR}{\mathbb{R}}

\newcommand{\cA}{\mathcal{A}}

\newcommand{\cD}{\mathcal{D}}

\newcommand{\cG}{\mathcal{G}}
\newcommand{\cH}{\mathcal{H}}

\newcommand{\cO}{\mathcal{O}}
\newcommand{\cP}{\mathcal{P}}

\newcommand{\cS}{\mathcal{S}}

\newcommand{\cX}{\mathcal{X}}

\newcommand{\bbeta}{\bm{\beta}}
\newcommand{\beeta}{\bm{\eta}}

\newcommand{\bmu}{\bm{\mu}}

\newcommand{\btheta}{\bm{\theta}}
\newcommand{\bxi}{\bm{\xi}}

\newcommand{\bnu}{\bm{\nu}}

\newcommand{\bSigma}{\bm{\Sigma}}

\newcommand{\sgn}[1]{\operatorname{sign} \{#1\}}
\DeclareMathOperator*{\argmin}{arg\,min}
\DeclareMathOperator*{\argmax}{arg\,max}

\DeclareMathOperator{\Var}{Var}

\newcommand{\bzero}{{\mathbf{0}}} 

\newcommand{\norm}[1]{\left\lVert#1\right\rVert}
\newcommand{\normn}[1]{\lVert#1\rVert}
\newcommand{\abs}[1]{\left\lvert#1\right\rvert}
\newcommand{\absn}[1]{\lvert#1\rvert}
\newcommand{\prth}[1]{\left(#1\right)}
\newcommand{\brck}[1]{\left[#1\right]}
\newcommand{\brce}[1]{\left\{#1\right\}}
\newcommand{\agl}[1]{\left\langle#1\right\rangle}

\title{\bf Asymptotic Inference for Multi-Stage Stationary Treatment Policy with Variable Selection}
\author[1]{Daiqi Gao}
\author[2]{Yufeng Liu}
\author[3]{Donglin Zeng}
\affil[1]{Harvard University}
\affil[2]{University of North Carolina at Chapel Hill}
\affil[3]{University of Michigan}
\date{}

\begin{document}
\maketitle

\begin{abstract}
    Dynamic treatment regimes or policies are a sequence of decision functions over multiple stages that are tailored to individual features. One important class of treatment policies in practice, namely multi-stage stationary treatment policies, prescribes treatment assignment probabilities using the same decision function across stages, where the decision is based on the same set of features consisting of time-evolving variables (e.g., routinely collected disease biomarkers). Although there has been extensive literature on constructing valid inference for the value function associated with dynamic treatment policies, little work has focused on the policies themselves, especially in the presence of high-dimensional feature variables. We aim to fill the gap in this work. Specifically, we first estimate the multi-stage stationary treatment policy using an augmented inverse probability weighted estimator for the value function to increase asymptotic efficiency, and further apply a penalty to select important feature variables. We then construct one-step improvements of the policy parameter estimators for valid inference. Theoretically, we show that the improved estimators are asymptotically normal, even if nuisance parameters are estimated at a slow convergence rate and the dimension of the feature variables increases with the sample size. Our numerical studies demonstrate that the proposed method estimates a sparse policy with a near-optimal value function and conducts valid inference for the policy parameters.
\end{abstract}


\section{Introduction}
Dynamic treatment regimes (DTRs) or policies have recently attracted considerable attention in precision medicine.
They are a sequence of decision functions to prescribe treatments over stages based on an individual's features which can evolve over stages.  
When the individual's features consist of clinical variables that are routinely collected over stages, one important class of policies, which we name as multi-stage stationary treatment policies (MSTPs), are to prescribe from the same set of treatments using the same decision function over all stages.  
\alert{Moreover, the policies are stochastic, meaning that individuals receive recommended treatments with a probability close but not equal to one.  
An MSTP is particularly critical for delivering interventions in mobile health (mHealth), where users receive treatments across multiple treatment stages, and the policy's effectiveness over all stages must be optimized.  
For example, OhioT1DM \citep{marling2020ohiot1dm} is a micro-randomized trial (MRT) designed to support health research for individuals with Type 1 Diabetes (T1D).  
The information including the current blood glucose level, the meal time, the exercise level, etc. are available at each stage.  
An MSTP that considers these features can guide decisions on insulin delivery, helping to maintain blood glucose levels within the target range over the long term.}
We focus on MSTPs in this paper because they are particularly useful for treating chronic diseases in clinical practices.  
First, the policies use the same decision function and the same set of variables so they are convenient for both implementation and interpretation.  
Second, the policies are dynamic by incorporating individual's evolving features in decisions, which are often known to be important for disease prognostics and thus are routinely collected during clinical visits or lab tests.  
Third, using stochastic decisions or stochastic policies is more flexible than relying on a deterministic policy whose estimation is known to be sensitive to evidence bias and noise. 
Moreover, the stochastic policies are useful for exploration \citep{gao2022non} and after-study analysis when applied to future studies \citep{boruvka2018assessing}.  

Many approaches have been developed to estimate optimal DTRs using data from a multi-stage study.  
For example, Q-learning derives the optimal DTRs by maximizing the so-called Q-functions, which are the conditional means of a total optimal reward outcome given the features at each stage and are often estimated using regression models in a backward fashion \citep{murphy2005generalization,zhao2011reinforcement,moodie2012q,zhu2019proper}.  
On the other hand, A-learning poses assumptions only on the interaction effect between the treatments and actions, and thus avoids potential problems of misspecified treatment-free effect \citep{murphy2003optimal,shi2018high,jeng2018high}.  
Some other policy search methods find the DTR within a function class that directly maximizes the value function, which is estimated using inverse probability weighting (IPW) with a convex surrogate loss \citep{zhao2015new} or the Augmented Inverse Propensity Weighted Estimator (AIPWE) \citep{zhang2013robust, liu2018augmented}.  
\alert{The latter is often called the doubly robust estimator.  
However, these methods typically focus on a small number of treatment stages where the environment is generally non-stationary and a time-dependent policy is preferred.  
In the reinforcement learning (RL) literature, \citet{nie2021learning} consider problems with a finite number of treatment stages (also called the horizon).  
They propose a method to learn a non-stationary policy without Markov assumptions using a doubly robust estimator and provide bounds on the convergence rate of the advantage function.  
Value iteration algorithms \citep{antos2007fitted} and doubly robust methods \citep{luckett2020estimating,liao2022batch} have been proposed to find the optimal stationary policy in infinite horizon Markov decision processes (MDPs) using off-policy data.  
A comprehensive review of the DTR and RL can be in found in \citet{kosorok2019precision} and \citet{sutton2018reinforcement}.  
However, none of the above provide inferential results for the policy.  
}

In addition to DTR estimation, a number of studies have considered obtaining valid inference for the value function associated with the estimated DTR.  
For example, \citet{luedtke2016statistical,zhu2019proper} and \citet{shi2020breaking} studied value inference, allowing for situations where the treatment is neither beneficial nor harmful for a subpopulation.  
Similar inference has been investigated in the RL framework under MDP assumptions \citep{kallus2020double,luckett2020estimating,liao2021off,shi2021deeply,shi2022statistical,kallus2022efficiently}.  
There are few methods to conduct inference for the treatment policies themselves. 
This is especially important for studying MSTPs with many feature variables: clinicians typically favor simple and parsimonious decisions due to concerns about implementation and the cost of collecting disease biomarkers during routine visits or lab tests.  
\citet{jeng2018high,zhu2019proper} derived the asymptotic distribution of the parameters in Q-learning or A-learning by assuming the regression models are correctly specified, which may not be plausible in the presence of many features.  
More recently, \citet{liang2022estimation} studied inference for deterministic treatment decisions in a high-dimensional setting but restricted their analysis to a single stage.  

This work aims to fill in the above gap by obtaining valid inference for MSTPs with high-dimensional features, assuming data from an MRT.  
Our stationary treatment policy is defined as a probability function of a linear combination of all the feature variables, and a tuning parameter is used in this function to approximate a deterministic decision.  
For inference, we first estimate the value function using the AIPWE \citep{bang2005doubly,liu2018augmented}, with a discussion on different constructions of the augmentation term.  
\alert{\citet{jiang2016doubly,thomas2016data} first proposed this estimator for off-policy evaluation (OPE) of a fixed target policy.  
\citet{kallus2020double} proved the semiparametric efficiency of this estimator in non-Markovian decision processes when the ratio between the behavior policy and the target policy is used, and in MDPs when a marginalized density ratio of the current state and action is used.}
Furthermore, to find a sparse estimator of MSTP, we impose an $L_1$ penalty for the purpose of variable selection.  
The classical inferential theory fails in this case due to the presence of high-dimensional parameters.  
The asymptotic distribution of the estimated parameter becomes intractable due to the non-ignorable estimation bias and the sparsity effect of the nuisance parameters.  
To validate the inference for the parameter estimators, we adopt the idea of one-step estimation \citep{zhang2014confidence,ning2017general} to remove bias in the regularized estimators.  

\alert{Our main contributions lie in the following aspects.  
(1) We learn a sparse high-dimensional policy and conduct its statistical inference. 
To the best of our knowledge, this is the first work that considers inference for multi-stage treatment rules themselves within policy search methods, especially in high-dimensional settings.  
(2) Theoretically, we show that the final estimators for the policy parameters are asymptotically normal even if the dimension of the feature variables increases exponentially with $\sqrt{n}$, or the models for $Q$-functions are misspecified and their parameters are estimated at a rate arbitrarily slow.  
The theoretical analysis is more involved than general high-dimensional inference problems as in \citet{ning2017general}, since we need to account for the high-dimensional plug-in estimator of the nuisance parameters and other high-dimensional components.  
The augmentation in our first step shares a similar spirit with debiased machine learning methods in \citet{chernozhukov2018double}, where nuisance parameters (Q-functions) are decorrelated from the policy parameters, except that we do not rely on data-splitting to ensure more reliable estimation with limited data.  
The one-step improvement in our second step further decorrelates the parameter of interest from the remaining high-dimensional parameters in MSTP.  
In other words, our proposed method incorporates two decorrelation procedures for inference.  
(3) The optimization and inference of multi-stage policies with off-policy data are computationally challenging.  
Importance sampling weights are utilized to adjust for the difference between the behavior policy used to collect the data and the target policy to be optimized.
The variance of the weights increases with the number of treatment stages.  
We leverage several techniques to stabilize the estimated asymptotic variance, including using the difference quotient \citep{lax2014calculus} to estimate the gradient and Hessian matrix of the loss function.  
}

The rest of this paper is organized as follows.  
In Section~\ref{sec:methodology.MSTP}, we define the objective of our problem and describe our method for learning the MSTP.  
Then we introduce the procedure for constructing the one-step estimator in theory and specify some implementation details.  
In Section~\ref{sec:theory.MSTP}, we show that the one-step estimator is asymptotically normal and provide its confidence intervals (CI).  
In Sections~\ref{sec:simulation.MSTP} and~\ref{sec:real.data}, we demonstrate our method in simulation studies and a real data example.  
Finally, we conclude this paper with some discussion in Section~\ref{sec:discussion.MSTP}.

\section{Methodology} \label{sec:methodology.MSTP}

\alert{We consider selecting treatment policies for $T$ stages, where the feature space $\cX$ and the treatment space $\cA = \{-1,1\}$ are the same across stages.  
At each stage, $\bX = [X_{1:d}] \in \cX$ is a $d$-dimensional feature vector, $A \in \cA$ is the treatment, and $R \in \bbR$ is the reward.  
Here, we use the notation $i:j$ to represent the index set $i, \dots, j$, where the indices $i, j \in \bbN$.  
The distributions of the feature $\bX$ and the treatment $A$ may vary across stages.  
An MSTP assigns treatments with the same probability in all stages, and is defined as a mapping $\pi: \cX \mapsto \cP(\cA)$ from the feature space to the space of probability distributions over the action space.  
Specifically, $\pi(a|\bX)$ denotes the probability of assigning treatment $a \in \cA$ when the feature is $\bX$.  
}

We assume that data are obtained from an offline, $T$-stage study, so the observed trajectory of the $i$-th subject is denoted as  
\[ \bD_i = \{ \bX_{i,1}, A_{i,1}, R_{i,1}, \bX_{i,2}, A_{i,2}, R_{i,2}, \dots, \bX_{i,T}, A_{i,T}, R_{i,T} \}, \]  
where the reward $R_{i,t} \in \bbR$ is an unknown function of the data $\{ \bX_{i,1}, A_{i,1}, R_{i,1}, \dots, \bX_{i,t}, A_{i,t} \}$ observed prior to or at time $t$.  
Let the domain of $\bD_i$ be defined as $\cD := (\cX \times \cA \times \bbR)^T$.  
Assume there are $n$ subjects and their trajectories $\{ \bD_i \}_{i=1}^n$ are independent and identically distributed.  
We assume that each action $A_{i,t}$ is taken randomly with probability depending on the history $\bH_{i,t} = \{ \bX_{i,1:(t-1)}, A_{i,1:(t-1)}, R_{i,1:(t-1)}, \bX_{i,t} \}$, and denote $\mu_t (a | \bH_{i,t})$ as the conditional probability of $A_{i,t}=a$ given $\bH_{i,t}$.  
The collection $\bmu := [\mu_{1:T}]$ is sometimes called the behavior policy in the RL literature.  

An important metric to evaluate an MSTP with parameter $\btheta$ is called the value function, which is defined as the sum of rewards $V(\btheta) = \bbE^{\btheta} (\sum_{t=1}^T R_t)$.  
\alert{The optimal MSTP in a policy class $\Pi$, whose parameter is denoted as $\btheta^*$, is the one maximizing $V(\btheta)$, i.e., the optimal parameter $\btheta^* \in \argmax_{\btheta: \pi^{\btheta} \in \Pi} V(\btheta)$.  
We focus on the policy class  
\[
\Pi = \brce{ \pi^{\btheta} : \pi^{\btheta} (a | \bX) = \frac{e^{a g(\bX, \btheta) / \tau}}{1 + e^{a g(\bX, \btheta) / \tau}}, \norm{\btheta}_2 \le 1, a \in \cA, \bX \in \cX} ,
\]  
where $\btheta = [\theta_{0:d}]$ and $\tau$ is a constant scaling parameter.  
Here, an MSTP $\pi^{\btheta} (a|\bX) \in \Pi$ is indexed by $\btheta$.  
The function $g$ is taken to be linear, defined as $g(\bX, \btheta) = \theta_0 + \sum_{j=1}^d X_j \theta_j$.  
It provides direct interpretability regarding the influence of each feature and ensures estimation efficiency with limited sample sizes, both of which are essential in clinical practices.  
Note that the scaling parameter $\tau$ adjusts the strength of influence of parameters on the action probability.  
A smaller $\tau$ leads to a decision closer to a deterministic policy.  
The constraint $\norm{\btheta}_2= 1$ ensures boundedness of the optimal policy within the class $\Pi$.  
Without such constraints, the optimal policy $\pi^{\btheta^*}$ will have $\btheta^* \to \infty$ to approximate the deterministic policy.
Without confusion, we use $\bbE^{\btheta}$ to denote the expectation under the policy indexed by $\btheta$, and use $\bbE$ to denote the expectation under the behavior policy.
Our goal is to estimate the optimal MSTP and obtain a proper inference for the policy parameter $\btheta$.  
}

To infer the optimal MSTP using the observed data, we make the following three standard assumptions \citep{murphy2003optimal}.  
Denote $\bX_{t} (a_{1:(t-1)})$ as the potential state at time $t$ if an action sequence $a_{1:(t-1)} \in \cA^{t-1}$ were taken.

\begin{asp}[Sequential ignorability] \label{asp:nuca}
    The sequence of potential outcomes (features) $\{ \bX_{t'} (A_{1:(t-1)}, a_{t:(t'-1)}) \}_{t' = t+1}^{T+1}$ is independent of the treatment $A_t$ given the observed information $\brce{\bX_{1}, A_{1}, \dots, \bX_{t-1}, A_{t-1}}$  
    for all $a_t \in \cA$, $t \in \{1:T\}$.
\end{asp}

\begin{asp}[Consistency] \label{asp:consistency.MSTP}
    The observed outcomes are consistent with the potential outcomes, $\bX_{t} = \bX_{t} (A_{1:(t-1)})$ for all $t \in \{2:(T+1)\}$.
\end{asp}

\begin{asp}[Positivity] \label{asp:positivity.MSTP}
    There exists a constant $p_0 > 0$ such that $\mu_t(a | \bH_{t}) \geq p_0$ for all $a \in \cA$ and $t \in \{1:T\}$.
\end{asp}

Note that the first assumption holds if the data are from an MRT, which we assume for the subsequent development.

\subsection{Estimate Policy Parameter with Variable Selection} \label{sec:estimate.policy.parameter}

\alert{To estimate the value function $V(\btheta)$ using the data with known $\mu_t$, we first denote $Q_t (\bx_t, a_t)$ as some function of rewards given $\bX_t = \bx_t$ and $A_t = a_t$.  
For example, $Q_t^{\btheta} (\bx_t, a_t) := \bbE^{\btheta} \{ \sum_{k=t}^T R_k \mid \bX_t = \bx_t, A_t = a_t \}$ represents the Q-function at stage $t$ under the policy $\pi^{\btheta}$.  
In general, the superscript $\btheta$ is omitted in $Q_t$, as the specific definition may or may not depend on $\pi^{\btheta}$, which will be further discussed in Section~\ref{sec:implementation}.  
Even when the definition of $Q_t$ explicitly depends on $\btheta$, as in the Q-function of the policy $\pi^{\btheta}$, it is often estimated first and plugged in subsequently during the estimation of $\btheta$.  
}
Let the expectation of $Q_t$ given $\bX_t$ be  
$$U_t (\bx_t) :=  \bbE \brce{ Q_t (\bX_t, A_t) | \bX_t = \bx_t},$$  
where $A_t \sim \pi^{\btheta} (\cdot | \bX_t)$.  
Denote $\bQ := [Q_{1:T}]$ and $\bU := [U_{1:T}]$.  
Then, based on the AIPWE given in \citet{kallus2020double},  
we construct the following estimator for $V(\btheta)$:  
\begin{equation*}
    \hat{V} (\btheta) 
    = \frac{1}{n} \sum_{i=1}^n \sum_{t=1}^T \brce{ 
    \rho_{i, 1:t}^{\btheta, \bmu} (\bH_{i,t}, A_{i,t}) [R_{i,t} - Q_t (\bX_{i,t}, A_{i,t})]
    + \rho_{i, 1:(t - 1)}^{\btheta, \bmu} (\bH_{i,t-1}, A_{i,t-1}) U_t (\bX_{i,t})},
\end{equation*}
where $\rho_{i, 1:t}^{\btheta, \bmu} (\bH_{i,t}, A_{i,t}) := \prod_{k=1}^t \frac{\pi^{\btheta} (A_{i,k} | \bX_{i,k})}{\mu_k (A_{i,k} | \bH_{i,k})}$ is often called the importance sampling weight.
\alert{For each stage $t$, to adjust for the difference between the behavior policy $\bmu$ and the target policy $\pi^{\btheta}$, $\hat{V} (\btheta)$ uses the step-wise weight for each reward within a trajectory.  
Compared to the trajectory-wise weight, which is the weight for the entire trajectory, the step-wise weight marginalizes over the weights for future actions and thus reduces the variance of the value estimator \citep{jiang2016doubly}. }
One major advantage of using the AIPWE for estimating $V(\btheta)$ is that the estimator $\hat{V}(\btheta)$ is unbiased for arbitrary $\bQ$, as long as $U_t$ is correctly evaluated as the conditional expectation of $Q_t$.
To see this, notice that the expectation of $\hat{V} (\btheta)$ is
\begin{equation*}
\begin{split}
    \bbE \hat {V} (\btheta)
    = \bbE^{\btheta} \brce{ \sum_{t=1}^T R_{i,t} }
    - & \bbE \Bigg\{ \sum_{t=1}^T \rho_{i, 1:(t - 1)}^{\btheta, \bmu} (\bH_{i,t-1}, A_{i,t-1}) \cdot \\
    & \bbE \brck{ \frac{\pi^{\btheta} (A_{i,t} | \bX_{i,t})}{\mu_t (A_{i,t} | \bH_{i,t})} {Q}_t (\bX_{i,t}, A_{i,t}) - {U}_t (\bX_{i,t}) \middle| \bH_{i,t} } \Bigg\}
\end{split}
\end{equation*}
given any $Q_t$ and $U_t$.
\alert{The first term on the right-hand side follows from inverse probability weighting, and the second term follows from the tower property of conditional expectation.
}
Now by the definition of $U_t$,  
\begin{align*}
    & \bbE \left\{ \frac{\pi^{\btheta} (A_{i,t} | \bX_{i,t})}{\mu_t (A_{i,t} | \bH_{i,t})} Q_t (\bX_{i,t}, A_{i,t}) - U_t (\bX_{i,t}) \middle| \bH_{i,t} \right\} \\
    & \qquad \qquad \qquad  
    = \sum_{a \in \cA} \mu_t (a | \bH_{i,t}) \frac{\pi^{\btheta} (a | \bX_{i,t})}{\mu_t (a | \bH_{i,t})} {Q}_t (\bX_{i,t}, a) - {U}_t (\bX_{i,t}) = 0
\end{align*}
once ${U}_t (\bX_{i,t}) = \bbE [Q_t (\bX_{i,t}, A_{i,t}) | \bX_{i, t}]$.  
Therefore, $\bbE \hat{V}(\btheta) = V (\btheta)$ for any $Q_t$.  
\alert{In OPE for any fixed policy $\pi$ in non-Markov decision processes, the value function estimator $\hat{V} (\btheta)$ is semiparametrically efficient when $Q_t$ is defined as the Q-function $Q_t^{\btheta}$ of $\pi^{\btheta}$ at stage $t$, and is estimated using sample splitting with a rate condition or without sample splitting with a Donsker condition.  
Besides, $\hat{V} (\btheta)$ is a doubly robust estimator, which guarantees $\sqrt{n}$-consistency even if one of the nuisance parameters $\bmu$ and $\bQ$ is misspecified in an observational study \citep{kallus2020double}.  
}

The above property implies that we can always obtain a consistent estimator for $\btheta^*$ with any working model for $Q_t$.  
\alert{However, a good choice of $Q_t$ may improve the asymptotic efficiency of the estimated policy parameter $\hat{\btheta}$.  
We will discuss the different methods for estimating $\hat{Q}_t$ in Section~\ref{sec:implementation}.  
}
To ensure the relationship between $Q_t$ and $U_t$ to be correct, we estimate $U_t$ using $\hat{U}_t (\bx) = \sum_{a \in \cA} \pi^{\btheta} (a | \bx) \hat{Q}_t (\bx, a)$.
After plugging $\hat{Q}_t$ and $\hat{U}_t$ into the expression of $\hat{V}(\btheta)$, we aim to maximize $\hat{V}(\btheta)$, or equivalently, minimize the function  
\begin{equation} \label{equ:lossfunc}
    \ell (\btheta, \hat{\beeta}) := \frac{1}{n} \sum_{i=1}^n l_i (\btheta, \hat{\beeta}),
\end{equation}
where $\beeta := \bQ$ is the nuisance parameter, and  
\begin{equation} \label{equ:lossfunc.i}
\begin{split}
    l_i (\btheta, \hat{\beeta}) := - \sum_{t=1}^T \brce{ 
    \rho_{i, 1:t}^{\btheta, \bmu} (\bH_{i,t}, A_{i,t}) [R_{i,t} - \hat{Q}_t (\bX_{i,t}, A_{i,t})]
    + \rho_{i, 1:(t - 1)}^{\btheta, \bmu} (\bH_{i,t-1}, A_{i,t-1}) \hat{U}_t (\bX_{i,t})
    }.
\end{split}
\end{equation}
Now we have $\bbE [\ell (\btheta, \beeta)] = - V (\btheta)$ for any $\beeta$.  
Assume $\beeta \in \cH$, where $\cH$ is a convex subset of some normed vector space. 
Suppose the estimator $\hat{\beeta}$ converges in probability uniformly to some deterministic limit $\bar{\beeta}$.  
Notably, only the limit $\bar{\beeta}$, and not the true value $\beeta^*$, will appear in the asymptotic distribution of $\hat{\btheta}$.  
Since $\btheta$ is high-dimensional, we include an $L_1$-penalty to obtain a sparse MSTP. Specifically, we estimate ${\btheta}$ as  
\begin{equation} \label{equ:objfunc}
    \hat{\btheta} = \argmin_{\btheta} [\ell (\btheta, \hat{\beeta}) + \lambda_{\btheta} \norm{\btheta}_1 ] \qquad \text{s.t. } \norm{\btheta}_2 \le 1 ,
\end{equation}
where $\lambda_{\btheta}$ is a tuning parameter for $\btheta$.  
\alert{The $L_1$ penalty is primarily chosen due to its convexity and the availability  of efficient optimization methods such as proximal gradient descent \citep{sra2011optimization}.  
Other nonconvex penalties like MCP \citep{zhang2010nearly} and SCAD \citep{fan2001variable} can also be incorporated into the proposed method \citep{ning2017general}.  
}

\subsection{Statistical Inference for Sparse High Dimensional Parameters}

To obtain valid inference, we follow \citet{ning2017general} to construct a one-step improvement of the estimator obtained in the previous section.  
\alert{Note that the classical Rao's score test is based on the asymptotically normal distribution of the profile score function under the null hypothesis.  
However, in high-dimensional settings, the asymptotic distribution becomes intractable due to the estimation bias in regularized estimators.  
}

\alert{When the parameter of interest is $\theta_j$, we denote $\bnu_j := (\theta_{1:(j-1)}, \theta_{(j+1):d})$ and let $\bxi := (\theta_j, \bnu_j)$.  
Now both $\beeta$ and $\bnu_j$ can be viewed as nuisance parameters.  
}  
Let the corresponding optimal parameters of $\theta_j$ and $\bnu_j$ be $\theta_j^*$ and $\bnu_j^*$.  
Given the estimated nuisance parameter $\hat{\beeta}$, we can define the matrix 
$\Ib := \bbE [\nabla_{\bxi \bxi} \ell (\btheta, \hat{\beeta})]$.
With $I_{\theta_j \theta_j}, \Ib_{\theta_j \bnu_j}, \Ib_{\bnu_j \bnu_j}, \Ib_{\bnu_j \theta_j}$ being the corresponding submatrices of $\Ib$ with respect to the partition of the parameters, let $I_{\theta_j | \bnu_j} = I_{\theta_j \theta_j} - \Ib_{\theta_j \bnu_j} \Ib_{\bnu_j \bnu_j}^{-1} \Ib_{\bnu_j \theta_j}$.  
Next, a decorrelated score function is defined as  
\[
S_j(\theta_j, \bnu_j, \beeta) = \nabla_{\theta_j} \ell (\theta_j, \bnu_j, \beeta) - \bw_j^T \nabla_{\bnu_j} \ell (\theta_j, \bnu_j, \beeta),
\]  
where $\bw_j^T = \Ib_{\theta_j \bnu_j} \Ib_{\bnu_j \bnu_j}^{-1}$.  
\alert{
The definition comes from the Taylor expansion of the score function, and ensures that the $S_j$ is uncorrelated with the nuisance score function.
The decorrelated score function is guaranteed to have asymptotic normality under the null hypothesis that $H_0: \theta_j^* = 0$, and can be used for hypothesis testing.}
Given the estimated parameters $\hat{\btheta}$ and $\hat{\beeta}$, the sparse estimator $\hat{\bw}_j$ of $\bw_j$ can be obtained using the Dantzig estimator  
\begin{equation} \label{equ:w}
    \hat{\bw}_j = \argmin \norm{\bw_j}_1, \qquad \text{s.t. } \norm{\nabla^2_{\theta_j \bnu_j} \ell (\hat{\btheta}, \hat{\beeta}) - \bw_j^T \nabla^2_{\bnu_j \bnu_j} \ell (\hat{\btheta}, \hat{\beeta})}_{\infty} \leq \lambda_{\bw_j},
\end{equation}
where $\lambda_{\bw_j}$ is tuned by cross-validation for each $j$ separately.  
This Dantzig estimator (\ref{equ:w}) is essentially the best sparse linear combination of the nuisance score function $\nabla_{\bnu_j} \ell (\theta_j, \bnu_j, \beeta)$ that approximates the score function $\nabla_{\theta_j} \ell (\theta_j, \bnu_j, \beeta)$ of the parameter of interest with error $\lambda_{\bw_j}$.  
It can also be estimated by penalized least squares using the gradient or Hessian matrix of $\ell$ \citep[see][Remark 3]{ning2017general}.  
Let $\hat{\bv}_j := (1, - \hat{\bw}_j^{T})^T$.  
Then $S_j(\theta_j, \bnu_j, \hat{\beeta})$ can be estimated as  
\begin{equation} \label{equ:score}
    \hat{S}_j (\hat{\theta}_j, \hat{\bnu}_{j}, \hat{\beeta}) = \nabla_{\theta_j} \ell (\hat{\theta}_j, \hat{\bnu}_{j}, \hat{\beeta}) - \hat{\bw}_j^T \nabla_{\bnu_j} \ell (\hat{\theta}_j, \hat{\bnu}_{j}, \hat{\beeta})
\end{equation}
by plugging in $\hat{\btheta}, \hat{\beeta}$ and $\hat{\bw}_j$.

Finally, building on the sparse estimator $\hat{\btheta}$ introduced in the previous section, a one-step estimator is defined as  
\begin{equation} \label{equ:ose}
    \tilde{\theta}_j := \hat{\theta}_j - \hat{S}_j (\hat{\theta}_j, \hat{\bnu}_{j}, \hat{\beeta}) / \hat{I}_{\theta_j | \bnu_j},  
    \qquad \text{where } \hat{I}_{\theta_j | \bnu_j} = \nabla_{\theta_j \theta_j}^2 \ell (\hat{\btheta}, \hat{\beeta}) - \hat{\bw}_j^T \nabla_{\bnu_j \theta_j}^2 \ell (\hat{\btheta}, \hat{\beeta})
\end{equation}
for $j \in \{1:d\}$.  
\alert{This estimator solves the equation $\hat{S}_j (\theta_j, \hat{\bnu}_{j}, \hat{\beeta}) = 0$ through a computationally efficient one-step method.  
Directly solving for the root of the decorrelated score function can be computationally demanding, especially when the score function is non-convex or has multiple roots.  
The one-step method provides an efficient alternative by solving the estimating equation  
$ \hat{S}_j (\hat{\theta}_j, \hat{\bnu}_{j}, \hat{\beeta}) + \hat{I}_{\theta_j | \bnu_j} (\theta_j - \hat{\theta}_j) = 0$.  
The resulting estimator is asymptotically equivalent to fully iterative solutions of the decorrelated score function.  
}

For the true parameter $\btheta^*$, define the matrix  
$\Ib^* := \bbE [\nabla_{\bxi \bxi} l_0 (\btheta^*, \bar{\beeta})]$ and the vectors $\bw_j^* = \Ib_{\bnu_j \bnu_j}^{*-1} \Ib_{\bnu_j \theta_j}^*$, $\bv_j^* = (1, - \bw_j^{*T})^T$
where $l_0$ is an independent copy of $l_i$ for any $i$.  
Now let  
\begin{equation} \label{equ:def.sigma.S}
    \bSigma^* := \Var [\nabla_{\bxi} l_0 (\btheta^*, \bar{\beeta})] \quad \text{and} \quad \sigma_j^* := \bv_j^{*T} \bSigma^* \bv_j^{*}.
\end{equation}
\alert{In general, $\bSigma^*$ could be different from $\Ib^*$ since our loss function is not the negative log-likelihood.}
As will be shown in Section~\ref{sec:theory.MSTP}, $\tilde{\theta}_j$ is asymptotically normal with mean $\theta_j^*$ and asymptotic variance $\sigma_j^* / [n I^{*2}_{\theta_j | \bnu_j}]$, which can be used for constructing the CIs.  
\alert{Note that after the decorrelation, the one-step estimator $\tilde{\theta}_j$ is still a consistent estimator of $\theta_j^*$.  
}

We summarize the theoretical steps of estimating the high-dimensional parameters and making inference with one-step estimators in Algorithm~\ref{alg:OSE}.  
Note that the confidence level applies to each individual CI rather than to multiple CIs simultaneously.  
Implementation methods will be specified in Section~\ref{sec:implementation}.

\begin{algorithm}[!t]
    \SetKwInOut{Input}{Input}
    \SetKwInOut{Output}{Output}
    \SetAlgoLined
    \Input{$n$ trajectories $\bD_{1:n}$.}
    \Output{The sparse estimator $\hat{\btheta}$, the one-step estimator $\tilde{\theta}_j$ with its CI for $j \in \{1:d\}$.}
    Estimate $\hat{\bQ}$ using an arbitrary working model\;
    Estimate the policy parameter $\hat{\btheta}$ by solving (\ref{equ:objfunc}) with $\hat{\bQ}$, where $\lambda_{\btheta}$ is tuned via cross-validation\;
    \For{$j \in \{1:d\}$}{
        Partition the sparse estimate $\hat{\btheta}$ as $(\hat{\theta}_0, \hat{\theta}_j, \hat{\bnu}_{j})$\;
        Obtain the Dantzig-type estimator $\hat{\bw}_j$ by solving (\ref{equ:w})\;
        Compute the decorrelated score function $\hat{S}_j (\theta_j, \hat{\bnu}_{j}, \hat{\beeta})$ using (\ref{equ:score})\;
        Calculate the one-step estimator $\tilde{\theta}_j$ using (\ref{equ:ose})\;
        Construct the $(1 - \alpha) \times 100\%$ CI for $\theta_j$ as
        \begin{equation} \label{equ:asy.CI.thetaj}
            \Bigg( \tilde{\theta}_j - \Phi^{-1}(1 - \alpha/2) \frac{\sqrt{\hat{\sigma}_j}}{\sqrt{n} \hat{I}_{\theta_j | \bnu_j}},
            \tilde{\theta}_j + \Phi^{-1}(1 - \alpha/2) \frac{\sqrt{\hat{\sigma}_j}}{\sqrt{n} \hat{I}_{\theta_j | \bnu_j}} \Bigg),
        \end{equation}
        where $\Phi$ is the cumulative distribution function of the standard normal distribution, and
        \[
        \hat{\sigma}_j = (1, -\hat{\bw}_j^T) \brck{\frac{1}{n} \sum_{i=1}^n \nabla_{\bxi} l_i (\hat{\btheta}, \hat{\beeta}) \nabla_{\bxi} l_i (\hat{\btheta}, \hat{\beeta})^T} (1, -\hat{\bw}_j^T)^T.
        \]
    }
    \caption{Optimization and inference of high-dimensional MSTP parameters}
    \label{alg:OSE}
\end{algorithm}

\subsection{Implementation} \label{sec:implementation}

\alert{The implementation of the theoretical method described in Algorithm~\ref{alg:OSE} is particularly challenging due to several reasons.  
First, the importance sampling weight $\rho_{i, 1:t}^{\btheta, \bmu}$ in (\ref{equ:lossfunc}) is unstable in numerical computation since it is a product of $t$ probability ratios.  
The variance of the value estimator can thus grow exponentially with $T$ \citep{liu2018breaking}.  
This problem is exacerbated in our case because the true optimal parameter $\btheta^*$ may be away from the behavior policy, further increasing the variance of the loss function.  
Second, an ideal nuisance parameter $\bQ$ might depend on the parameter $\btheta$ to be optimized.  
For example, as mentioned in Section~\ref{sec:estimate.policy.parameter}, estimating $\bQ$ as the Q-functions of $\pi^{\btheta}$ ensures that the value function estimator $\hat{V} (\btheta)$ is semiparametrically efficient \citep{kallus2020double}.  
However, solving (\ref{equ:objfunc}) becomes computationally expensive when $\btheta$ appears both in the policy $\pi^{\btheta}$ and in the Q-function $Q_t^{\btheta}$, and this can further exacerbate numerical instability.  
Third, the function (\ref{equ:objfunc}) itself is difficult to maximize due to the non-convexity of the loss function $\ell (\btheta, \hat{\beeta})$, the non-differentiability of the $L_1$ penalty, and the $L_2$ constraint.  
Examples of non-convex loss functions are provided in Section~\ref{sec:simulation.MSTP} and Appendix~\ref{sec:true.value.function}.  
In this section, we will address these challenges.  
}

\paragraph{Stabilizing the importance sampling weight.}
\alert{The importance sampling weight $\rho_{i, 1:t}^{\btheta, \bmu}$ appears in the loss function $\ell (\btheta, \hat{\beeta})$ in (\ref{equ:objfunc}) when finding the sparse estimator $\hat{\btheta}$, and also appears in the gradient and Hessian matrix of $\ell (\btheta, \hat{\beeta})$ in (\ref{equ:w}), (\ref{equ:score}), (\ref{equ:ose}), and (\ref{equ:asy.CI.thetaj}) when constructing the CIs for $\tilde{\theta}_j$ (see Appendix~\ref{sec:proof.gradient} for the gradient and Hessian matrix of $\ell (\btheta, \hat{\beeta})$).  
}
A practical solution for stabilizing the weight is to use the weighted importance sampling, which scales the probability ratio by the average ratio across all episodes at this stage.  
Specifically, the sparse estimator $\hat{\btheta}$ can be found by  
\begin{equation} \label{equ:wobjfunc}
\begin{split}
    \hat{\btheta} 
    = \argmin_{\btheta} \Bigg\{ - 
    \frac{1}{n} \sum_{i=1}^n \sum_{t=1}^T \frac{\rho_{i, 1:t}^{\btheta, \bmu}}{\omega^{\btheta, \bmu}_{1:t}} [R_{i,t} - \hat{Q}_t (\bX_{i,t}, & A_{i,t})]
    + \frac{\rho_{i, 1:(t-1)}^{\btheta, \bmu}}{\omega^{\btheta, \bmu}_{1:{t-1}}} \hat{U}_t (\bX_{i,t}) \\
    & + \lambda_{\btheta} \norm{\btheta}_1 \Bigg\}
    \qquad \text{s.t. } \norm{\btheta}_2 \le 1 ,
\end{split}
\end{equation}
where $\omega^{\btheta, \bmu}_{1:t} = \frac{1}{n} \sum_{j=1}^n \rho_{j, 1:t}^{\btheta, \bmu}$ and $\lambda_{\check{\btheta}}, \lambda_{\btheta}$ are tuning parameters for the initial and final sparse estimators, respectively.  
Similar weighted ratios have been used and discussed in \citet{precup2000temporal,thomas2015safe,thomas2016data}.  

In addition, our simulation shows that the gradient and the Hessian matrix can be more unstable than the loss function due to the product of more terms of probabilities.  
Therefore, we propose two amendments to solve this problem in practice.  
First, for the gradient and the Hessian matrix in (\ref{equ:w}), (\ref{equ:score}), and (\ref{equ:ose}), we calculate them numerically using the symmetric difference quotient (see Appendix~\ref{sec:numerical.gradient.hessian}).  
Since the $\ell_2$-norm of $\hat{\btheta}$ is restricted to be 1, we carefully handle the special case when $\hat{\theta}_0 = 0$, where only Newton's difference quotient is available.  
Second, we use bootstrap to numerically find the CI of $\tilde{\theta}_j$ in step (\ref{equ:asy.CI.thetaj}).  
That is, we randomly select $B$ bootstrap samples and compute the sparse estimator $\hat{\btheta}_b$ and the one-step estimator $\tilde{\btheta}_{b, j}$ for $j \in \{1:d\}$ on each bootstrap sample.  
The CI of $\tilde{\theta}_j$ is then constructed as the $\alpha/2$ and $1 - \alpha/2$ quantiles of $\{ \tilde{\theta}_{b,j} \}_{b = 1}^B$, corresponding to a confidence level of $(1 - \alpha)$.

\paragraph{Estimating the nuisance parameters.}
When the nuisance parameter $\bQ$ depends on the policy parameter $\btheta$, we propose to estimate $\bQ$ using an initial estimator $\check{\btheta}$ that does not depend on $\bQ$ and is a consistent estimator of $\btheta^*$.  
For example, the initial estimate $\check{\btheta}$ could be the one optimizing the same function (\ref{equ:objfunc}) by setting $\hat Q_t=0$, i.e., without augmentation.  
Similar to (\ref{equ:wobjfunc}), the weighted importance sampling is used for numerical stability.  
Then the initial estimator $\check{\btheta}$ is found by  
\begin{equation} \label{equ:winitobjfunc}
    \check{\btheta} = \argmin_{\btheta} \left\{- 
    \frac{1}{n} \sum_{i=1}^n \sum_{t=1}^T \frac{\rho_{i, 1:t}^{\btheta, \bmu}}{\omega^{\btheta, \bmu}_{1:t}} R_{i,t}
    + \lambda_{\check{\btheta}} \norm{\btheta}_1 \right\}
    \qquad \text{s.t. } \norm{\btheta}_2 \le 1.
\end{equation}
This estimate is consistent when the behavior policy $\bmu$ is known and does not require estimation of the nuisance parameters.  

Based on this initial estimate $\check{\btheta}$, we model the nuisance parameter $\bQ$ using a linear model of basis functions for each stage separately.  
Assume that $\phi(\bx_t)$ is a basis function of the covariates $\bx_t$ at some stage $t$ of dimension $d^{\prime}$ and it includes an intercept.  
In practice, $\phi(\bx_t)$ can be taken to be the linear function, polynomial function, Gaussian radial basis functions, splines, wavelet basis, etc. of $\bx_t$ \citep{luckett2020estimating,shi2022statistical}.  
Let $\Phi (\bX_{t}, A_{t}) := [\phi^T (\bX_t), A_t \cdot \phi^T (\bX_t)]^T$  
and we can fit the model $Q_t (\bX_{t}, A_{t}) = \Phi (\bX_{t}, A_{t})^T \bbeta_{t}$ at stage $t$, where $\bbeta_{t}$ is in the dimension $2 d^{\prime}$.  
We discuss three different ways for defining the nuisance parameter $\bQ$.  

\alert{In the first method, we define $Q_t^{\btheta} (\bx_t, a_t) := \bbE^{\btheta} [\sum_{k=t}^T R_k | \bX_t = \bx_t, A_t = a_t]$ as the Q-function at stage $t$.  
Based on the tower property of conditional expectation, we have $Q_t^{\btheta} (\bx_t, a_t) := \bbE^{\btheta} [R_t + Q_{t + 1}^{\btheta} (\bX_{t + 1}, A_{t + 1}) | \bX_t = \bx_t, A_t = a_t]$.  
Following the idea of Q-learning \citep{zhu2019proper}, we can estimate $\hat{\bbeta}_{T}$, \dots, $\hat{\bbeta}_{1}$ backwardly by plugging in the initial $\check{\btheta}$.  
Denote the estimated Q-function at stage $t$ for policy $\pi^{\check{\btheta}}$ as  
\begin{equation} \label{equ:estimate.q.function}
    q^{\check{\btheta}}_{i, t} = R_{i, t} + \sum_{a \in \cA} \pi^{\check{\btheta}} (a | \bX_{i, t + 1}) \Phi (\bX_{i, t + 1}, a)^T \hat{\bbeta}_{t + 1},
\end{equation}
where $\hat{\bbeta}_{T + 1}$ is taken as the zero vector.  
}  
To find a sparse estimate of $\bbeta_t$, we minimize the square loss function with the $L_1$ penalty, i.e.,  
\begin{equation} \label{equ:beta_t1}
    \hat{\bbeta}_{t}^{(1)} = \argmin_{\bbeta_{i,t} \in \bbR^{2 d^{\prime}}} \frac{1}{n} \sum_{i=1}^n \brce{q^{\check{\btheta}}_{i, t} - \Phi (\bX_{i, t}, A_{i, t})^T \bbeta_t}^2 + \lambda_{\bbeta_{t}^{(1)}} \normn{\bbeta_t}_1,
\end{equation}
where $\lambda_{\bbeta_{t}^{(1)}}$ is a tuning parameter.  

The second method is a heuristic algorithm to minimize the variance of the estimator $\hat{V}_t (\btheta, \hat{\beeta})$ of the value function at stage $t$, where  
\[
\hat{V}_t (\btheta, \hat{\beeta}) = \frac{1}{n} \sum_{i=1}^n \brce{ \rho_{i, 1:t}^{\btheta, \bmu} [R_{i,t} - \hat{Q}_t (\bX_{i,t}, A_{i,t})]
    + \rho_{i, 1:(t-1)}^{\btheta, \bmu} \hat{U}_t (\bX_{i,t})}.
\]
\alert{Note that the variance can be decomposed as $\Var (\hat{V}_t (\btheta, \hat{\beeta})) = \bbE \{ \hat{V}_t (\btheta, \hat{\beeta}) \}^2 - \{ \bbE \hat{V}_t (\btheta, \hat{\beeta}) \}^2$.  
Since $\bbE \hat{V}_t (\btheta, \hat{\beeta}) = \bbE^{\btheta} R_t$ for any $\hat{\beeta}$, the second term is irrelevant to the nuisance parameter $\hat{\beeta}$.  
For computational efficiency, we only need to minimize $\bbE \{ \hat{V}_t (\btheta, \hat{\beeta}) \}^2$ using the sample average.
}
With initial estimate $\check{\btheta}$ and the weighted importance sampling, we have  
\begin{equation} \label{equ:beta_t2}
\begin{split}
    \hat{\bbeta}_{t}^{(2)} = & \argmin_{\bbeta_{i,t} \in \bbR^{2 d^{\prime}}} 
    \frac{1}{n} \sum_{i=1}^n \Bigg\{ 
    \frac{\rho_{i, 1:t}^{\check{\btheta}, \bmu}}{\omega^{\check{\btheta}, \bmu}_{1:t}} R_{i,t} - \\
    & \Bigg[ \frac{\rho_{i, 1:t}^{\check{\btheta}, \bmu}}{\omega^{\check{\btheta}, \bmu}_{1:t}} \Phi (\bX_{i, t}, A_{i, t}) - 
    \frac{\rho_{i, 1:(t-1)}^{\check{\btheta}, \bmu}}{\omega^{\check{\btheta}, \bmu}_{1:(t-1)}} \sum_{a \in \cA} \pi^{\check{\btheta}} (a | \bX_{i, t}) \Phi (\bX_{i, t}, a) \Bigg]^T \bbeta_t
    \Bigg\}^2
    + \lambda_{\bbeta_{t}^{(2)}} \normn{\bbeta_t}_1,
\end{split}
\end{equation}
where $\lambda_{\bbeta_{t}^{(2)}}$ is a tuning parameter.  

\alert{Note that both methods are essentially minimizing the asymptotic variance of the value function estimator.  
While the first method estimates $Q_t$ as the Q-function of $\pi^{\btheta^*}$, which achieves semiparametric efficiency \citep{kallus2020double}, the second method directly minimizes the finite sample variance, which will converge to the asymptotic variance.
Furthermore, (\ref{equ:beta_t1}) and (\ref{equ:beta_t2}) both minimize the square loss with an $L_1$ penalty, and can be easily solved by existing Lasso packages with properly constructed predictors and responses.  
}  
To improve the finite sample performance, we refit $\hat{\bbeta}_t$ on its nonzero components for all $t$ using linear regression.  
A refitted Lasso estimator means that we re-estimate the parameter on the support of the original Lasso estimator using the original loss function without the $L_1$ penalty.  
It has been shown that a refitted Lasso estimator usually leads to better finite sample performance than the original Lasso estimator \citep{zhang2014confidence,ning2017general}.  
This refitted estimator may be less biased and less sensitive to the choice of tuning parameters of the penalty.  

In the naive method, we can take $\lambda_{\bbeta_{t}^{(0)}} = \bzero$, so that $Q_t = 0$ for all $t$.  
Then the estimated nuisance parameter $Q_t$ at stage $t$ is  
\begin{equation} \label{equ:Q_t}
    \hat{Q}_t^{(m)} (\bX_{t}, A_{t}) = \Phi (\bX_{1:t}, A_{1:t})^T \hat{\bbeta}_{t}^{(m)}
\end{equation}
for methods $m = 0, 1, 2$.

\paragraph{Optimizing constrained nonconvex nondifferentiable functions.}
To solve for $\check{\btheta}$ and $\hat{\btheta}$, note that (\ref{equ:winitobjfunc}) and (\ref{equ:wobjfunc}) are both constrained nonconvex nondifferentiable optimization problems.  
To deal with the $L_1$ penalty, we use the proximal coordinate descent algorithm \citep{sra2011optimization}.  
To ensure that the $L_2$ norm of the estimated parameter is bounded by one, we normalize the parameter by its $L_2$ norm in each iteration of the coordinate descent.  
Since nonconvex problems may converge to a local minimizer, we start the optimization from different starting points for better numerical results.  
Similar to the procedure for estimating the nuisance parameters, we refit $\check{\btheta}$ and $\hat{\btheta}$ on their support using the trust-region constrained algorithm, which is suitable for minimization problems with constraints. 
Although we require $\normn{\btheta}_2 \le 1$, the estimated $\hat{\btheta}$ actually has $\normn{\hat{\btheta}}_2 = 1$ to approximate a deterministic policy.  
The optimization algorithm is summarized in Algorithm~\ref{alg:PCD}.

The full implementation algorithm for finding the estimate and the CI of high-dimensional policy parameters is summarized in Algorithm~\ref{alg:OSE.bs}.  
Note that all the tuning parameters are selected using the full dataset and then fixed during bootstrap.  

\begin{algorithm}[!t]
    \SetKwInOut{Input}{Input}
    \SetKwInOut{Output}{Output}
    \SetAlgoLined
    \Input{$n$ trajectories $\bD_{1:n}$, the number of bootstrap iterations $B$, the confidence level $1 - \alpha$, the method $m$ for estimating the nuisance parameter.}
    \Output{The sparse estimator $\hat{\btheta}$, the one-step estimator $\tilde{\theta}_j$ with its CI for $j \in \{1:d\}$.}  
    Obtain an initial estimator $\check{\btheta}$ by solving (\ref{equ:winitobjfunc}) using Algorithm~\ref{alg:PCD}, where $\lambda_{\check{\btheta}}$ is tuned by cross-validation\;  
    Estimate the nuisance parameter $\hat{\bQ}^{(m)}$ with $\check{\btheta}$ using Algorithm~\ref{alg:solve.beta}, where $\lambda_{\bbeta_1}, \dots, \lambda_{\bbeta_T}$ are tuned by cross-validation at each stage $t \in \{1:T\}$\;  
    Obtain the sparse estimator $\hat{\btheta}$ by solving (\ref{equ:wobjfunc}) with $\hat{Q}_t^{(m)}$ using Algorithm~\ref{alg:PCD}, where $\lambda_{\btheta}$ is tuned by cross-validation\;  
    \For{$j \in \{1:d\}$}{
        Calculate the one-step estimator $\tilde{\theta}_j$ following the steps in (\ref{equ:w}), (\ref{equ:score}), and (\ref{equ:ose}), where $\lambda_{\bw_j}$ is tuned by cross-validation, and the gradient and Hessian matrix are estimated numerically by (\ref{equ:grad.num}) and (\ref{equ:Hess.num})\;
    }
    \For{$b \in \{1:B\}$}{
        Randomly draw a bootstrap sample of size $n$\;
        Obtain the bootstrap estimator $\hat{\btheta}_{b}$ by refitting (\ref{equ:wobjfunc}) with the bootstrap sample, $\hat{Q}_t^{(m)}$ and $\lambda_{\btheta}$ using Algorithm~\ref{alg:PCD}\;
        \For{$j \in \{1:d\}$}{
            Calculate the one-step estimator $\tilde{\theta}_{b,j}$ following the steps in (\ref{equ:w}), (\ref{equ:score}), and (\ref{equ:ose}) with $\lambda_{\bw_j}$ and the numerically estimated gradient and Hessian matrix\;
        }
    }
    \For{$j \in \{1:d\}$}{
        Obtain the $(1 - \alpha) \times 100\%$ CI of $\tilde{\theta}_j$ by finding the $\alpha/2$ and $1 - \alpha/2$ quantiles of $\{ \tilde{\theta}_{b,j} \}_{b = 1}^B$.
    }
    \caption{Implementation for the optimization and inference of high-dimensional MSTP parameters}
\label{alg:OSE.bs}
\end{algorithm}

\section{Theoretical Results} \label{sec:theory.MSTP}


Denote $s_{\bxi} := \normn{\btheta^*}_0$ and $s_{\bw_j} := \normn{\bw_j^*}_0$ to be the numbers of nonzero elements in the corresponding vectors.  
We assume the following assumptions hold for the variables and the convergence rate of the nuisance parameters.  

\begin{asp} \label{asp:bound.var}
    Assume the rewards $R_t$ are bounded in the sense that $\normn{R_t}_{\infty} \leq r$ for some $r > 0$ and for all $t \in \{1:T\}$.  
    Assume the covariates are bounded such that $\normn{\bX_{t}}_{\infty} \leq z$ and $\absn{\bv_j^{*T} \bX_{t}} \leq z$ for some $z > 0$ and for all $t \in \{1:T\}$.
    Suppose that there exists some constant $u > 0$ such that $\abs{\theta_0^*} \ge u$.  
\end{asp}



\begin{asp} \label{asp:conv.rate}
    We assume that $(s_{\bxi} + s_{\bw_j}) \log d / \sqrt{n} \to 0$.  
    In addition, suppose that $\hat{\beeta} \in \cH_n$ with probability no less than $1 - \Delta_n$, where  
    $$\cH_n := \brce{\beeta \in \cH: \normn{\eta_t}_{\bbP, \infty} \le r, \normn{\eta_t - \bar{\eta}_t}_{\bbP, 2} \leq \delta_n \text{ for all } t},$$  
    and $\delta_n = o(1), \Delta_n = o(1)$ are positive constants.  
\end{asp}  

\begin{asp} \label{asp:V.sc}
    Assume $V(\btheta)$ is $\kappa$-strongly concave at $\btheta^*$. That is, there exists $\kappa > 0$ such that  
    $\agl{\nabla_{\bxi} V(\btheta^*), \Delta_{\bxi}} - [V(\btheta^* + \Delta_{\btheta}) - V(\btheta^*)] \geq \frac{\kappa}{2} \norm{\Delta_{\bxi}}_2^2$
    for all $\Delta_{\bxi} \in \bbB(R)$ for some radius $R$, where $\bbB(R)$ is the ball with radius $R$ defined by the $L_2$ norm, and $\Delta_{\btheta} = (\pm \sqrt{1 - \norm{\Delta_{\bxi}}_2^2}, \Delta_{\bxi})$ is the concatenated error bound for $\theta$.  
\end{asp}

Assumption~\ref{asp:bound.var} requires the boundedness of the variables.  
The conditions about the covariates $\bX_t$ follow the example of generalized linear models in \citet{ning2017general}. 
We assume that the random policy that assigns both actions with equal probabilities will not be the optimal policy.  
Without loss of generality, we assume the intercept is nonzero to achieve this.  
If the intercept is zero, we can shift the $j$th feature by a constant $u / \theta_j$ for some coordinate $j$ where $\theta_j \ne 0$.  

Assumption~\ref{asp:conv.rate} deals with the convergence rate of the nuisance parameters.  
Since we assume that we are using data from an MRT and $\bmu$ is known, we do not need the model for $Q_t$ to be correctly specified.  
Moreover, $\hat{Q}_t$ can converge to its limit at any rate.  
This assumption can be easily satisfied by almost all learning methods, including regularized methods like Lasso, ridge regression, or elastic net.  

Assumption~\ref{asp:V.sc} is used to verify the restricted strong convexity, which is one of the sufficient conditions for proving the convergence of parameters regularized by the $L_1$ penalty.  
\alert{The local concavity at $\btheta^*$ is empirically verified with numerical examples in Appendix~\ref{sec:true.value.function}.
}  

\alert{
\begin{lem} \label{lem:theta.conv}
    Under Assumptions~\ref{asp:nuca}-\ref{asp:V.sc},  
    when $\lambda_{\btheta} \simeq \sqrt{\log d / n}$,  
    we have  
    \begin{equation}
        \normn{\hat{\bxi} - \bxi^*}_1 = \cO_{\bbP} (s_{\bxi} \sqrt{\log d / n}). \label{equ:l1conv.theta}
    \end{equation}
\end{lem}

Lemma~\ref{lem:theta.conv} shows that the sparse estimator $\hat{\bxi}$, which is a permutation of $\hat{\btheta}$, is a consistent estimator of $\bxi^*$.  
We will then show that the one-step estimator $\tilde{\theta}_j$ is a consistent estimator of $\theta_j^*$ with a $\sqrt{n}$ convergence rate.  
}

\begin{asp} \label{asp:Sigma.pd}
    Suppose that the covariance matrix $\bSigma^* = \Var [\nabla_{\bxi} l_0 (\btheta^*, \bar{\beeta})]$ is positive definite and finite.  
\end{asp}

\alert{Assumption~\ref{asp:Sigma.pd} is used in the multivariate central limit theorem to prove the asymptotic normality of the score function.
}  
We follow the proof of \citet{ning2017general} to decorrelate $\bnu_j$ from the parameter $\theta_j$ that we are interested in, and the assumptions will be used to verify the conditions in Theorem 3.2 \citep{ning2017general}.  
The main challenge lies in the nuisance parameters $\beeta$, which need to be estimated and are also high-dimensional.  
We will use the technique in \citet{chernozhukov2018double} to decorrelate the nuisance parameters.

\begin{thm} \label{thm:main}
    Under Assumptions~\ref{asp:nuca}-\ref{asp:Sigma.pd}, the one-step estimator $\tilde{\theta}_j$ satisfies  
    \begin{equation}
        \sqrt{n} (\tilde{\theta}_j - \theta_j^*) I^{*}_{\theta_j | \bnu_j} / \sigma_j^{*1/2} \Rightarrow N(0, 1)  
    \end{equation}
    for $j \in \{1:d\}$,  
    where $\sigma_j^*$ is defined in (\ref{equ:def.sigma.S}).  
\end{thm}

Theorem \ref{thm:main} shows that the plug-in one-step estimator is asymptotically normal.  
Since $\hat{I}_{\theta_j | \bnu_j}$ is consistent for $I^{*}_{\theta_j | \bnu_j}$ and $\hat{\sigma}_j$ is consistent for $\sigma_j^*$,  
we can construct CIs as in (\ref{equ:asy.CI.thetaj}) based on the theorem.  
Note that $\sigma_j^*$ depends on the nuisance parameter $\hat{\beeta}$.  
Therefore, although the limit and the convergence rate of $\hat{\beeta}$ do not affect the convergence rate of $\tilde{\theta}_j$, the limit $\bar{\beeta}$ does influence the asymptotic variance of $\tilde{\theta}_j$.  

\section{Simulation Study} \label{sec:simulation.MSTP}

\begin{figure}[t]
    \centering
    \includegraphics[width=\textwidth]{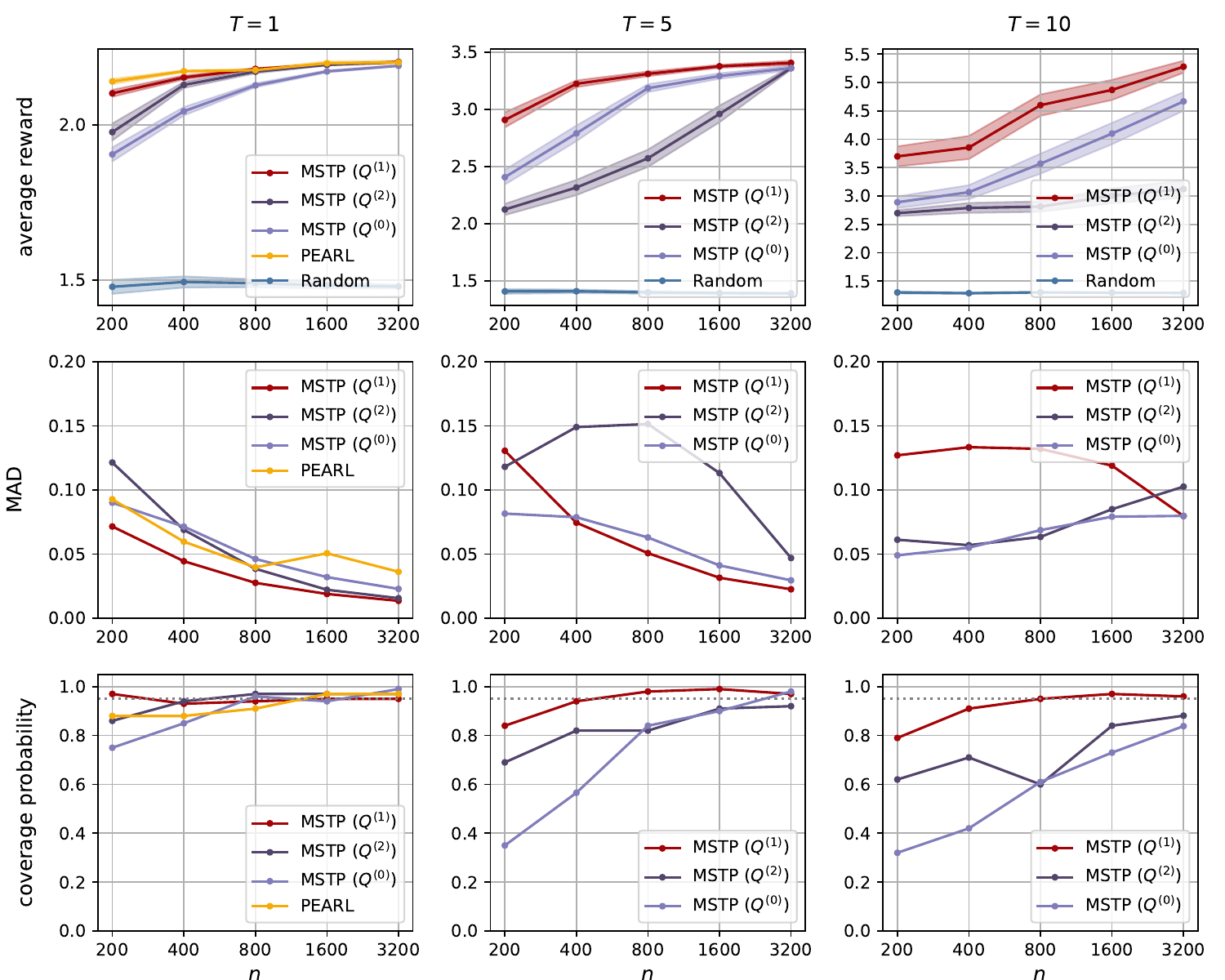}
    \caption{The first row presents the average reward of MSTP, PEARL, and the random policy in Scenario 1. The second and third rows show the MAD and coverage probability of $\theta_1$. The columns correspond to different values of $T$. The dotted line indicates the nominal coverage probability for $\theta_1$.}
    \label{fig:simulation.v1}
\end{figure}

\begin{figure}[t]
    \centering
    \includegraphics[width=\textwidth]{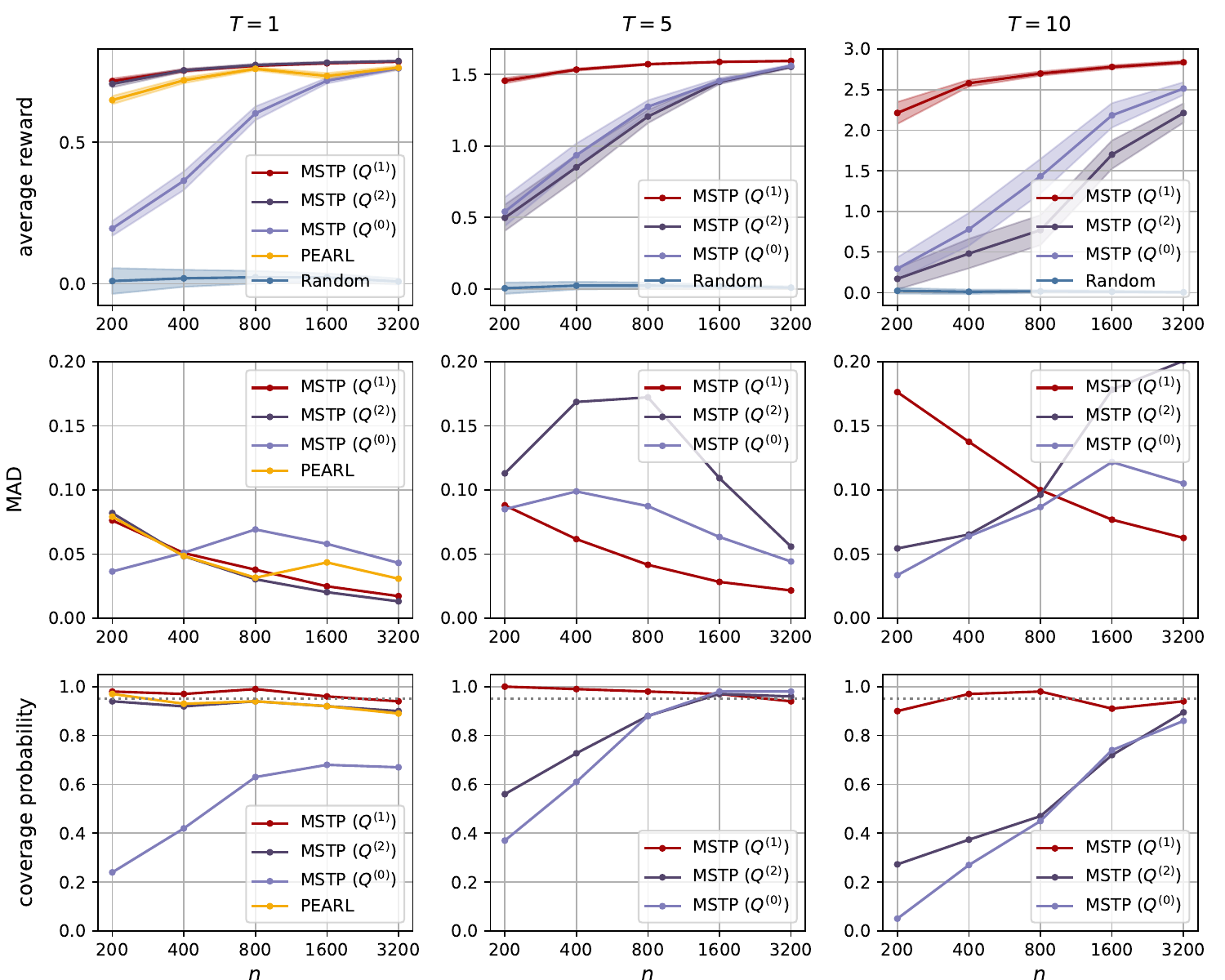}
    \caption{The first row presents the average reward of MSTP, PEARL, and the random policy in Scenario 2. The second and third rows show the MAD and coverage probability of $\theta_1$. The columns correspond to different values of $T$. The dotted line indicates the nominal coverage probability for $\theta_1$.}
    \label{fig:simulation.v2}
\end{figure}

\alert{In this section, we test our proposed algorithm for the optimization and inference of low-dimensional parameters in high-dimensional settings in two simulated scenarios.  

Assume that the data are from an MRT, and the action $A_{i,t}$ takes values from $\{-1,1\}$ with equal probability at each stage $t$ for all individuals $i$.  
The variable $X_{i, t, j}$ represents the $j$th variable of individual $i$ at stage $t$.  
The first two variables for $j = 1, 2$ are important variables that influence the reward, while the other variables for $j \ge 3$ are noise variables.  
The initial states are generated from a multivariate normal distribution, i.e., $\bX_{i,1} \sim N(\bzero, \Sigma)$ for each individual $i \in \{1:n\}$.  
In the covariance matrix $\Sigma$, the variance of variable $j$ is $\Sigma_{j,j} = 4$ for all $j \in \{1:d\}$.  
To allow dependence between high-dimensional features, the covariance between the important variables is $\Sigma_{1, 2} = \Sigma_{2, 1} = 1$, and the covariance between variables $j - 1$ and $j$ is $\Sigma_{j - 1, j} = \Sigma_{j, j - 1} = 0.2$ for $j \in \{3:d\}$.  
Other entries in $\Sigma$ take a value of zero.  
For stages $t \in \{2:(T + 1)\}$, the variables and the rewards are generated as follows.  
In Scenario 1,  
\begin{align*}
    X_{i,t,1} =& 0.8 \tilde{X}_{i,t-1,1} + 0.3 A_{i,t-1} \tilde{X}_{i,t-1,1} + 0.2 \tilde{X}_{i,t-1,2} + \epsilon_{i,t,1}, \\
    X_{i,t,2} =& 0.8 \tilde{X}_{i,t-1,2} - 0.3 A_{i,t-1} \tilde{X}_{i,t-1,2} + A_{i,t-1} \tanh \{( \tilde{X}_{i,t-1,1} - \tilde{X}_{i,t-1,2}) / 2 \} + \epsilon_{i,t,2}, \\
    X_{i,t,j} =& 0.9 \tilde{X}_{i,t-1,j} + \epsilon_{i,t,j}, \quad j \in \{3:d\}, \\
    R_{i,t-1} =& \log \brce{ 1 + \exp (X_{i,t,1} + X_{i,t,2}) } - 0.5 A_{i,t},
\end{align*}
and in Scenario 2,  
\begin{align*}
    X_{i,t,1} =& 0.8 \tilde{X}_{i,t-1,1} + 0.3 A_{i,t-1} \tilde{X}_{i,t-1,1} + 0.1 \tilde{X}_{i,t-1,2} + \epsilon_{i,t,1}, \\
    X_{i,t,2} =& 0.8 \tilde{X}_{i,t-1,1} + 0.2 A_{i,t-1} \tilde{X}_{i,t-1,2} + 0.2 \tilde{X}_{i,t-1,2} + \epsilon_{i,t,2}, \\
    X_{i,t,j} =& 0.9 \tilde{X}_{i,t-1,j} + \epsilon_{i,t,j}, \quad j \in \{3:d\}, \\
    R_{i,t-1} =& X_{i,t,1} + X_{i,t,2} - 0.6 A_{i,t},
\end{align*}
for $i \in \{1:n\}$.  
Here, $\tilde{X}_{i,t,j}$ is the sequence of exponentially weighted moving averages of $X_{i,t,j}$ such that  
$\tilde{X}_{i,1,j} = X_{i,1,j}$ and  
$\tilde{X}_{i,t,j} = 0.2 \tilde{X}_{i,t-1,j} + 0.8 X_{i,t,j}$ for all $j \in \{1:d\}$ and $t \geq 2$.  
Let $\epsilon_{i,t,j} \overset{i.i.d.}{\sim} N(0, 0.4^2)$ for all $i$, $j$, and $t \ge 2$.  
Under these scenarios, the Markov assumption is violated since the reward $R_{i, t}$ is a function of $\bX_{i, t+1}$, which depends on the states in all previous stages. 

In this simulation, we experiment with three different constructions of the nuisance parameter $\bQ$ as discussed in Section~\ref{sec:implementation}.  
The number of stages $T$ is taken to be 1, 5, or 10.  
The dimension is fixed at $d = 50$.  
The sample size $n$ ranges among 200, 400, 800, 1600, and 3200.  
In our simulation, the scaling parameter $\tau$ is fixed at $0.2$, since our experiments show that a larger $\tau$ leads to a stochastic policy far from the true optimal deterministic policy, and a smaller $\tau$ may cause unstable computation.  
We take the number of bootstraps to be $B = 100$.  
The value function of each estimated policy is calculated based on an independent test set of size $200{,}000$ generated by this policy.  
We repeat the whole procedure $W = 100$ times for each scenario.  

The simulation results for Scenario 1 and Scenario 2 are presented in Figures~\ref{fig:simulation.v1} and~\ref{fig:simulation.v2}, respectively.  
We report the average reward for the policy $\pi^{\hat{\btheta}}$ based on the sparse estimator $\hat{\btheta}$, and the mean absolute deviation (MAD) and the coverage probability of the one-step estimator $\tilde{\btheta}$.  
We find the true minimizer $\btheta^*$ of the loss function within the class $\Pi$ by grid-search for each $T$.  
In particular, we estimate the value function on an independent test set of size $200{,}000$ for $\btheta$ on the grids inside the unit ball.  
The true value of $\btheta^*$ in different scenarios is reported in Appendix~\ref{sec:true.value.function}.  
}

We compare our proposed method with Penalized Efficient Augmentation and Relaxation Learning (PEARL) \citep{liang2022estimation} when $T = 1$.  
PEARL is a method for estimating the optimal Individualized Treatment Rule (ITR) and conducting statistical inference from high-dimensional data in single-stage decision problems.  
It utilizes the data-splitting method to account for the slow convergence rate of nuisance parameter estimations.  
In addition, it follows the inference procedure in \citet{ning2017general}, first finding a sparse estimator and then obtaining the one-step estimator.  
We use the package \texttt{ITRInference} for implementation \citep{liang2022estimation}.  
\alert{Since an ITR generated by PEARL takes the selected treatment with probability one, any positive multiplication of $\hat{\btheta}_{\text{PEARL}}$ will yield the same ITR.  
However, due to the strict convexity of their loss function, the estimated optimal parameter $\hat{\btheta}_{\text{PEARL}}$ is always unique.  
Therefore, the package does not impose requirements on the scale of the parameters.  
Their simulation study estimated the coverage probability of the constructed CI based on the limit of parameter estimates, which may not be identical to the true parameters for data generation.  
In practice, we rescale all parameters so that the average one-step estimator $1/W \sum_{w=1}^W \tilde{\btheta}_{\text{PEARL}, w, 1}$ at coordinate $j = 1$ matches $\btheta^*_{1}$ at $j = 1$, where $\btheta^*$ is the true optimal parameter of a deterministic policy.  
}

\alert{For a better comparison across different values of $T$, we report the average reward $\bar{V} (\btheta) = \frac{1}{T} V (\btheta)$ instead of the value function $V (\btheta)$.  
The first rows in Figures~\ref{fig:simulation.v1} and~\ref{fig:simulation.v2} present the mean of $\bar{V} (\hat{\btheta})$ and its 95\% CI based on the standard error across $W$ replications.  
The sparse estimator $\hat{\btheta}$ constructed using different nuisance parameters $Q^{(0)}, Q^{(1)}, Q^{(2)}$ is compared with the random policy used to generate the data.  
We observe that in both scenarios, the nuisance parameter $Q^{(1)}$ yields the largest average reward and the smallest standard error, compared to the other two nuisance parameters and the random policy.  
Although both $Q^{(1)}$ and $Q^{(2)}$ essentially minimize the asymptotic variance of the value function estimator, $Q^{(2)}$ depends on the importance sampling weight and becomes numerically unstable as $T$ increases.  
When $T = 1$, we further compare with PEARL, since it is only applicable in single-stage decision problems.  
PEARL estimates a deterministic policy, which yields a higher average reward in Scenario 1.  
Although the proposed method estimates a stochastic policy, it is stable when $T = 1$ and yields a larger average reward in Scenario 2.

We further report the MAD and the coverage probability of the important variable $\theta_1$ in the second and third rows of Figures~\ref{fig:simulation.v1} and~\ref{fig:simulation.v2}.  
The MAD and the coverage probability of the second important variable $\theta_2$ and the noise variables $\theta_{3:d}$ are reported in Appendix~\ref{sec:simulatio.additional.mad.cp}.  
For the one-step estimator of PEARL, the MAD is approximated as $\tilde{\sigma}_{\text{PEARL}} / 1.4826$, where $\tilde{\sigma}_{\text{PEARL}}$ is the estimated standard deviation in the asymptotically normal distribution of the one-step estimator.  
We observe that the coverage probabilities of $\tilde{\btheta}$ are concentrated near the nominal coverage of $95\%$ when the sample size is large enough.  
When $T$ is small, the CIs achieve the nominal coverage probability even with a small sample size.  
However, the difficulty of policy learning and inference increases as $T$ increases.  
More samples are needed to learn the optimal policy and reach the nominal coverage probability.  
The MAD generally decreases with the increasing sample size if the sparse estimator is well estimated.  
The MAD of MSTP with $Q^{(2)}$ or $Q^{(0)}$ may be small when $n = 200, 400$ for $j = 1, 2$ since $\hat{\theta}_1$ or $\hat{\theta}_2$ are estimated to be zero on a large portion of bootstrap samples, but the one-step estimator cannot correct the bias with a small sample size.  
The MAD of $\tilde{\btheta}$ based on $Q^{(1)}$ is smaller than that of the other two nuisance parameters and PEARL, while achieving a coverage probability even closer to the nominal value.  
Therefore, the above experiments demonstrate the influence of the nuisance parameters on the asymptotic efficiency, and the advantage of using $Q^{(1)}$ in the doubly robust estimator.  
}

\section{Real Data Analysis} \label{sec:real.data}

\alert{The OhioT1DM dataset is a collection of data from individuals with T1D collected through continuous glucose monitors (CGMs), insulin usage, and self-reported events \citep{marling2020ohiot1dm}.  
It is designed to support research in blood glucose prediction, diabetes management, and related healthcare applications.  
In the 2018 cohort, 6 patients participated in an 8-week study.  
We apply the proposed method to the OhioT1DM dataset to learn a treatment policy for patients with T1D.  

After excluding extremely rare events, we use 12 features to construct the MSTP, including the current glucose level, the long-acting basal insulin level, carbohydrate estimate for a meal, and others.  
Since the glucose level, heart rate, galvanic skin response (GSR), skin temperature, air temperature, and step counts are recorded every 5 minutes using CGM, the treatment decision is defined as whether to deliver short-acting bolus insulin every 5 minutes.  
The reward is defined as a function of the glucose level in the next 5 minutes, following \citet{shi2020does}.  
We take $T = 12$, so that the value function is the sum of rewards over one hour.  
The detailed description of the features and the data cleaning methods are provided in Appendix~\ref{sec:real.data.details}.  

Based on the results in Section~\ref{sec:simulation.MSTP}, we use the nuisance parameter $Q^{(1)}$ in Algorithm~\ref{alg:OSE.bs}.  
In our analysis, the propensity score is estimated using logistic regression with an $L_1$ penalty.  
All features have been standardized.  
The reward is standardized before the analysis but is returned to the original scale in the following results.  

To estimate the mean and standard error of the average reward, we merge the training and test examples in the original dataset and obtain a combined dataset of 7824 trajectories.  
We then randomly split the combined dataset into training and test sets with probabilities 0.8 and 0.2 for $W = 100$ replications.  
The average reward of the behavior policy is estimated by taking the average of all rewards in the training set.  
The average reward of the estimated policy is calculated using the IPW estimator with normalized weights on the test set.  
After applying Algorithm~\ref{alg:OSE.bs} for policy learning, the average reward for the behavior policy is $-6.260$, with a standard error of $0.008$; the average reward for the estimated policy is $-2.367$, with a standard error of $0.100$.  
This suggests that the estimated policy could significantly improve upon the behavior policy.  

We use the training and test set splitting in the original dataset to construct the policy using the sparse estimator $\hat{\btheta}$ and to construct the CIs of $\tilde{\btheta}$, which are reported in Table~\ref{tab:policy.tidm}.  
Based on the results in Table~\ref{tab:policy.tidm}, the current glucose level, meal, hypoglycemic episode, heart rate, and GSR are selected, but only the glucose level is significant.  
The estimated policy suggests delivering bolus insulin when the current glucose level is high and avoiding it when the glucose level is already low.  
Since ``hypoglycemic episode'' indicates low glucose levels, one might expect its coefficient to be negative, but the estimated coefficient is positive.  
This discrepancy arises because hypoglycemic episodes are self-reported, reflecting patients' awareness of the condition and prompting them to consume meals or snacks immediately.  
Out of 78 self-reported events in the entire dataset, patients consumed a meal in 74 events.  
As a result, ``hypoglycemic episode'' becomes an indicator of an increase in blood glucose levels in subsequent times.  
Nevertheless, the coefficient is not significant based on its CI, as this variable is highly correlated with glucose levels.  


\begin{table}[t] 
    \centering
    \begin{tabular}{ccccc}
        \toprule
        $j$ & Feature & $\hat{\theta}_j$ & $\tilde{\theta}_j$ & CI of $\tilde{\theta}_j$ \\
        \midrule
        0 & Intercept & 0.000 & & \\
        1 & Basal & 0.000 & -0.001 & (-0.354, 0.039) \\
        2 & Glucose level & 0.872 & 0.870 & (0.302, 0.991) \\
        3 & Meal & 0.463 & 0.460 & (-0.032, 0.553) \\
        4 & Sleep & 0.000 & -0.001 & (-0.093, 0.045) \\
        5 & Work & 0.000 & -0.001 & (-0.057, 0.060) \\
        6 & Hypoglycemic episode & 0.084 & 0.081 & (-0.041, 0.113) \\
        7 & Exercise & 0.000 & -0.000 & (-0.061, 0.290) \\
        8 & Heart rate & 0.102 & 0.097 & (-0.196, 0.171) \\
        9 & GSR & 0.084 & 0.079 & (-0.194, 0.547) \\
        10 & Skin temperature & 0.000 & -0.002 & (-0.119, 0.085) \\
        11 & Air Temperature & 0.000 & -0.002 & (-0.095, 0.091) \\
        12 & Steps & 0.000 & -0.000 & (-0.150, 0.772) \\
        \bottomrule
        \end{tabular}
    \caption{The estimated sparse policy parameter $\hat{\theta}_j$ and the CIs of the one-step estimator $\tilde{\theta}_j$.  
    }  
    \label{tab:policy.tidm}
\end{table}
}

\section{Discussion} \label{sec:discussion.MSTP}

In this work, we focus on the multi-stage decision problem and propose a method for estimating the high-dimensional MSTP and the CIs of its parameters.  
We first estimate the MSTP based on the AIPWE of the value function with an $L_1$ penalty to encourage sparsity and an $L_2$ constraint to avoid unboundedness.  
Then, we compute the one-step estimators, which are asymptotically normal and suitable for statistical inference.  
The proposed one-step estimator is shown to achieve nominal coverage probabilities in simulation studies.  
While the choice of nuisance parameter estimation does not affect the convergence rate of the low-dimensional policy parameter, it influences the asymptotic efficiency.  
We compare different estimators $\hat{\bQ}$ in the simulation study and show that $Q_t$ constructed as the Q-function generates the highest value function and the smallest MAD of the estimated policy parameters.  
Our code for simulation can be accessed at \href{https://github.com/DqGao/MSTP}{\texttt{https://github.com/DqGao/MSTP}}.

\alert{We assume that the behavior policy $\bmu$ is known, as in MRTs.  
When data from an observational study are used for learning, $\bmu$ is unknown and needs to be estimated.  
Our framework naturally extends to observational studies.  
The one-step estimator can still achieve the $\sqrt{n}$ convergence rate as long as either the Q-function or the propensity score is correctly specified.  
Since we are most interested in the significance of the selected variables, we focus on constructing confidence intervals for each parameter separately.  
Similar methods can be used to construct confidence regions for multi-dimensional parameters \citep{ning2017general}, if the joint effect of multiple features is of interest.  
In this work, we focus on binary treatments.  
Extending the method to multiple treatments is an interesting open problem \citep{qi2020multi,ma2022learning,ma2023learning}.  
A possible approach is to replace the sigmoid probability in the policy class $\Pi$ with a softmax probability, but the conditions for valid statistical inference require further investigation.

For decision problems with a long horizon $T$, the ratio $\rho_{1:t}^{\btheta, \bmu}$ can become extremely unstable, especially when the feature dimension $d$ is large.  
To handle this instability, an alternative method based on the marginalized distribution of the current state and action has been proposed in \citet{liu2018breaking,kallus2020double,liao2022batch}.  
However, it requires the Markov assumption, which is often violated in mHealth applications \citep{shi2020does}.  
Moreover, this approach requires either the estimation of a marginal distribution for each stage $t$, or the convergence of the offline data to the stationary distribution \citep{kallus2020double,liao2022batch}.  
Particularly, estimating the marginal distribution becomes challenging with high-dimensional continuous features.  
Due to the curse of dimensionality \citep{hastie2017elements}, the data for density ratio estimation can be very sparse.
\citet{luckett2020estimating} only uses the importance sampling weight at one decision stage based on the Bellman optimality equation, but it requires the value function to be estimated at each step of the gradient descent algorithm with respect to the policy parameters.
Another approach under the Markov assumption is to use partial history importance weighting with assumptions on mixing and overlap between the behavior and target policies \citep{hu2023off}.  
If the Markov chain under the target policy mixes quickly, the policy before mixing has little impact on later rewards.  
Thus, only the importance sampling weight after mixing is needed, efficiently reducing the variance of the value function estimator for a long horizon.  
}

\section*{Acknowledgment}
This research was funded by NIH grants R01GM124104, R01GM126550, P50DA054039, P41EB028242, R01HL125440-06A1, and UH3DE028723, and NSF grants DMS 2100729 and SES 2217440.

\appendix
\section{Implementation Details}

\alert{In this section, we present the implementation details omitted in Section~\ref{sec:implementation}.  

\subsection{Estimating the Nuisance Parameters}  

In Section~\ref{sec:implementation}, we discussed three different methods for estimating the nuisance parameter $\bQ$.  
In Algorithm~\ref{alg:solve.beta}, we present the full algorithm for estimating $\hat{\bQ}^{(m)}$ for $m = 0, 1, 2$.  
The tuning parameter $\lambda_{\bbeta_t}^{(m)}$ is selected as the one that minimizes the mean squared error in (\ref{equ:beta_t1}) or (\ref{equ:beta_t2}) on the validation set.  
While for the method $m = 1$, the parameters $\bbeta_{t}^{(1)}$ must be estimated backwardly from the last stage to the first one, the parameters $\bbeta_{t}^{(2)}$ for the method $m = 2$ do not require a specific estimation order.  

\begin{algorithm}[!t]
    \SetKwInOut{Input}{Input}
    \SetKwInOut{Output}{Output}
    \SetAlgoLined
    \Input{The method $m$, the initial estimate $\check{\btheta}$, the tuning parameter $\lambda_{\bbeta_t}^{(m)}$.}
    \Output{The estimated nuisance parameters $\hat{\bQ}^{(m)} = [\hat{Q}_{1:T}^{(m)}]$.}
    \uIf{$m = 0$}{
        Take $\lambda_{\bbeta_{t}^{(0)}} = \bzero$\;
    }
    \uElseIf{$m = 1$}{
        \For{$t = T, \dots, 1$}{
            Obtain $q^{\check{\btheta}}_{i, t}$ with $\hat{\bbeta}_{t + 1}$ by (\ref{equ:estimate.q.function})\;
            Estimate $\hat{\bbeta}_{t}^{(1)}$ by solving (\ref{equ:beta_t1}) with $\check{\btheta}$ and $\lambda_{\bbeta_t}^{(1)}$ using a Lasso solver\;
            Refit $\hat{\bbeta}_{t}^{(1)}$ on its nonzero components using linear regression\;
        }
    }
    \ElseIf{$m = 2$}{
        \For{each $t \in \{1:T\}$}{
            Estimate $\hat{\bbeta}_{t}^{(2)}$ by solving (\ref{equ:beta_t2}) with $\check{\btheta}$ and $\lambda_{\bbeta_t}^{(2)}$ using a Lasso solver\;
            Refit $\hat{\bbeta}_{t}^{(2)}$ on its nonzero components using linear regression\;
        }
    }
    Find $\hat{Q}_t^{(m)}$ by plugging $\hat{\bbeta}_{t}^{(m)}$ into (\ref{equ:Q_t}) for each $t$.  
    \caption{Estimate the Nuisance Parameter $\bQ$}
    \label{alg:solve.beta}
\end{algorithm}
}

\subsection{Optimization with $L_1$ Penalty and $L_2$ Constraint}  

We present the optimization method for a nonconvex loss function with an $L_1$ penalty and $L_2$ constraint as described in Section~\ref{sec:implementation} in Algorithm~\ref{alg:PCD}.  
Notice that to find the global minimizer of a nonconvex function, we need to try different starting points.  
Since running the full algorithm from multiple starting points is computationally heavy,  
we compare the function values of multiple points for each coordinate separately at the beginning.  

\alert{The tuning parameter $\lambda_{\check{\theta}}$ for the initial estimator $\check{\theta}$ or $\lambda_{\theta}$ for the sparse estimator $\hat{\theta}$ is selected as the one that maximizes the estimated value function $\hat{V} (\btheta)$ on the validation set.  
Note that for $\lambda_{\check{\theta}}$, $Q_t = 0$ in $\hat{V} (\btheta)$.  
We use the function \texttt{scipy.optimize.minimize} with \texttt{method=`trust-constr'} in Python to solve the constrained minimization problem when refitting $\check{\btheta}$ and $\hat{\btheta}$ on their support.  
}

\begin{algorithm}[!t]
    \SetKwInOut{Input}{Input}
    \SetKwInOut{Output}{Output}
    \SetAlgoLined
    \Input{A loss function $\ell$ to be minimized with $L_1$ penalty and $L_2$ constraint, the tuning parameter $\lambda$ for the $L_1$ penalty, a starting point $\btheta^{(0)}$, the error threshold $e$, the maximum number of iterations $M$.}
    \Output{The estimated parameter $\check{\btheta}$ or $\hat{\btheta}$.}
    \For{each coordinate $j \in \{0:d\}$}{
        Find the best starting point of the coordinate $j$ using grid search\;
    }
    Normalize $\btheta^{(1)}$ by its $\ell_2$-norm\;
    Initialize the number of iterations $m = 1$\;
    \While{$m \le M$ and $\normn{\btheta^{(m)} - \btheta^{(m - 1)}}_2 \ge e$}{
        $m \leftarrow m + 1$\;
        \For{each coordinate $j \in \{0:d\}$}{
            Find the minimizer of $\ell$ with respect to the current coordinate using the BFGS algorithm
            \[
            \tilde{\theta}^{\prime (m)}_j = \argmin_{\theta_j} \ell((\theta^{(m)}_{0:(j-1)}, \theta_j, \theta^{(m-1)}_{(j+1):d}));
            \]
            Shrink $\tilde{\theta}^{\prime (m)}_j$ using soft-thresholding
            \[
            \theta^{(m)}_j = \sgn{\tilde{\theta}^{\prime (m)}_j} \max(\absn{\tilde{\theta}^{\prime (m)}_j} - \lambda, 0);
            \]
            Normalize $(\theta^{(m)}_{0:j}, \theta^{(m-1)}_{(j+1):d})$ by its $\ell_2$-norm\;
        }
    }
    Re-estimate $\check{\btheta}$ or $\hat{\btheta}$ by minimizing the loss function $\ell$ with the $L_2$ constraint on the support of $\btheta^{(m)}$ using the trust-region constrained algorithm\;
    \caption{Optimization of a Loss Function with $L_1$ Penalty and $L_2$ Constraint}
    \label{alg:PCD}
\end{algorithm}

\subsection{Obtaining the One-Step Estimator}  

The tuning parameter $\lambda_{\bw_j}$ is chosen for each feature $j$ separately.  
It is selected as the value that minimizes the projection error in (\ref{equ:w}) on the validation set.  
To estimate the Dantzig-type estimator in (\ref{equ:w}), we use the package \texttt{cvxpy} to solve the constrained convex minimization problem.

\subsection{Numerical Computation of the Gradient and Hessian matrix of Loss Function} \label{sec:numerical.gradient.hessian}

The direct estimation of the gradient and Hessian matrix of $\rho_{i, 1:t}^{\btheta, \bmu}$ is unstable due to the probability product, which may cause even larger variability when constructing CIs for the one-step estimator and lead to undercoverage of the true parameter.  
Therefore, we propose to calculate them numerically.  
An estimate of the partial derivative at the coordinate $j \in \{1:d\}$ is calculated using the symmetric difference quotient as  
\begin{equation} \label{equ:grad.num}
    \widehat{\nabla}_{\theta_j} \ell (\btheta, \beeta) = \frac{\ell (\btheta^+, \beeta) - \ell (\btheta^-, \beeta)}{2 h_j}, 
\end{equation}
where  
\begin{align}
    h_j &= \min \brce{\frac{1}{\sqrt{nT}}, \sqrt{1 - \sum_{l \ne 0, j} \theta_l^2} - \absn{\theta_j}}, \nonumber \displaybreak[1]\\
    \theta_j^+ &= \theta_j + h_j, \nonumber \\
    \theta_0^+ &= \sgn{\theta_0} \sqrt{1 - \sum_{l \ne 0, j} \theta_l^2 - (\theta_j^+)^2}, \label{equ:num.grad.theta0} \\
    \theta_l^+ &= \theta_l \quad \text{for } l \ne 0, j, \label{equ:num.grad.thetal} 
\end{align}
and $\btheta^-$ can be defined similarly with $\theta_j^- = \theta_j - h_j$.  
Here, $1 / \sqrt{nT}$ is the common value for $h_j$ used in numerical computations.  
However, since the $\ell_2$-norm is restricted to be 1 in our case, $h_j$ sometimes needs to be even smaller depending on the initial estimate $\theta_j$.  
If $\theta_0 = 0$, then $\sqrt{1 - \sum_{l \ne 0, j} \theta_l^2} - \absn{\theta_j} = 0$.  
In this case, when $\theta_j \ne 0$, we estimate the gradient using Newton's difference quotient as  
\[
\widehat{\nabla}_{\theta_j} \ell (\btheta, \beeta) = \frac{\ell (\btheta^+, \beeta) - \ell (\btheta, \beeta)}{- h_j \sgn{\theta_j}}, 
\]
where  
\begin{align*}
    h_j &= \min \brce{\frac{1}{\sqrt{nT}}, 2 \abs{\theta_j}}, \\
    \theta_j^+ &= \theta_j - h_j \sgn{\theta_j},
\end{align*}
and the other coordinates $\theta_l^+$ for $l \ne j$ are calculated as in (\ref{equ:num.grad.theta0}) and (\ref{equ:num.grad.thetal}).  
When $\theta_0 = \theta_j = 0$, we use the gradient of $\btheta^{\prime}$ as an approximation, where  
\begin{equation} \label{equ:num.grad.thetaj.0}
    \theta_0^{\prime} = 0, \qquad
    \theta_j^{\prime} = \frac{1}{\sqrt{nT}}, \qquad
    \theta_l^{\prime} = \theta_l \sqrt{1 - (\theta_j^{\prime})^2} \text{ for } l \ne 0, j.  
\end{equation}

Similarly, we use the symmetric difference quotient to find an estimate of the Hessian matrix:  
\begin{equation} \label{equ:Hess.num}
    \widehat{\nabla}_{\theta_j \theta_k} \ell (\btheta, \beeta) = \frac{\ell (\btheta^{++}, \beeta) - \ell (\btheta^{+-}, \beeta) - \ell (\btheta^{-+}, \beeta) + \ell (\btheta^{--}, \beeta)}{(2 h_{jk})^2}.
\end{equation}  
When $j \ne k$,  
\begin{align}
    h_{jk} &= \min \brce{\frac{1}{\sqrt{nT}}, \frac{1 - \sqrt{\sum_{l \ne 0} \theta_l^2}}{\sqrt{2}}}, \nonumber \\
    \theta_j^{++} &= \theta_j + h_{jk}, \qquad
    \theta_k^{++} = \theta_k + h_{jk}, \nonumber \\
    \theta_0^{++} &= \sgn{\theta_0} \sqrt{1 - \sum_{l \ne 0, j, k} \theta_l^2 - (\theta_j^{++})^2 - (\theta_k^{++})^2}, \label{equ:num.Hess.jk.theta0} \\
    \theta_l^{++} &= \theta_l \quad \text{for } l \ne 0, j, k,  \label{equ:num.Hess.jk.thetal}
\end{align}
and  
$\btheta^{+-}, \btheta^{-+}, \btheta^{--}$  
can be defined similarly with  
\begin{align*}
    \theta_j^{+-} = \theta_j + h_{jk}, \qquad \theta_k^{+-} = \theta_k - h_{jk}, \\
    \theta_j^{-+} = \theta_j - h_{jk}, \qquad \theta_k^{-+} = \theta_k + h_{jk}, \\
    \theta_j^{--} = \theta_j - h_{jk}, \qquad \theta_k^{--} = \theta_k - h_{jk}.
\end{align*}  
When $j = k$,  
\begin{align}
    h_{jj} &= \min \brce{\frac{1}{\sqrt{nT}}, \frac{1}{2} \prth{\sqrt{1 - \sum_{l \ne 0, j} \theta_l^2} - \absn{\theta_j}}}, \nonumber \\
    \theta_j^{++} &= \theta_j + 2 h_{jj}, \nonumber \\
    \theta_0^{++} &= \sqrt{1 - \sum_{l \ne 0, j} \theta_l^2 - (\theta_j^{++})^2}, \label{equ:num.Hess.jj.theta0} \\
    \theta_l^{++} &= \theta_l \quad \text{for } l \ne 0, j, k, \label{equ:num.Hess.jj.thetal}
\end{align}
and $\btheta^{+-}, \btheta^{-+}, \btheta^{--}$  
can be defined similarly with  
$$\theta_j^{+-} = \theta_j^{-+} = \theta_j, \qquad \theta_j^{--} = \theta_j - 2 h_{jj}.$$  
If $\theta_0 = 0$ and $\theta_j, \theta_k \ne 0$, we instead use Newton's difference quotient to estimate the Hessian matrix as  
\begin{equation*}
    \widehat{\nabla}_{\theta_j \theta_k} \ell (\btheta, \beeta) = \frac{\ell (\btheta^{++}, \beeta) - \ell (\btheta^{+\cdot}, \beeta) - \ell (\btheta^{\cdot+}, \beeta) + \ell (\btheta, \beeta)}{(h_{jk} \sgn{\theta_j})(h_{jk} \sgn{\theta_k})}.
\end{equation*}
When $j \ne k$,  
\begin{align*}
    h_{jk} &= \min \brce{\frac{1}{\sqrt{nT}}, 2 \abs{\theta_j}, 2 \abs{\theta_k}}, \\
    \theta_j^{++} &= \theta_j - h_{jk} \sgn{\theta_j}, \qquad
    \theta_k^{++} = \theta_k - h_{jk} \sgn{\theta_k},
\end{align*}
and the other coordinates $\theta_l^{++}$ for $l \ne 0$ are calculated as in (\ref{equ:num.Hess.jk.theta0}) and (\ref{equ:num.Hess.jk.thetal}).  
Similarly, $\btheta^{+-}, \btheta^{-+}$ can be defined with  
\begin{align*}
    \theta_j^{+\cdot} = \theta_j + h_{jk}, \qquad \theta_k^{+\cdot} = \theta_k, \\
    \theta_j^{\cdot+} = \theta_j, \qquad \theta_k^{\cdot+} = \theta_k + h_{jk}.
\end{align*}
When $j = k$,  
\begin{align*}
    h_{jj} &= \min \brce{\frac{1}{\sqrt{nT}}, \abs{\theta_j}}, \\
    \theta_j^{++} &= \theta_j - 2 h_{jj} \sgn{\theta_j},
\end{align*}
and the other coordinates $\theta_l^{++}$ for $l \ne 0$ are calculated as in (\ref{equ:num.Hess.jj.theta0}) and (\ref{equ:num.Hess.jj.thetal}).  
Similarly, $\btheta^{+\cdot}, \btheta^{\cdot+}$ can be defined with  
$$\theta_j^{+\cdot} = \theta_j^{\cdot+} = \theta_j - h_{jj} \sgn{\theta_j}.$$  
When $\theta_0 = \theta_j = 0$ and $\theta_k \ne 0$, we use the gradient of $\btheta^{\prime}$ for approximation, where $\btheta^{\prime}$ is defined in (\ref{equ:num.grad.thetaj.0}).  
When $\theta_0 = \theta_j = \theta_k = 0$, we use the gradient of $\btheta^{\prime \prime}$ for approximation, where  
\begin{align*}
    \theta_0^{\prime \prime} &= 0, \qquad
    \theta_j^{\prime \prime} = \frac{1}{\sqrt{nT}}, \qquad
    \theta_k^{\prime \prime} = \frac{1}{\sqrt{nT}}, \qquad \\
    \theta_l^{\prime \prime} &= \theta_l \sqrt{1 - (\theta_j^{\prime \prime})^2 - (\theta_k^{\prime \prime})^2} \quad \text{for } l \ne 0, j, k.
\end{align*}

\section{Additional Details in Simulation Study}  

\alert{In this section, we provide the simulation details in Section~\ref{sec:simulation.MSTP}.  

\subsection{MAD and Coverage Probability} \label{sec:simulatio.additional.mad.cp}  

\begin{figure}[!t]
    \centering
    \includegraphics[width=\textwidth]{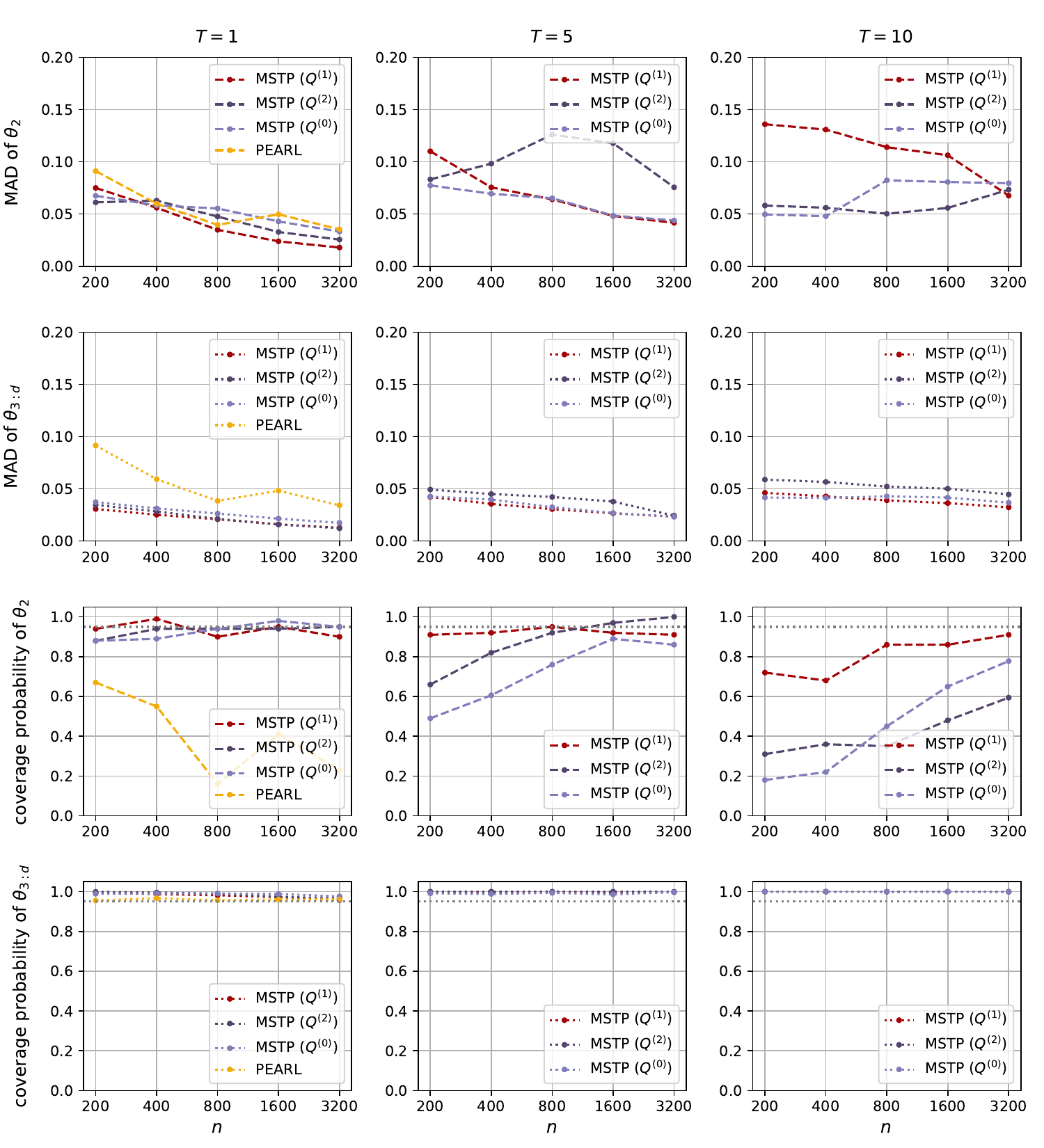}
    \caption{The MAD and coverage probability of $\theta_2$ and their average $\theta_{3:d}$ in Scenario 1. The columns correspond to different values of $T$, and the rows correspond to different metrics. The dotted line indicates the nominal coverage probability.}
    \label{fig:simulation.v1.theta2_other}
\end{figure}

\begin{figure}[!t]
    \centering
    \includegraphics[width=\textwidth]{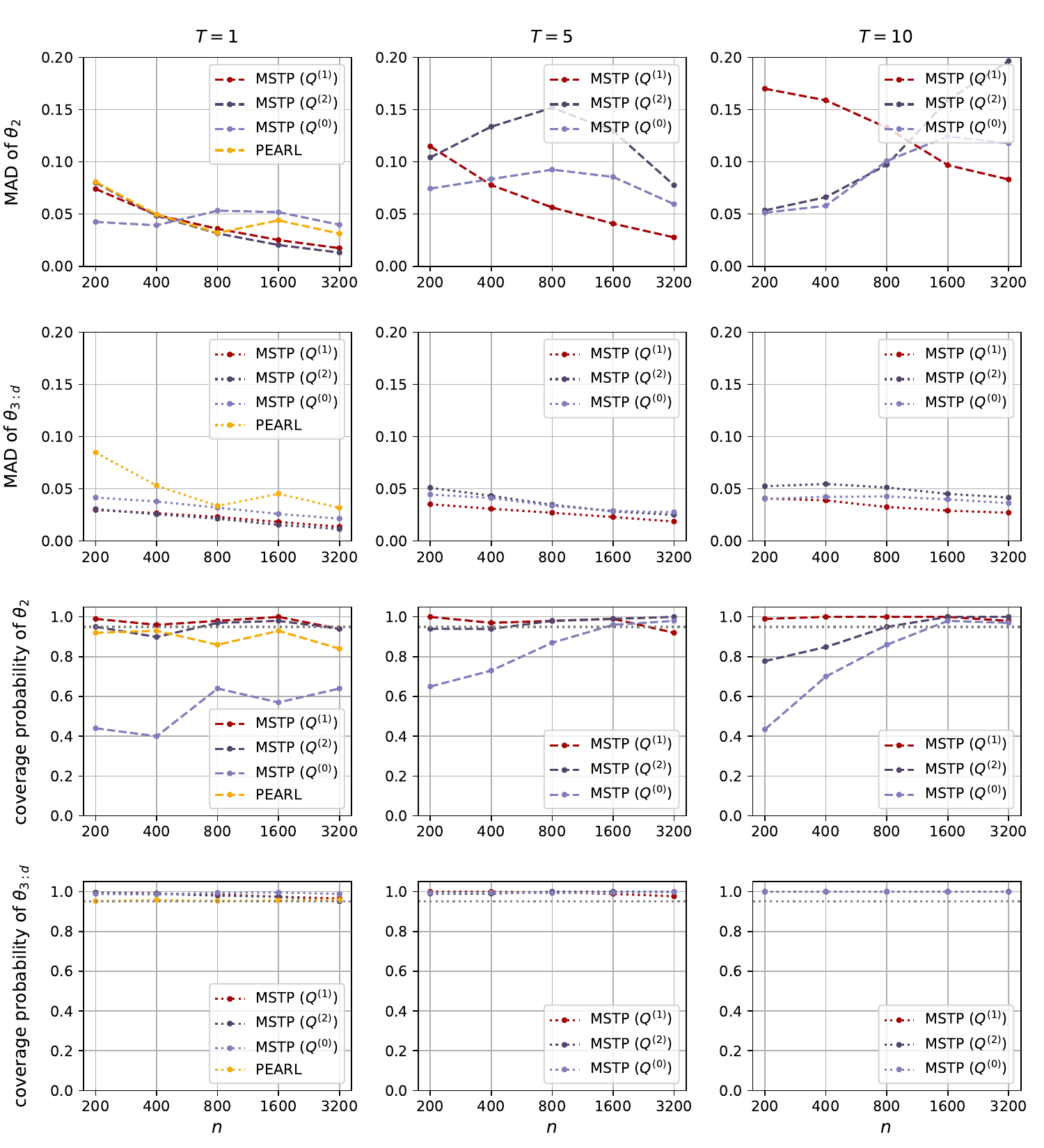}
    \caption{The MAD and coverage probability of $\theta_2$ and their average $\theta_{3:d}$ in Scenario 2. The columns correspond to different values of $T$, and the rows correspond to different metrics. The dotted line indicates the nominal coverage probability.}
    \label{fig:simulation.v2.theta2_other}
\end{figure}

In this section, we present the MAD and the coverage probability of the second important variable $\theta_2$ and their average over the noise variables $\theta_{3:d}$ in Figures~\ref{fig:simulation.v1.theta2_other} and~\ref{fig:simulation.v2.theta2_other}.  
The conclusion is similar to that for $\theta_1$.  
Note that the coverage probability of $\theta_{3:d}$ remains 100\% for all sample sizes when $T = 5$ and $T = 10$.  
This result is due to the bootstrap estimation procedure.  
Each bootstrap estimate $\tilde{\theta}_{b, j}$ of the one-step estimator is obtained by refitting the sparse estimator $\hat{\btheta}_b$ on the bootstrap sample and conducting one-step estimation again.  
Since $\hat{\btheta}_{b, j}$ for $j \in \{3:d\}$ are estimated to be zero in most bootstrap samples, the one-step estimators cover the true value well enough.  
This is not an overestimation since the MAD of $\theta_{3:d}$ is much smaller than the MAD of $\theta_1$ and $\theta_2$.  
For the one-step estimator of PEARL, after rescaling the CIs based on $\theta_1$, $\tilde{\theta}_1$ achieves the nominal coverage probability.  
However, the CI of $\tilde{\theta}_2$ cannot cover $\theta^*$ well in Scenario 1.  
This is because the average ratio $\tilde{\theta}_2 / \tilde{\theta}_1$ for PEARL changes from -0.226 when $n = 200$ to -0.675 when $n = 3200$.

\subsection{Failure of the Sparse Estimator in Statistical Inference}  

In this section, we show that the sparse estimator $\hat{\btheta}$ does not provide valid statistical inference results.  
Specifically, we use bootstrap to construct a CI for each $\hat{\theta}_j$, $j \in \{1:d\}$, where the nuisance parameter is constructed using $Q^{(1)}$.  
Figure~\ref{fig:compare.cp} provides the coverage probability of the CIs for $\hat{\theta}_j$ and $\tilde{\theta}_j$ for $j = 1, 2$, and the average coverage probability of $\hat{\theta}_{3:d}$ and $\tilde{\theta}_{3:d}$ over the noise variables.  
The results suggest that the CI based on $\hat{\theta}_j$ does not achieve the nominal coverage probability, especially when $T = 10$.  
This demonstrates the failure of the sparse estimator in statistical inference and the necessity of using the one-step estimator.

\begin{figure}[!t]
    \centering
    \includegraphics[width=\linewidth]{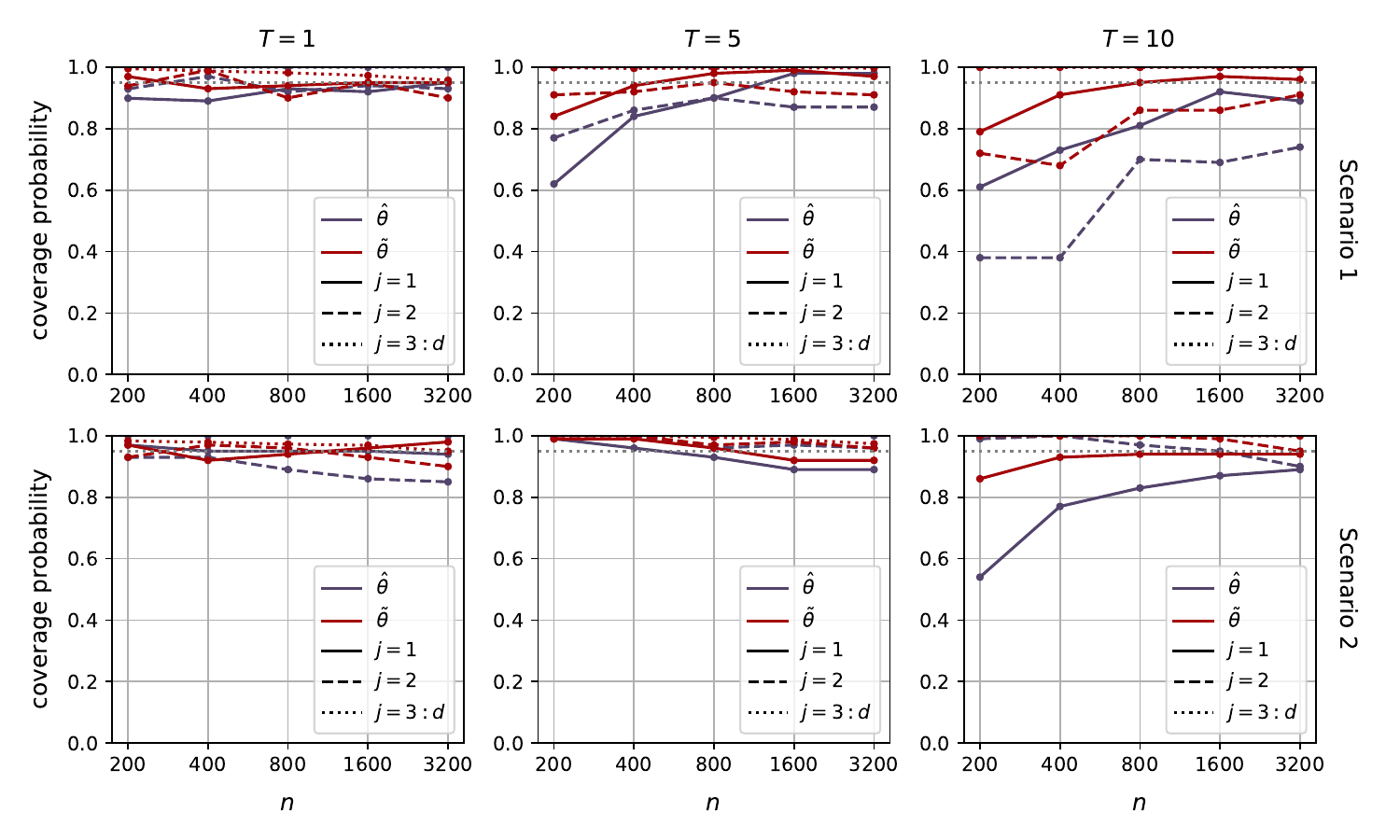}
    \caption{Comparison of the coverage probability of the CIs for the sparse estimator $\hat{\btheta}$ and the one-step estimator $\tilde{\btheta}$. The columns correspond to different values of $T$, and the rows correspond to different simulation scenarios.}
    \label{fig:compare.cp}
\end{figure}

\subsection{Validation of the Estimated MAD}  

In this section, we compare the estimated MAD and the empirical MAD of the MSTP estimated with $Q^{(1)}$.  
The estimated MAD refers to the MAD of the $B$ bootstrap estimators $\tilde{\theta}_{b, j}$, averaged over $W$ replications.  
The empirical MAD refers to the MAD of the $W$ one-step estimators $\tilde{\theta}_j$.  
Figure~\ref{fig:compare.mad} shows that the estimated MAD and the empirical MAD are approximately equal, verifying that the bootstrapped CI is a valid approximation for the CI of $\tilde{\theta}_j$.  
Although the bootstrapped CI might be overly conservative when $T = 10$ and $n$ is small in Scenario 1, the estimated MAD still recovers the empirical MAD as the sample size increases.

\begin{figure}[!t]
    \centering
    \includegraphics[width=\linewidth]{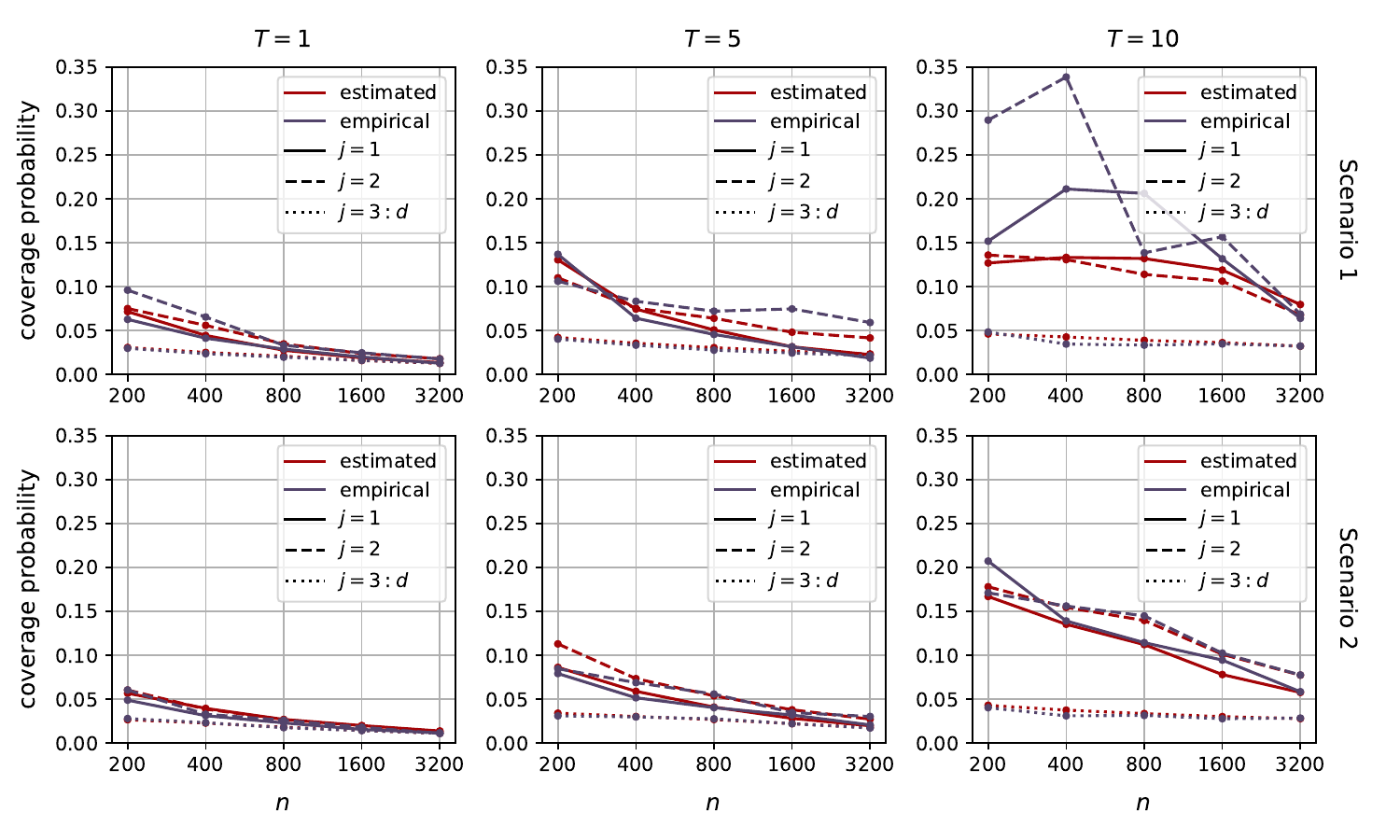}
    \caption{Comparison of the estimated and empirical MAD. The columns correspond to different values of $T$, and the rows correspond to different simulation scenarios.}
    \label{fig:compare.mad}
\end{figure}

\subsection{True Value Function}  \label{sec:true.value.function}  

Tables~\ref{tab:true.optimal.v1} and~\ref{tab:true.optimal.v2} provide the optimal parameter $\btheta^*$ and the optimal average reward $\bar{V}^* = \max_{\btheta} \bar{V} (\btheta)$ in the simulation experiments in Section~\ref{sec:simulation.MSTP}, where $\bar{V} (\btheta) = \frac{1}{T} V (\btheta)$.  
Figures~\ref{fig:true.optimal.v1} and~\ref{fig:true.optimal.v2} present the true average reward for all $\bar{V} (\btheta)$.  
To find $\btheta^*$ within the class $\Pi$, we estimate the average reward $\bar{V} (\btheta)$ on an independent test set of size $200{,}000$ for each $\btheta = [\theta_0, \theta_1, \theta_2]$, where $\theta_1, \theta_2$ take values on the grids inside the unit ball and $\theta_0 = \pm \sqrt{1 - \theta_1^2 - \theta_2^2}$.  
The figures empirically verify that the value function is generally non-concave over the whole parameter space $\normn{\btheta}_2 \le 1$, but it is locally concave around the true optimal parameter $\btheta^*$.

\begin{table}[!ht]
    \centering
    \begin{tabular}{cccc}
        \toprule
        $T$             & 1 & 5 & 10 \\
        \midrule
        $\theta_0^*$    & -0.81 & -0.38 & -0.22 \\
        $\theta_1^*$    &  0.53 &  0.67 &  0.61 \\
        $\theta_2^*$    & -0.26 & -0.64 & -0.76 \\
        $V^*$           & 2.200 & 3.455 & 5.791 \\
        \bottomrule
    \end{tabular}
    \caption{The optimal policy parameter $\btheta^*$ and optimal average reward $\bar{V}^*$ in Scenario 1.}
    \label{tab:true.optimal.v1}
\end{table}

\begin{table}[!ht]
    \centering
    \begin{tabular}{cccc}
        \toprule
        $T$             & 1 & 5 & 10 \\
        \midrule
        $\theta_0^*$    & -0.84 & -0.54 & -0.33 \\
        $\theta_1^*$    &  0.44 &  0.73 &  0.83 \\
        $\theta_2^*$    &  0.31 &  0.42 &  0.45 \\
        $V^*$           & 0.794 & 1.599 & 2.978 \\
        \bottomrule
    \end{tabular}
    \caption{The optimal policy parameter $\btheta^*$ and optimal average reward $\bar{V}^*$ in Scenario 2.}
    \label{tab:true.optimal.v2}
\end{table}

\begin{figure}[!htbp]
    \centering
    \includegraphics[width=\linewidth]{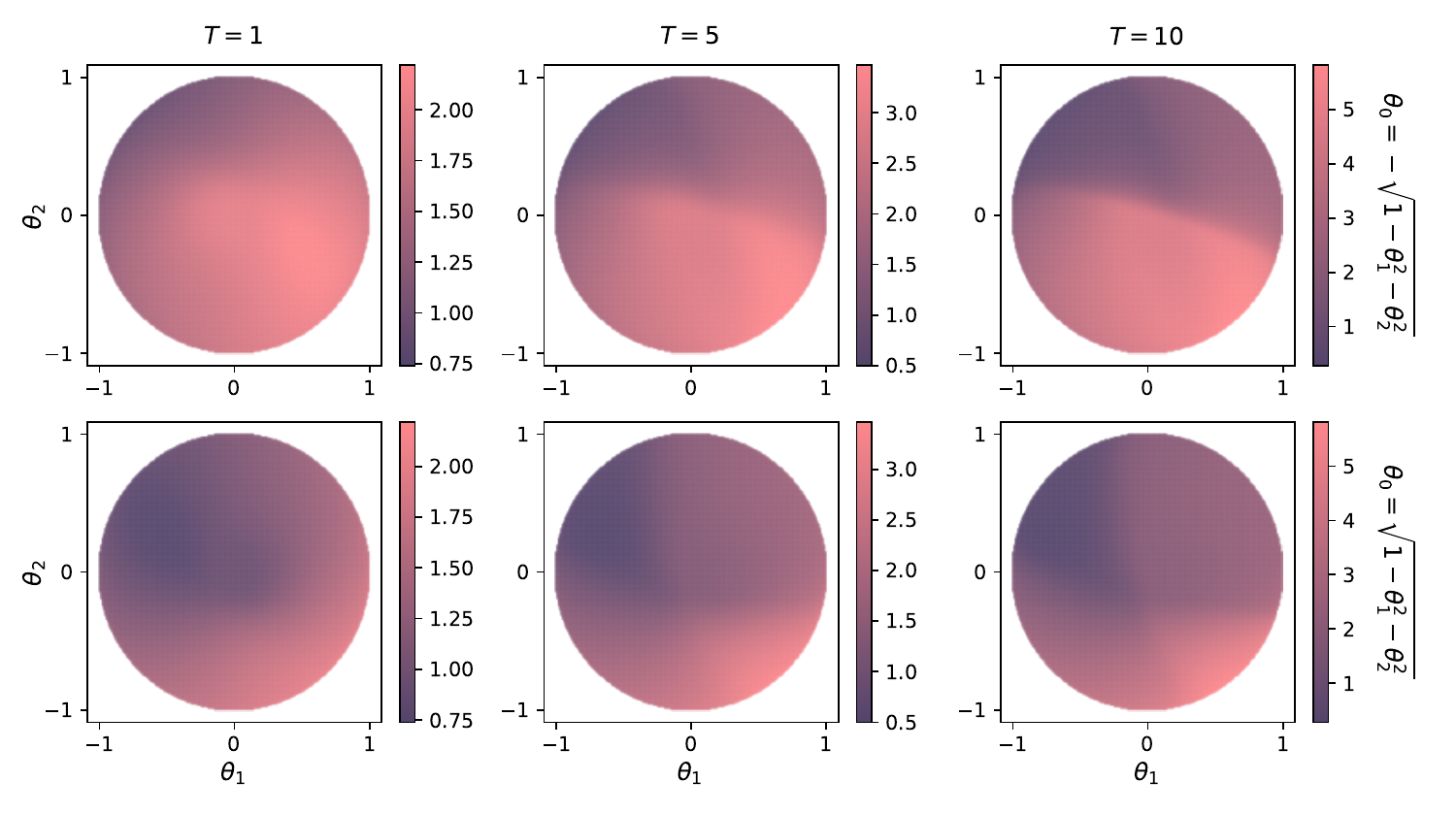}
    \caption{The true value function in Scenario 1. In the first row, $\btheta = [- \sqrt{1 - \theta_1^2 - \theta_2^2}, \theta_1, \theta_2]$; in the second row, $\btheta = [\sqrt{1 - \theta_1^2 - \theta_2^2}, \theta_1, \theta_2]$.}
    \label{fig:true.optimal.v1}
\end{figure}

\begin{figure}[!htbp]
    \centering
    \includegraphics[width=\linewidth]{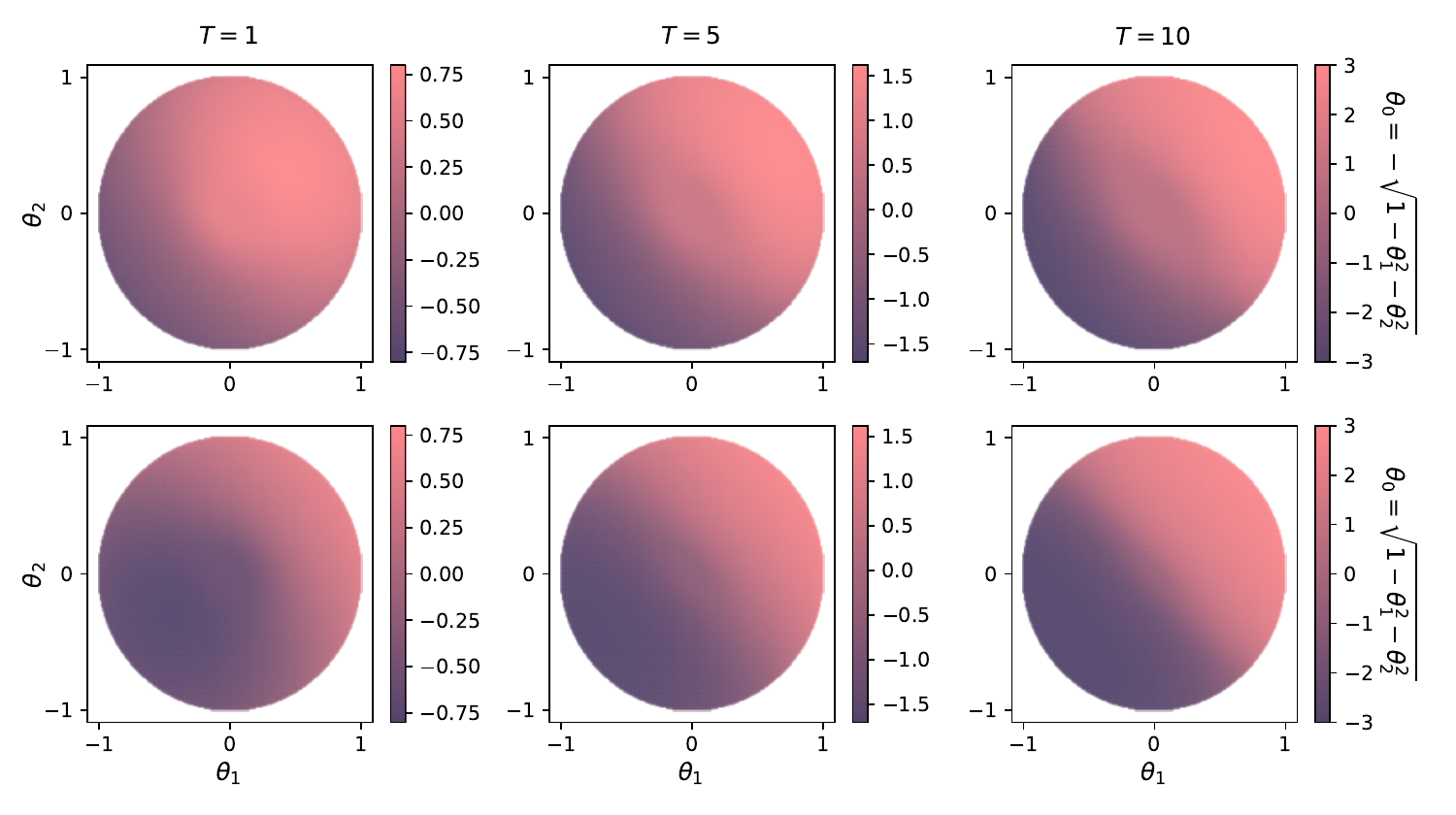}
    \caption{The true value function in Scenario 2. In the first row, $\btheta = [- \sqrt{1 - \theta_1^2 - \theta_2^2}, \theta_1, \theta_2]$; in the second row, $\btheta = [\sqrt{1 - \theta_1^2 - \theta_2^2}, \theta_1, \theta_2]$.}
    \label{fig:true.optimal.v2}
\end{figure}
}

\section{Additional Details in Real Data Analysis} \label{sec:real.data.details}  

\alert{In this section, we present the details of the features, actions, and rewards.  
The latest OhioT1DM dataset contains two cohorts: a 2018 cohort and a 2020 cohort.  
The two cohorts used different fitness bands, which provide access to different physiological data.  
Since the basis heart rate, basis air temperature, basis steps, and sleep quality data are not available for the 2020 cohort, we focus on the 2018 cohort for analysis.  

There are 17 relevant features recorded in the original dataset.  
We use the following 12 features to construct the MSTP.  
For a detailed description of the OhioT1DM dataset, please refer to \citet{marling2020ohiot1dm}.
\begin{itemize}
    \item \textit{Glucose level}. The CGM data, recorded every 5 minutes.  
    \item \textit{Basal}. The rate of continuous delivery of basal insulin. This rate remains constant for several hours until a new rate is set. The temporary basal variable is incorporated into this variable by superseding the normal basal rate.  
    \item \textit{Meal}. The self-reported carbohydrate estimate for meals. To represent the long-term effect of a meal, the carbohydrate estimate is passed along the next hour but discounted at a rate of 0.9. Specifically, if a patient has consumed a meal between the $(t - 1)$th and the $t$th decision time, the feature is defined as $X_{\text{meal}, s} = 0.9^{s-t} X_{\text{meal}, t}$ for $s \in \{t:(t + 11)\}$.  
    \item \textit{Sleep}. The sleep quality at the decision time. If a patient is awake at this time, the variable takes a value of zero. This variable uses only the sleep quality detected by the fitness band to ensure access to real-time data.  
    \item \textit{Work}. The physical exertion during work time.  
    \item \textit{Hypoglycemic episode}. Whether a hypoglycemic event occurred before the decision time.  
    \item \textit{Exercise}. The physical exertion during exercise time.  
    \item \textit{Heart rate}. The heart rate, aggregated every 5 minutes.  
    \item \textit{GSR}. The galvanic skin response (GSR), aggregated every 5 minutes.  
    \item \textit{Skin temperature}. The skin temperature, aggregated every 5 minutes.  
    \item \textit{Air temperature}. The air temperature, aggregated every 5 minutes.  
    \item \textit{Steps}. The step count, aggregated every 5 minutes.  
\end{itemize}
Although some features in the OhioT1DM dataset, including work time and its physical exertion, hypoglycemic episodes, and exercise time and duration, are self-reported after the event, a functional fitness band would be able to assess these features in real time.  
Hence, we still use these variables in estimating the MSTP, assuming that real-time data will be available to apply the estimated policy in future studies.  

The self-reported sleep quality is omitted from the feature set since the automatic measure is already provided by the fitness band.  
The variable \textit{finger stick} is omitted since it represents blood glucose values measured by self-monitoring and conveys similar information to the CGM data.  
The variable \textit{stressor}, which represents the time of self-reported stress, is omitted because there were only 5 self-reported stress events across all 6 users.  
The variable \textit{illness} is omitted because illness typically occurs over a period of time, but the ending time of each illness is missing.  
Additionally, there were only 11 self-reported illness events for the 6 users in total.  

\begin{figure}
    \centering
    \includegraphics[width=\linewidth]{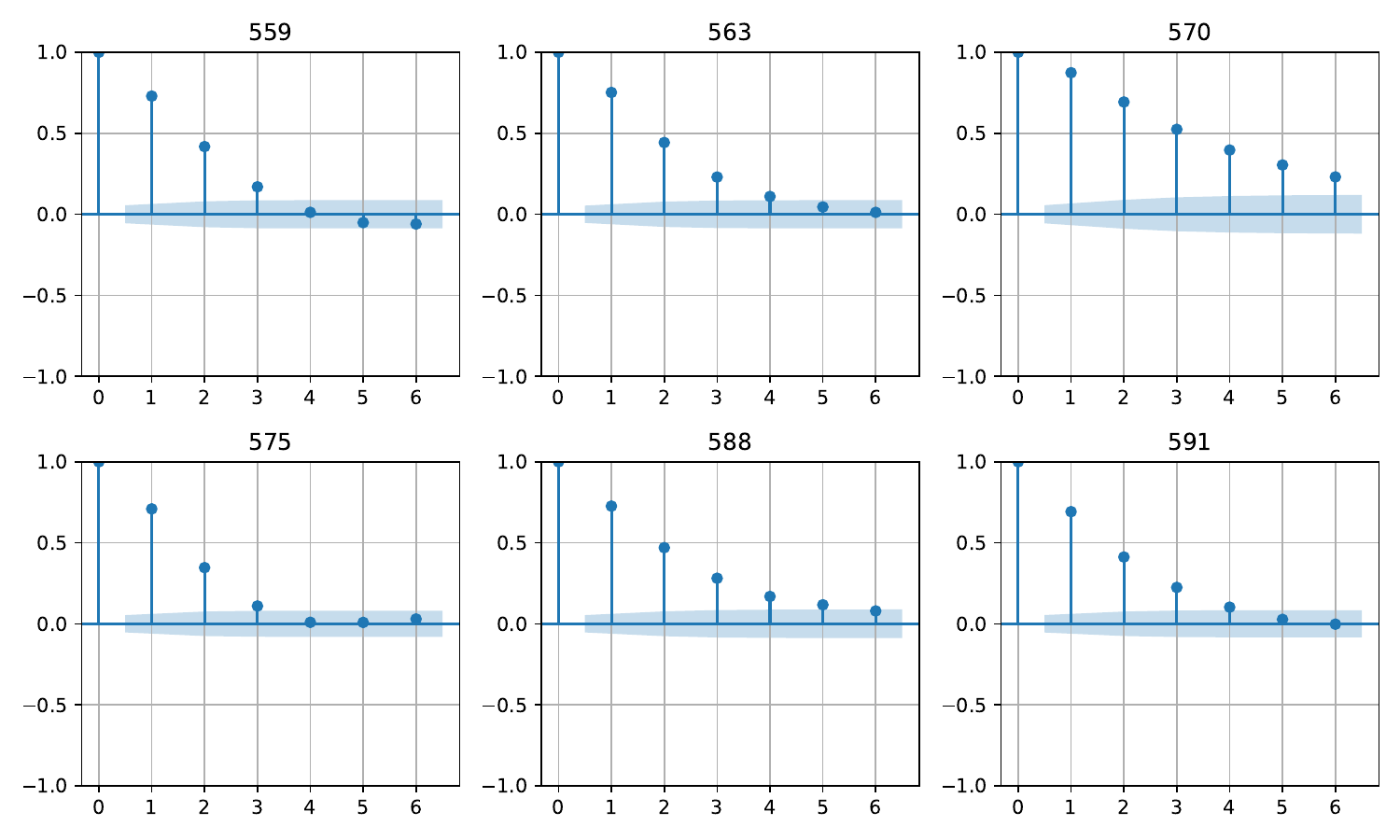}
    \caption{The autocorrelation of the initial glucose level for each hour across the 6 users in the 2018 cohort.}
    \label{fig:acf}
\end{figure}
}

We define the treatment $A_t$ based on whether bolus insulin is delivered or not, regardless of the dosage.
Let $j$ be the coordinate of the glucose level in the feature vector, and let $X_{t, j}$ represent the glucose level at time $t$.  
The reward is defined as the Index of Glycemic Control,  
\begin{equation*}
    R_t = 
    \begin{cases}
        - \frac{1}{30} (80 - X_{t + 1, j})^2, & X_{t + 1, j} < 80, \\
        0, & 80 \le X_{t + 1, j} \le 140, \\
        - \frac{1}{30} (X_{t + 1, j} - 140)^{1.35}, & X_{t + 1, j} \ge 140, \\
    \end{cases}
\end{equation*}
following \citet{shi2020does}.  

Since we assume each trajectory is independent in theory, we separate the 8-week data of each user into multiple episodes.  
Figure~\ref{fig:acf} plots the autocorrelation function of the initial glucose level of each hour.  
We observe that the autocorrelation decreases to zero after 4 hours for most users.  
Therefore, we use the first hour of data from each 4-hour block for the real data analysis, while excluding the remaining 3 hours of data.  
In this way, temporal dependence across treatment stages is allowed within each trajectory, but independence is generally preserved between trajectories.

\section{Theoretical Proofs} \label{sec:proof}



\subsection{Gradient and Hessian Matrix of the Loss Function} \label{sec:proof.gradient}

In this section, we provide the theoretical gradient and Hessian matrix of the loss function.
By the fact that any estimated $\btheta$ will satisfy the equation $\norm{\btheta}_2 = 1$ to approximate a deterministic policy, we can rewrite the intercept as $\theta_0 = \pm \sqrt{1 - \theta_1^2 - \dots - \theta_d^2}$. 
When $\theta_0$ is nonnegative, the gradient and Hessian matrix of the loss function are
\begin{equation} \label{equ:loss.gradient}
\begin{split}
    \nabla_{\bxi} \ell (\btheta, \hat{\beeta}) 
    = - \frac{1}{n} \sum_{i=1}^n \sum_{t=1}^T \Bigg\{ &
    \frac{\nabla_{\bxi} [\prod_{k=1}^t \pi^{\btheta} (A_{i,k} | \bX_{i,k})]}{\prod_{k=1}^t \mu_k (A_{i,k} | \bH_{i,k})} [R_{i,t} - \hat{Q}_t (\bX_{i,t}, A_{i,t})] \\
    & + \frac{\nabla_{\bxi} [\prod_{k=1}^{t-1} \pi^{\btheta} (A_{i,k} | \bX_{i,k})]}{\prod_{k=1}^{t-1} \mu_k (A_{i,k} | \bH_{i,k})} \sum_{a \in \cA} \pi^{\btheta} (a | \bX_{i,t}) \hat{Q}_t (\bX_{i,t}, a), \\
    & + \frac{\prod_{k=1}^{t-1} \pi^{\btheta} (A_{i,k} | \bX_{i,k})}{\prod_{k=1}^{t-1} \mu_k (A_{i,k} | \bH_{i,k})} \sum_{a \in \cA} \nabla_{\bxi} \pi^{\btheta} (a | \bX_{i,t}) \hat{Q}_t (\bX_{i,t}, a) \Bigg\},
\end{split}
\end{equation}
\begin{equation} \label{equ:loss.hessian}
\begin{split}
    \nabla_{\bxi \bxi}^2 \ell (\btheta, \hat{\beeta}) 
    = - \frac{1}{n} \sum_{i=1}^n \sum_{t=1}^T \Bigg\{ &
    \frac{\nabla_{\bxi \bxi}^2 [\prod_{k=1}^t \pi^{\btheta} (A_{i,k} | \bX_{i,k})]}{\prod_{k=1}^t \mu_k (A_{i,k} | \bH_{i,k})} [R_{i,t} - \hat{Q}_t (\bX_{i,t}, A_{i,t})] \\
    & + \frac{\nabla_{\bxi \bxi}^2 [\prod_{k=1}^{t-1} \pi^{\btheta} (A_{i,k} | \bX_{i,k})]}{\prod_{k=1}^{t-1} \mu_k (A_{i,k} | \bH_{i,k})} \sum_{a \in \cA} \pi^{\btheta} (a | \bX_{i,t}) \hat{Q}_t (\bX_{i,t}, a), \\
    & + 2 \frac{\nabla_{\bxi} [\prod_{k=1}^{t-1} \pi^{\btheta} (A_{i,k} | \bX_{i,k})]}{\prod_{k=1}^{t-1} \mu_k (A_{i,k} | \bH_{i,k})} \sum_{a \in \cA} \nabla_{\bxi} \pi^{\btheta} (a | \bX_{i,t}) \hat{Q}_t (\bX_{i,t}, a), \\
    & + \frac{\prod_{k=1}^{t-1} \pi^{\btheta} (A_{i,k} | \bX_{i,k})}{\prod_{k=1}^{t-1} \mu_k (A_{i,k} | \bH_{i,k})} \sum_{a \in \cA} \nabla_{\bxi \bxi}^2 \pi^{\btheta} (a | \bX_{i,t}) \hat{Q}_t (\bX_{i,t}, a) \Bigg\},
\end{split}
\end{equation}
where
\begin{gather*}
    \nabla_{\bxi} \brck{\prod_{k=1}^t \pi^{\btheta} (a_k | \bx_k)} 
    = \brck{\prod_{k=1}^t \pi^{\btheta} (a_k | \bx_k)} \brck{\sum_{k=1}^t \frac{a_k (\bx_k - \bxi / \theta_0) / \tau}{1 + e^{a_k \bx_k^T \btheta / \tau}}}, \\
    \begin{split}
        \nabla_{\bxi \bxi}^2 & \brck{\prod_{k=1}^t \pi^{\btheta} (a_k | \bx_k)}
        = \brck{\prod_{k=1}^t \pi^{\btheta} (a_k | \bx_k)} \brck{\sum_{k=1}^t \frac{a_k (\bx_k - \bxi / \theta_0) / \tau}{1 + e^{a_k \bx_k^T \btheta / \tau}}} \brck{\sum_{k=1}^t \frac{a_k (\bx_k - \bxi / \theta_0) / \tau}{1 + e^{a_k \bx_k^T \btheta / \tau}}}^T \\
        &- \brck{\prod_{k=1}^t \pi^{\btheta} (a_k | \bx_k)} \sum_{k=1}^t \frac{(1 + e^{a_k \bx_k^T \btheta / \tau}) a_k / (\tau \theta_0) \bI + e^{a_k \bx_k^T \btheta / \tau} [a_k (\bx_k - \bxi / \theta_0) / \tau] [a_k (\bx_k - \bxi / \theta_0) / \tau]^T}{(1 + e^{a_k \bx_k^T \btheta / \tau})^2}.
    \end{split}
\end{gather*}
Similarly, we can find the gradient and Hessian matrix when $\theta_0$ is negative.
Note that the nuisance parameter $\hat{\beeta} = \hat{Q}_{1:T}$ is estimated using the initial estimator $\check{\btheta}$ and is then fixed in when minimizing the loss function.
Therefore, the gradient and Hessian matrix with respect to $\bxi$ does not involve the gradient of $\hat{\beeta}$ with respect to $\bxi$.

\subsection{Proof of Theorem~\ref{thm:main}} \label{sec:proof.inference}

Since $R_t$ is bounded for all $t$ by Assumption~\ref{asp:bound.var}, we know that $\bar{\beeta}$ is bounded by $r$ with probability one. 
For a real valued function $f: \cD \mapsto \bbR$, write the empirical process as 
$$\bbG_n = \frac{1}{\sqrt{n}} \sum_{i=1}^n [f(\bD_i) - \bbE f(\bD_i)].$$
Denote $[\cdot]_j$ as the $j$th dimension of a vector and $[\cdot]_{j.}$ the $j$th row of a matrix.

For $\bar{\cH} := \{ \beeta - \bar{\beeta}: \beeta \in \cH \}$ define the pathwise derivative of the nuisance parameter $D_q: \bar{\cH} \mapsto \bbR^{d}$,
\[ D_q [\beeta - \bar{\beeta}] := \nabla_q \bbE \brck{\nabla_{\bxi} \ell (\btheta^*, \bar{\beeta} + q (\beeta - \bar{\beeta}))}, \qquad \beeta \in \cH, \]
for all $q \in [0, 1)$.
This derivative exists by our construction of $\ell$.
We will use the Neyman orthogonality to decorrelate the high-dimensional nuisance parameters.
The definition is taken from \citet{chernozhukov2018double}.

\begin{dfn}[Neyman orthogonality]
    The score $\nabla_{\bxi} \ell$ obeys the orthogonality condition at $(\btheta^*, \bar{\beeta})$ with respect to the nuisance realization set $\cH_n \in \cH$ if
    \[ \bbE [\nabla_{\bxi} \ell (\btheta^*, \bar{\beeta})] = 0 \]
    and the pathwise derivative map $D_q [\beeta - \bar{\beeta}]$ exists for all $q \in [0, 1)$ and $\beeta \in \cH_n$ and vanishes at $q = 0$; namely,
    \[ D_0 [\beeta - \bar{\beeta}] = \bzero, \qquad \text{for all } \beeta \in \cH_n. \]
\end{dfn}

\begin{lem} \label{lem:Neyman.second.rate}
    Under Assumptions~\ref{asp:nuca}-\ref{asp:bound.var}, 
    the gradient $\nabla_{\bxi} \ell$ satisfies the Neyman orthogonality.
    In addition, the following results hold for the nuisance parameters:
    \begin{gather}
        \sup_{\beeta \in \cH_n, q \in (0,1)} \norm{\nabla^2_{q q} \bbE [\nabla_{\bxi} \ell (\btheta^*, \bar{\beeta} + q (\beeta - \bar{\beeta}))]}_{\infty} = 0, \label{equ:derivative.theta.q2} \\
        \sup_{\beeta \in \cH_n, q \in (0,1)} \norm{\nabla_q \bbE [\nabla_{\bxi \bxi} \ell (\btheta^*, \bar{\beeta} + q (\beeta - \bar{\beeta}))]}_{\infty} = 0. \label{equ:derivative.theta2.q} 
    \end{gather}
\end{lem}

\begin{proof}
Since $\bbE [\ell (\btheta, \beeta)] = - V (\btheta)$ for any $\beeta$
and $\nabla_{\bxi} V (\btheta^*) = 0$ by the definition of $\btheta^*$,
we have $\bbE [\nabla_{\bxi} \ell (\btheta^*, \beeta)] = 0$ for any $\beeta$,
when the regularity conditions are satisfied.
Therefore, 
\[ D_0 [\beeta - \bar{\beeta}] = \nabla_{\bQ} \bbE [\nabla_{\bxi} \ell (\btheta^*, \beeta)] |_{\beeta = \bar{\beeta}} \cdot (\beeta - \bar{\beeta}) = \bzero \]
since $\nabla_{\bQ} \bbE [\nabla_{\bxi} \ell (\btheta^*, \beeta)] = \bzero$.

Similarly, for each dimension $j$ we also have $\nabla^2_{\bQ \bQ} \bbE [\nabla_{\theta_j} \ell (\btheta^*, \beeta)] = \bzero$ since $\bbE [\nabla_{\theta_j} \ell (\btheta^*, \beeta)]$ is a constant zero.
Consequently,
\begin{align*}
    & \brck{\nabla^2_{q q} \bbE [\nabla_{\bxi} \ell (\btheta^*, \bar{\beeta} + q (\beeta - \bar{\beeta}))]}_j \\
    = & \nabla^2_{q q} \bbE [\nabla_{\theta_j} \ell (\btheta^*, \bar{\beeta} + q (\beeta - \bar{\beeta}))] \\
    = & (\beeta - \bar{\beeta})^T \nabla^2_{\bQ \bQ} \bbE [\nabla_{\theta_j} \ell (\btheta^*, \bar{\beeta} + q (\beeta - \bar{\beeta}))] (\beeta - \bar{\beeta}) \\
    = & 0
\end{align*}
and (\ref{equ:derivative.theta.q2}) follows.
Since $\bbE [\nabla_{\bxi \bxi} \ell (\btheta^*, \beeta)] = \nabla_{\bxi \bxi} V (\btheta^*)$ for any $\beeta$,
we have that 
$$\nabla_q \bbE [\nabla_{\bxi \bxi} \ell (\btheta^*, \bar{\beeta} + q (\beeta - \bar{\beeta}))] = \nabla_q \nabla_{\bxi \bxi} V (\btheta^*) = 0$$ 
and thus (\ref{equ:derivative.theta2.q}) follows.
\end{proof}

\begin{lem} \label{lem:nuisance.EP}
    Under Assumptions~\ref{asp:nuca}-\ref{asp:conv.rate}, we have
    \begin{gather}
        \sup_{\beeta \in \cH_n} \norm{ \nabla_{\bxi} \ell (\btheta^*, \beeta) - \nabla_{\bxi} \ell (\btheta^*, \bar{\beeta})}_{\infty} = o_{\bbP} (1 / \sqrt{n}), \label{equ:first.order} \\
        \sup_{\beeta \in \cH_n} \norm{ \bv_j^{*T} \nabla_{\bxi \bxi}^2 \ell (\btheta^*, \beeta) - \bv_j^{*T} \nabla_{\bxi \bxi}^2 \ell (\btheta^*, \bar{\beeta})}_{\infty} = o_{\bbP} (\sqrt{\log d / n}). \label{equ:second.order}
    \end{gather}
\end{lem}

\begin{proof}
To show (\ref{equ:first.order}), first note that
\begin{align}
\norm{ \nabla_{\bxi} \ell (\btheta^*, \beeta) - \nabla_{\bxi} \ell (\btheta^*, \bar{\beeta})}_{\infty} 
\leq & \frac{1}{\sqrt{n}} \norm{ \bbG_n [\nabla_{\bxi} l (\btheta^*, \beeta) - \nabla_{\bxi} l (\btheta^*, \bar{\beeta})]}_{\infty} \label{equ:first.order.EP} \\
& + \norm{ \bbE [\nabla_{\bxi} \ell (\btheta^*, \beeta) - \nabla_{\bxi} \ell (\btheta^*, \bar{\beeta})]}_{\infty} \label{equ:first.order.ortho}
\end{align}
for any $\beeta \in \cH_n$.
We will bound (\ref{equ:first.order.ortho}) using Neyman orthogonality and bound (\ref{equ:first.order.EP}) using the results of empirical process.

To bound (\ref{equ:first.order.ortho}),
define 
$$ \bh_{\beeta} (q) := \bbE [\nabla_{\bxi} \ell (\btheta^*, \bar{\beeta} + q (\beeta - \bar{\beeta})) - \nabla_{\bxi} \ell (\btheta^*, \bar{\beeta})] $$
for $q \in [0, 1)$.
By Taylor's expansion, there exists $\tilde{q} \in (0,1)$ such that
$$\bh_{\beeta} (1) = \bh_{\beeta} (0) + \nabla_q \bh_{\beeta} (0) + \frac{1}{2} \nabla^2_{q q} \bh_{\beeta} (\tilde{q}).$$
We have $\bh_{\beeta} (0) = \bzero$ by definition and $\nabla_q \bh_{\beeta} (0) = \bzero$ by Lemma~\ref{lem:Neyman.second.rate}.
In addition, the second derivative satisfies $\normn{\nabla^2_{q q} \bh_{\beeta} (\tilde{q})}_{\infty} = 0$ by Lemma~\ref{lem:Neyman.second.rate}.
Therefore, $\norm{\bh_{\beeta} (1)}_{\infty} = 0$ for any $\beeta \in \cH_n$
and thus
$$\sup_{\beeta \in \cH_n} \norm{ \bbE [\nabla_{\bxi} \ell (\btheta^*, \beeta) - \nabla_{\bxi} \ell (\btheta^*, \bar{\beeta})]}_{\infty} = o (1 / \sqrt{n}).$$

To bound (\ref{equ:first.order.EP}),
define
$$\bg_{\beeta_1, \beeta_2} (q) = \nabla_{\bxi} \ell (\btheta^*, \beeta_1 + q (\beeta_2 - \beeta_1)) - \nabla_{\bxi} \ell (\btheta^*, \beeta_1)$$
for any $\beeta_1, \beeta_2 \in \cH_n$.
Then we have 
$$\bg_{\beeta_1, \beeta_2} (1) = \nabla_{\bxi} \ell (\btheta^*, \beeta_2) - \nabla_{\bxi} \ell (\btheta^*, \beeta_1).$$
By Taylor's expansion, there exists $\tilde{q} \in (0,1)$ such that
\[ \bg_{\beeta_1, \beeta_2} (1) = \bg_{\beeta_1, \beeta_2} (0) + \nabla_q \bg_{\beeta_1, \beeta_2} (\tilde{q}) = \nabla_{\btheta, \beeta}^2 \ell (\btheta^*, \beeta_1 + \tilde{q} (\beeta_2 - \beeta_1)) (\beeta_2 - \beeta_1) \] 
since $\bg_{\beeta_1, \beeta_2} (0) = 0$ by definition.
Hence by Cauchy-Schwartz inequality we have 
\begin{equation} \label{equ:Lipschitz}
    [\nabla_{\bxi} \ell (\btheta^*, \beeta_2) - \nabla_{\bxi} \ell (\btheta^*, \beeta_1)]_j \leq \normn{[\nabla_{\btheta, \beeta}^2 \ell (\btheta^*, \beeta_1 + \tilde{q} (\beeta_2 - \beeta_1))]_{j.}}_2 \norm{\beeta_2 - \beeta_1}_2
\end{equation}
for the $j$th dimension,
which implies that the functions $[\nabla_{\bxi} \ell (\btheta^*, \beeta)]_j$ and $[\nabla_{\bxi} \ell (\btheta^*, \beeta) - \nabla_{\bxi} \ell (\btheta^*, \bar{\beeta})]_j$ are Lipschitz in the parameter $\beeta$.
Note that with Assumption~\ref{asp:bound.var} and the boundedness of $\hat{Q}_t$, we have 
$$\normn{\nabla_{\btheta, \beeta}^2 \ell (\btheta^*, \beeta_1 + \tilde{q} (\beeta_2 - \beeta_1))}_2 \le C,$$ 
where $C$ is a constant.


Therefore, the bracketing number $N_{[]} (\epsilon, \cG_{n,j}, L_2 (\bbP))$ of the function set 
$$\cG_{n,j} := \{ [\nabla_{\bxi} \ell (\btheta^*, \beeta) - \nabla_{\bxi} \ell (\btheta^*, \bar{\beeta})]_j: \beeta \in \cH_n \}$$ 
is upper bounded by the covering number $N (\epsilon / (2C), \cH_n, \norm{\cdot}_2)$ of the nuisance parameter set $\cH_n$ \citep[Theorem 2.7.11]{van1996weak}.
Since $\norm{\beeta - \bar{\beeta}}_2 = o_{\bbP} (1)$, the covering number of the nuisance set $N (\epsilon / (2C), \cH_n, \norm{\cdot}_2)$ is finite.
Let 
$$J_{[]} (\delta, \cG_{n,j}, \norm{\cdot}) := \int_0^{\delta} \sqrt{1 + \log N_{[]} (\epsilon, \cG_{n,j}, \norm{\cdot})} d \epsilon$$ 
be the bracketing integral.
The minimum envelop function of class $\cG_{n,j}$ is defined as $G_{n,j} (x) := \sup_{g \in G_{n,j}} \abs{g(x)}$.
Suppose $\cG_{n,j}$ is covered by the brackets $[l_1, u_1], \dots, [l_{N_{\epsilon}}, u_{N_{\epsilon}}]$, where $N_{\epsilon} := N_{[]} (\epsilon, \cG_{n,j}, L_2 (\bbP))$ is the bracketing number for any $\epsilon > 0$.
Then we can write the minimum envelop function as 
$$G_{n,j} (x) = \max_{j \in \{1:n\}_{\epsilon}} \{ \absn{l_{N_{\epsilon}} (x)}, \absn{u_{N_{\epsilon}} (x)} \}.$$
The $L_{\bbP,2}$ norm of $G_{n,j} (x)$ is then 
$$ \norm{G_{n,j}}_{\bbP, 2}^2 = \bbE G_{n,j}^2 (\bD) \leq \bbE \sum_{j = 1}^{N_{\epsilon}} (l_{j}^2 (\bD) + u_{j}^2 (\bD)).$$
Note that
\begin{align*}
    \bbE l_{j}^2 (\bD) = & \bbE [g (\bD) + (l_j (\bD) - g (\bD))]^2 \\
    \leq & 2 [\bbE g^2 (\bD) + \bbE (l_j (\bD) - g (\bD))^2] \\
    \leq & 2[\bbE g^2 (\bD) + \epsilon^2]
\end{align*}
for some $g \in \cG_{n,j}$ contained in the $j$th bracket for any $j \in \{1:N_{\epsilon}\}$.
According to (\ref{equ:Lipschitz}), 
$$\bbE g^2 (\bD) \leq C^2 \normn{\beeta - \bar{\beeta}}_{\bbP, 2} \le \delta_n $$
by Assumption~\ref{asp:conv.rate}
and thus 
$$\bbE l_{j}^2 (\bD) \le 2 [\epsilon^2 + \delta_n ].$$
The same holds for upper brackets $u_j$.
Since 
$$N_{\epsilon} \le N \prth{\frac{\epsilon}{2C}, \cH_n, \norm{\cdot}_2} \leq \delta_n  \frac{2C}{\epsilon},$$ 
we get that
\begin{equation*}
    \norm{G_{n,j}}_{\bbP, 2}^2 \leq 4 [\epsilon^2 + \delta_n ] N_{\epsilon} 
    \leq 4 [\epsilon^2 + \delta_n ] \cdot 2C \delta_n  \frac{1}{\epsilon}.
\end{equation*}
So $\norm{G_{n,j}}_{\bbP, 2}^2 \to 0$ when $n \to \infty$.
Since 
\begin{equation} \label{equ:Esup.EP}
    \bbE^* \sup_{g \in \cG_{n,j}} \bbG_n (g) \lesssim J_{[]} (\norm{G_{n,j}}_{\bbP, 2}, \cG_{n,j}, L_2 (\bbP))
\end{equation}
by \citet[Theorem 2.14.2]{van1996weak}, the left-hand side of (\ref{equ:Esup.EP}) is in the order of $o(1)$.
Finally, by Markov's inequality, 
$$\sup_{\beeta \in \cH_n} \frac{1}{\sqrt{n}} \bbG_n [\nabla_{\bxi} \ell (\btheta^*, \beeta) - \nabla_{\bxi} \ell (\btheta^*, \bar{\beeta})]_j = o_{\bbP} (1/\sqrt{n}).$$
The bound on (\ref{equ:first.order.EP}) follows by taking the maximum over all the dimensions.

Combining the upper bounds of (\ref{equ:first.order.ortho}) and (\ref{equ:first.order.EP}) we can show (\ref{equ:first.order}).

For (\ref{equ:second.order}), note that
\begin{align}
& \norm{ \bv_j^{*T} \nabla_{\bxi \bxi}^2 \ell (\btheta^*, \beeta) - \bv_j^{*T} \nabla_{\bxi \bxi}^2 \ell (\btheta^*, \bar{\beeta})}_{\infty} \\
\leq & \frac{1}{\sqrt{n}} \norm{ \bbG_n [\bv_j^{*T} \nabla_{\bxi \bxi}^2 \ell (\btheta^*, \beeta) - \bv_j^{*T} \nabla_{\bxi \bxi}^2 \ell (\btheta^*, \bar{\beeta})]}_{\infty} \label{equ:second.order.EP} \\
& + \norm{ \bbE [\bv_j^{*T} \nabla_{\bxi \bxi}^2 \ell (\btheta^*, \beeta) - \bv_j^{*T} \nabla_{\bxi \bxi}^2 \ell (\btheta^*, \bar{\beeta})]}_{\infty} \label{equ:second.order.ortho}
\end{align}
for any $\beeta \in \cH_n$.

To bound (\ref{equ:second.order.ortho}),
define
$$ \tilde{\bh}_{\beeta} (q) := \bbE [\bv_j^{*T} \nabla_{\bxi \bxi}^2 \ell (\btheta^*, \bar{\beeta} + q (\beeta - \bar{\beeta})) - \bv_j^{*T} \nabla_{\bxi \bxi}^2 \ell (\btheta^*, \bar{\beeta})] $$
for $q \in [0, 1)$.
By Taylor's expansion, there exists $\tilde{q} \in (0,1)$ such that
$$\tilde{\bh}_{\beeta} (1) = \tilde{\bh}_{\beeta} (0) + \nabla_q \tilde{\bh}_{\beeta} (\tilde{q}).$$
We have $\tilde{\bh}_{\beeta} (0) = \bzero$ by definition and $\nabla_q \tilde{\bh}_{\beeta} (\tilde{q}) = o (\sqrt{\log d / n})$ by Lemma~\ref{lem:Neyman.second.rate}.

Using similar arguments as that for (\ref{equ:first.order.EP}), we can conclude that (\ref{equ:second.order.EP}) is in the order of $o_{\bbP} (1/\sqrt{n})$. 
Combining the upper bounds of (\ref{equ:second.order.ortho}) and (\ref{equ:second.order.EP}) gives the results.
\end{proof}

\begin{lem}[Concentration of the gradient and Hessian] \label{lem:concen.gradHess}
    Under Assumptions~\ref{asp:nuca}-\ref{asp:conv.rate}, we have
    \begin{gather}
        \norm{\nabla_{\bxi} \ell (\btheta^*, \hat{\beeta}) - \bbE \nabla_{\bxi} \ell (\btheta^*, \bar{\beeta})}_{\infty} = \cO_{\bbP} (\sqrt{\log d / n}), \label{equ:concen.grad} \\
        \norm{\bv_j^{*T} \nabla_{\bxi \bxi}^2 \ell (\btheta^*, \hat{\beeta}) - \bbE (\bv_j^{*T} \nabla_{\bxi \bxi}^2 \ell (\btheta^*, \bar{\beeta}))}_{\infty} = \cO_{\bbP} (\sqrt{\log d / n}).  \label{equ:concen.hess}
    \end{gather}
\end{lem}

\begin{proof}
Since $\bbP (\hat{\beeta} \notin \cH_n) \le \Delta_n$ and $\Delta_n$ converges to zero, we can focus on the event when $\beeta \in \cH_n$.
To prove (\ref{equ:concen.grad}), note that
\begin{equation*}
\begin{split}
    & \norm{\nabla_{\bxi} \ell (\btheta^*, \hat{\beeta}) - \bbE \nabla_{\bxi} \ell (\btheta^*, \bar{\beeta})}_{\infty} \\
    \leq & \norm{ \nabla_{\bxi} \ell (\btheta^*, \hat{\beeta}) - \nabla_{\bxi} \ell (\btheta^*, \bar{\beeta})}_{\infty} + \norm{\nabla_{\bxi} \ell (\btheta^*, \bar{\beeta}) - \bbE \nabla_{\bxi} \ell (\btheta^*, \bar{\beeta})}_{\infty}.
\end{split}
\end{equation*}
By (\ref{equ:first.order}) in Lemma~\ref{lem:nuisance.EP}, we only need to show that 
\begin{equation} \label{equ:concen.grad.tmp}
    \norm{\nabla_{\bxi} \ell (\btheta^*, \bar{\beeta}) - \bbE \nabla_{\bxi} \ell (\btheta^*, \bar{\beeta})}_{\infty} = \cO_{\bbP} (\sqrt{\log d / n}).
\end{equation}
Write 
$$\bh (\bD_{1:n}) := - \nabla_{\bxi} \ell (\btheta^*, \bar{\beeta}),$$
which is a $d$-dimensional real-valued function, and denote the $j$-th dimension of $\bh$ as $h_j$.
According to (\ref{equ:loss.gradient}), we have
\begin{equation*}
    \begin{split}
        \bh (\bD_{1:n}) 
        = \frac{1}{n} \sum_{i=1}^n \sum_{t=1}^T \Big\{ & \nabla_{\bxi} \rho_{i, 1:t}^{\btheta^*, \bar{\bmu}} [R_{i,t} - \bar{\eta}_t (\bX_{i,t}, A_{i,t})] + [\nabla_{\bxi} \rho_{i, 1:(t-1)}^{\btheta^*, \bar{\bmu}}] \bar{U}_t (\bX_{i,t}) \\
        & + \rho_{i, 1:(t-1)}^{\btheta^*, \bar{\bmu}} [\nabla_{\bxi} \bar{U}_t (\bX_{i,t})] \Big\},
    \end{split}
\end{equation*}
where $\bar{U}_t (\bX_{i,t}) = \sum_{a \in \cA} \pi^{\btheta} (a, \bX_{i, t}) \bar{\eta}_t (\bX_{i,t}, a)$.
When $\theta_0^* \ge 0$, since 
\[ \rho_{i, 1:t}^{\btheta^*, \bar{\bmu}} \leq \frac{1}{p_0^t} \qquad \text{and} \qquad \abs{\frac{A_{i,k} (X_{i,k,j} - \theta_j^* / \theta_0^*) / \tau}{1 + e^{A_{i,k} \bX_{i,k}^T \btheta^* / \tau}}} \leq \frac{z - u^{-1}}{\tau}, \]
we have that the $j$th dimension of $\nabla_{\bxi} \rho_{i, 1:t}^{\btheta^*, \bar{\bmu}}$ is upper bounded by $\frac{zt}{\tau p_0^t}$.
Besides, we know that $|R_{i,t} - \bar{\eta}_t (\bX_{i,t}, A_{i,t})| \leq 2r$ since $\bar{\eta}_t$ is bounded by $r$ according to Assumption~\ref{asp:bound.var} and the definition of $\bar{\eta}_t$.
Thus the $j$th dimension of the first part is changed by
\begin{equation*}
    \begin{split}
    \frac{1}{n} \sum_{t=1}^T \abs{ [\nabla_{\bxi} \rho_{i, 1:t}^{\btheta^*, \bar{\bmu}}]_j [R_{i,t} - \bar{\eta}_t (\bX_{i,t}, A_{i,t})] - [\nabla_{\bxi} \rho_{i, 1:t}^{\prime \btheta^*, \bar{\bmu}}]_j [R_{i,t}^{\prime} - \bar{\eta}_t (\bX_{i,t}^{\prime}, A_{i,t}^{\prime})]} \qquad \\
    \leq \frac{2}{n} \sum_{t=1}^T \frac{2rt (z - u^{-1})}{\tau p_0^t}
    \end{split}
\end{equation*}
if the $i$th trajectory $\bD_i$ is changed into $\bD_i^{\prime}$.
Similarly, since $|[\nabla_{\bxi} \pi^{\btheta^*} (a | \bx)]_j| \leq \frac{z - u^{-1}}{\tau}$, 
we have 
\[ |[\nabla_{\bxi} \rho_{i, 1:(t-1)}^{\btheta^*, \bar{\bmu}}]_j \bar{U}_t (\bX_{i,t})| \le \frac{r (t-1) (z - u^{-1})}{\tau p_0^{t-1}} \]
and
\[ |\rho_{i, 1:(t-1)}^{\btheta^*, \bar{\bmu}} [\nabla_{\bxi} \bar{U}_t (\bX_{i,t})]_j| \le \frac{2r (z - u^{-1})}{\tau p_0}. \]
Therefore, the upper bound on the change of $h_j$ when changing $\bD_i$ is
$ c_i := C / n $
for some constant $C > 0$ depending on $r, z, p_0, \tau, T$.
We have similar results when $\theta_0^* < 0$.
By McDiarmid's inequality, we get that 
\[ \bbP \prth{\abs{h_j (\bD_{1:n}) - \bbE h_j (\bD_{1:n})} \geq \epsilon} \leq 2 \exp \brce{- \frac{2 \epsilon^2}{\sum_{i=1}^n c_i^2}} \leq 2 \exp \brce{- C^{\prime} n \epsilon^2} \]
for some constant $C^{\prime} > 0$.
Now the union bound inequality yields 
\[ \bbP \prth{\norm{\bh (\bD_{1:n}) - \bbE \bh (\bD_{1:n})}_{\infty} \geq \epsilon} \leq 2 d \exp \brce{- C^{\prime} n \epsilon^2}. \]
With $\epsilon = C^{\prime \prime} \sqrt{\log d / n}$ for some $C^{\prime \prime} > 0$, we have $\bh (\bD_{1:n}) = \cO_{\bbP} (\sqrt{\log d / n})$ and equation (\ref{equ:concen.grad.tmp}) follows.

Similarly, by (\ref{equ:second.order}) in Lemma~\ref{lem:nuisance.EP}, we only need to prove 
\[ \norm{\bv_j^{*T} \nabla_{\bxi \bxi}^2 \ell (\btheta^*, \bar{\beeta}) - \bbE (\bv_j^{*T} \nabla_{\bxi \bxi}^2 \ell (\btheta^*, \bar{\beeta}))}_{\infty} = \cO_{\bbP} (\sqrt{\log d / n}) \]
for (\ref{equ:concen.hess}).
Now $\bv_j^{*T} \nabla_{\bxi \bxi}^2 \ell (\btheta^*, \bar{\beeta})$ can be divided into 5 parts.
When $\bar{\eta}_t$ is upper bounded by the constant $r$, each dimension for each part is bounded by $C^{\prime \prime} / n$ for some constant $C^{\prime \prime} > 0$, since $v^*$ is a constant.
The result follows from the same arguments as before.
\end{proof}

\begin{lem}[Central limit theorem for the score function] \label{lem:CLT}
    Under Assumptions~\ref{asp:nuca}-\ref{asp:positivity.MSTP} and~\ref{asp:Sigma.pd}, it holds that
    \[ \sqrt{n} \bv_j^{*T} \nabla_{\bxi} \ell(\btheta^*, \hat{\beeta}) \Rightarrow N(0, \sigma_j^{*}), \]
    where $\sigma_j^{*} \geq C$ for some constant $C > 0$.
\end{lem}

\begin{proof}
Since $\bbE \nabla_{\bxi} \ell(\btheta^*, \bar{\beeta}) = 0$, note that 
\begin{equation} \label{equ:CLT.decompose}
    \sqrt{n} \nabla_{\bxi} \ell(\btheta^*, \hat{\beeta}) 
    = \sqrt{n} [\nabla_{\bxi} \ell(\btheta^*, \hat{\beeta}) - \nabla_{\bxi} \ell(\btheta^*, \bar{\beeta})]
    + \sqrt{n} [\nabla_{\bxi} \ell(\btheta^*, \bar{\beeta}) - \bbE \nabla_{\bxi} \ell(\btheta^*, \bar{\beeta})].
\end{equation}
The equation (\ref{equ:first.order}) in Lemma~\ref{lem:nuisance.EP} shows that the first difference on the right-hand side of (\ref{equ:CLT.decompose}) is in the order of $o_{\bbP} (1)$ when $\hat{\beeta} \in \cH_n$.
Besides, the probability of $\hat{\beeta} \notin \cH_n$ converges to zero.
For the second difference in (\ref{equ:CLT.decompose}), when $\bSigma^*$ is finite by Assumption~\ref{asp:Sigma.pd}, the multivariate central limit theorem \citep[Theorem 5]{ferguson2017course} shows that
\[ \sqrt{n} [\nabla_{\bxi} \ell(\btheta^*, \bar{\beeta}) - \bbE \nabla_{\bxi} \ell(\btheta^*, \bar{\beeta})] \Rightarrow N(0, \bSigma^*). \]
The convergence follows since $N(0, \bv_j^{*T} \bSigma^* \bv_j^{*}) + o_{\bbP} (1) = N(0, \bv_j^{*T} \bSigma^* \bv_j^{*})$.
In addition, Assumption~\ref{asp:Sigma.pd} guarantees that $\bv_j^{*T} \bSigma^* \bv_j^{*} \geq C$, since $\bv_j^{*T}$ is nonzero at least in its first argument.
\end{proof}

\begin{lem} \label{lem:theta.conv2}
    Under Assumptions~\ref{asp:nuca}-\ref{asp:V.sc},
    when $\lambda_{\btheta} \simeq \sqrt{\log d / n}$,
    we have
    \begin{equation}
        (\hat{\bxi} - \bxi^*)^T \nabla_{\bxi \bxi}^2 \ell (\btheta^*, \hat{\beeta}) (\hat{\bxi} - \bxi^*) = \cO_{\bbP} (s_{\bxi} \log d / n). \label{equ:l1conv.thetaHess} 
    \end{equation}
\end{lem}

Here, we present the proof for Lemmas~\ref{lem:theta.conv} and~\ref{lem:theta.conv2} together.

\begin{proof}
    Denote $\hat{\Delta}_{\bxi} := \hat{\bxi} - \bxi^*$.
    Let $\cS_{\bxi}$ and $\cS_{\bw_j}$ be the support set of $\bxi^*$ and $\bw_j^*$ respectively.
    According to \citet[inequality (88)]{chernozhukov2018plug}, we have
    \begin{equation} \label{equ:BregDist.singlebound}
    \begin{split}
    \agl{\nabla_{\bxi} \ell (\hat{\btheta}, \hat{\beeta}), \hat{\Delta}_{\bxi}} 
    \le & \lambda_{\bxi} (\normn{\bxi^*}_1 - \normn{\hat{\bxi}}_1)
    = \lambda_{\bxi} (\normn{\bxi^*_{\cS_{\bxi}}}_1 - \normn{\bxi^*_{\cS_{\bxi}} + \hat{\Delta}_{\bxi, S_j}}_1 - \normn{\hat{\Delta}_{\bxi, {\cS_{\bxi}}^c}}_1) \\
    \le & \lambda_{\bxi} (\normn{\hat{\Delta}_{\bxi, {\cS_{\bxi}}}}_1 - \normn{\hat{\Delta}_{\bxi, {\cS_{\bxi}}^c}}_1) .
    \end{split}
    \end{equation}
    Define the empirical symmetric Bregman distance as
    \[ H (\bxi, \bxi^*, \beeta) = \agl{\nabla_{\bxi} \ell (\btheta, \beeta) - \nabla_{\bxi} \ell (\btheta^*, \beeta), \bxi - \bxi^*} .\]
    Since all variables are bounded, by Taylor's expansion on each dimension we have
    \[ \sup_{\bxi} \norm{\nabla_{\bxi \bxi} \ell (\btheta, \beeta) - \nabla_{\bxi \bxi} \ell (\btheta, \beeta^{\prime})}_{\infty} \leq L \norm{\beeta - \beeta^{\prime}}_2 \]
    for any $\beeta, \beeta^{\prime} \in \cH_n$.
    Therefore, the empirical symmetric Bregman distance is Lipschitz in the nuisance parameters:
    \begin{equation} \label{equ:BregDist.Lip}
    \abs{H (\bxi, \bxi^*, \beeta) - H (\bxi, \bxi^*, \beeta^{\prime})} \le L \norm{\beeta - \beeta^{\prime}}_2 \normn{\bxi - \bxi^*}_1^2
    \end{equation}
    for any $\beeta, \beeta^{\prime} \in \cH_n$ and for any $\bxi$ \citep[Lemma 3]{chernozhukov2018plug}.

    By Assumption~\ref{asp:V.sc}, we know $\bbE \ell (\btheta, \bar{\beeta}) = - V (\bxi)$ is $\kappa$-strongly convex at $\bxi^*$.
    We will use \citet[Lemma 2]{chernozhukov2018plug} to prove the restricted strong convexity:
    \begin{equation}  \label{equ:BregDist.RSC}
        H (\bxi^* + \Delta_{\bxi}, \bxi^*, \bar{\beeta}) \ge \kappa \normn{\Delta_{\bxi}}_2^2 - \tau_{n, \delta} \normn{\Delta_{\bxi}}_1^2
    \end{equation}
    holds with probability $1 - \delta$ for all $\Delta_{\bxi} \in \bbB (R)$ and $\tau_{n, \delta} \simeq 1 / (\delta \sqrt{n})$.
    To verify the condition, we need to show that 
    \[ \sup_{\bxi \in \{ \bxi^* + \Delta_{\bxi}: \Delta_{\bxi} \in \bbB (R)\}} \norm{\bv_j^{*T} \nabla_{\bxi \bxi} \ell (\btheta, \bar{\beeta}) - \bv_j^{*T} \bbE \nabla_{\bxi \bxi} \ell (\btheta, \bar{\beeta})}_{\infty} \le \tau_{n, \delta} \]
    with probability $1 - \delta$.
    Similar as the bound for (\ref{equ:first.order.EP}), for each dimension $j$, we can use Taylor's expansion to show that $\nabla_{\bxi \bxi} \ell (\btheta, \bar{\beeta})$ is Lipschitz in the parameter $\bxi$ with constant $C$.
    Consequently, the bracketing number $N_{[]} (\epsilon, \cG_{n,j}, L_2 (\bbP))$ of 
    $$\cG_{n,j} := \{ [\bv_j^{*T} \nabla_{\bxi \bxi} \ell (\btheta^* + \Delta_{\bxi}, \bar{\beeta})]_j: \Delta_{\bxi} \in \bbB (R) \}$$ 
    is upper bounded by the covering number $N (\epsilon / (2C), \bbB (R), \norm{\cdot}_2)$.
    For a finite $R$, the set $\bbB (R)$ is bounded and $N (\epsilon, \bbB (R), \norm{\cdot}_2) \simeq (1/\epsilon)^d$.
    Therefore, 
    $$\bbE^* \sup_{g \in \cG_{n,j}} \bbG_n (g) \lesssim J_{[]} (\norm{G_{n,j}}_{\bbP, 2}, \cG_{n,j}, L_2 (\bbP)) \le J_{[]} (\infty, \cG_{n,j}, L_2 (\bbP)) < \infty$$ 
    by \citet[Theorem 2.14.2]{van1996weak}.
    By Markov's inequality and union bound inequality, we get that 
    $$\tau_{n, \delta} \simeq d J_{[]} (\norm{G_{n,j}}_{\bbP, 2}, \cG_{n,j}, L_2 (\bbP)) / (\delta \sqrt{n}).$$

    Combining the Lipschitz bound (\ref{equ:BregDist.Lip}) and the restricted strong convexity bound (\ref{equ:BregDist.RSC}), we have
    \begin{equation*}
    \begin{split}
        H (\hat{\bxi}, \bxi^*, \hat{\beeta})
        \ge & H (\hat{\bxi}, \bxi^*, \bar{\beeta}) - L \normn{\hat{\beeta} - \bar{\beeta}}_2 \normn{\hat{\Delta}_{\bxi}}_1^2 \\
        \ge & \kappa \normn{\hat{\Delta}_{\bxi}}_2^2 - (\tau_{n, \delta} + L \normn{\hat{\beeta} - \bar{\beeta}}_2) \normn{\hat{\Delta}_{\bxi}}_1^2.
    \end{split}
    \end{equation*}
    On the other hand,
    \begin{equation*}
    \begin{split}
        H (\hat{\bxi}, \bxi^*, \hat{\beeta})
        \le & \agl{\nabla_{\bxi} \ell (\hat{\btheta}, \hat{\beeta}) - \nabla_{\bxi} \ell (\btheta^*, \hat{\beeta}), \hat{\bxi} - \bxi^*} \\
        \le & \lambda_{\bxi} (\normn{\hat{\Delta}_{\bxi, {\cS_{\bxi}}}}_1 - \normn{\hat{\Delta}_{\bxi, {\cS_{\bxi}}^c}}_1) + \norm{\nabla_{\bxi} \ell (\btheta^*, \hat{\beeta})}_{\infty} \normn{\hat{\Delta}_{\bxi}}_1
    \end{split}
    \end{equation*}
    according to (\ref{equ:BregDist.singlebound}).
    The assumption that $\lambda_{\bxi} \simeq \sqrt{\log d / n}$ and Lemma~\ref{lem:concen.gradHess} implies 
    $$\norm{\nabla_{\bxi} \ell (\btheta^*, \hat{\beeta})}_{\infty} \lesssim \lambda_{\bxi} / 2$$ 
    with high probability.
    Conditioning on this event, the above two bounds yield
    \begin{equation*}
    \begin{split}
    \kappa \normn{\hat{\Delta}_{\bxi}}_2^2 
    \le & \frac{\lambda_{\bxi}}{2} (3 \normn{\hat{\Delta}_{\bxi, {\cS_{\bxi}}}}_1 - \normn{\hat{\Delta}_{\bxi, {\cS_{\bxi}}^c}}_1) + (\tau_{n, \delta} + L \normn{\hat{\beeta} - \bar{\beeta}}_2) \normn{\hat{\Delta}_{\bxi}}_1^2 \\
    \le & \frac{3 \lambda_{\bxi}}{2} \sqrt{s_{\bxi}} \normn{\hat{\Delta}_{\bxi}}_2 + (\tau_{n, \delta} + L \normn{\hat{\beeta} - \bar{\beeta}}_2) s_{\bxi} \normn{\hat{\Delta}_{\bxi}}_2^2.
    \end{split}
    \end{equation*}
    Since $\bbP (\beeta \notin \cH_n) \to 0$, we focus on the events when $\beeta \in \cH_n$.
    When $n$ is large enough such that $(\tau_{n, \delta} + L \delta_n) s_{\bxi} \le \kappa / 2$, we have 
    $$\normn{\hat{\Delta}_{\bxi}}_2 \le \frac{3 \lambda_{\bxi}}{4} \sqrt{s_{\bxi}} \simeq \sqrt{s_{\bxi} \log d / n}.$$
    Besides, since 
    $$\frac{\lambda_{\bxi}}{2} (3 \normn{\hat{\Delta}_{\bxi, {\cS_{\bxi}}}}_1 - \normn{\hat{\Delta}_{\bxi, {\cS_{\bxi}}^c}}_1) \ge \frac{\kappa}{2} \normn{\hat{\Delta}_{\bxi}}_2^2 \ge 0,$$
    we also have $\normn{\hat{\Delta}_{\bxi, {\cS_{\bxi}}^c}}_1 \le 3 \normn{\hat{\Delta}_{\bxi, {\cS_{\bxi}}}}_1$
    and thus 
    $$\normn{\hat{\Delta}_{\bxi}}_1 \le 4 \normn{\hat{\Delta}_{\bxi, {\cS_{\bxi}}}}_1 \le 4 \sqrt{s_{\bxi}} \normn{\hat{\Delta}_{\bxi, {\cS_{\bxi}}}}_2 \lesssim s_{\bxi} \sqrt{\log d / n}.$$
    Now consider the randomness of $\norm{\nabla_{\bxi} \ell (\btheta^*, \hat{\beeta})}_{\infty}$ and we have $\normn{\hat{\Delta}_{\bxi}}_1 = \cO_{\bbP} (s_{\bxi} \sqrt{\log d / n})$.

    To show (\ref{equ:l1conv.thetaHess}), note that
    \begin{align*}
        & \abs{H (\hat{\bxi}, \bxi^*, \hat{\beeta}) - \hat{\Delta}_{\bxi}^T \nabla_{\bxi \bxi}^2 \ell (\btheta^*, \hat{\beeta}) \hat{\Delta}_{\bxi}} \\
        = & \abs{\hat{\Delta}_{\bxi}^T \brck{ \nabla_{\bxi \bxi}^2 \ell (\tilde{q} \hat{\btheta} + (1 - \tilde{q}) \btheta^*, \hat{\beeta}) - \nabla_{\bxi \bxi}^2 \ell (\btheta^*, \hat{\beeta}) } \hat{\Delta}_{\bxi}} \\
        \le & \norm{ \hat{\Delta}_{\bxi}^T \brck{ \nabla_{\bxi \bxi}^2 \ell (\tilde{q} \hat{\btheta} + (1 - \tilde{q}) \btheta^*, \hat{\beeta}) - \nabla_{\bxi \bxi}^2 \ell (\btheta^*, \hat{\beeta}) } }_{\infty} \norm{\hat{\Delta}_{\bxi}}_1 
    \end{align*}
    by Taylor's expansion for some $\tilde{q} \in [0, 1]$ as in the proof for Lemma~\ref{lem:nuisance.EP}.
    Define
    \begin{equation*}
        g_j (q') = \hat{\Delta}_{\bxi}^T \brck{ \nabla_{\bxi \theta_j}^2 \ell (q' [\tilde{q} \hat{\btheta} + (1 - \tilde{q}) \btheta^*] + (1 - q') \btheta^*, \hat{\beeta}) }
    \end{equation*}
    for coordinate $j$.
    Then we have
    \begin{equation*}
        g_j (1) = \hat{\Delta}_{\bxi}^T \brck{ \nabla_{\bxi \theta_j}^2 \ell (\tilde{q} \hat{\btheta} + (1 - \tilde{q}) \btheta^*, \hat{\beeta}) },
    \end{equation*}
    and
    \begin{equation*}
        g_j (0) = \hat{\Delta}_{\bxi}^T \brck{ \nabla_{\bxi \theta_j}^2 \ell (\btheta^*, \hat{\beeta}) }.
    \end{equation*}
    With mean value theorem again, there exists $\tilde{q}'$ such that
    \begin{equation*}
        g_j (1) = g_j (0) + \frac{d}{d q'} g_j (\tilde{q}').
    \end{equation*}
    i.e.,
    \begin{equation} \label{equ:L.Taylor.1dim}
    \begin{split}
    & \hat{\Delta}_{\bxi}^T \brck{ \nabla_{\bxi \theta_j}^2 \ell (\tilde{q} \hat{\btheta} + (1 - \tilde{q}) \btheta^*, \hat{\beeta}) - \nabla_{\bxi \theta_j}^2 \ell (\btheta^*, \hat{\beeta}) } \\
    = & \hat{\Delta}_{\bxi}^T \nabla_{\bxi} \brck{ \nabla_{\bxi \theta_j}^2 \ell (\tilde{q}' [\tilde{q} \hat{\btheta} + (1 - \tilde{q}) \btheta^*] + (1 - \tilde{q}') \btheta^*, \hat{\beeta}) - \nabla_{\bxi \theta_j}^2 \ell (\btheta^*, \hat{\beeta}) } \tilde{q} \hat{\Delta}_{\bxi} \\
    \end{split}
    \end{equation}
    for some $\tilde{q}' \in [0, 1]$.
    Since all variables are bounded, we have 
    \begin{equation*}
        \hat{\Delta}_{\bxi}^T \nabla_{\bxi} \brck{ \nabla_{\bxi \theta_j}^2 \ell (\tilde{q}' [\tilde{q} \hat{\btheta} + (1 - \tilde{q}) \btheta^*] + (1 - \tilde{q}') \btheta^*, \hat{\beeta}) - \nabla_{\bxi \theta_j}^2 \ell (\btheta^*, \hat{\beeta}) } \hat{\Delta}_{\bxi}
        \le C \hat{\Delta}_{\bxi}^T \nabla_{\bxi \bxi}^2 \ell (\btheta^*, \hat{\beeta}) \hat{\Delta}_{\bxi}
    \end{equation*}
    for some constant $C$ for all $j$.
    Hence, 
    \begin{equation}  \label{equ:Taylor.third}
        \abs{H (\hat{\bxi}, \bxi^*, \hat{\beeta}) - \hat{\Delta}_{\bxi}^T \nabla_{\bxi \bxi}^2 \ell (\btheta^*, \hat{\beeta}) \hat{\Delta}_{\bxi}}
        \le C \hat{\Delta}_{\bxi}^T \nabla_{\bxi \bxi}^2 \ell (\btheta^*, \hat{\beeta}) \hat{\Delta}_{\bxi} \norm{\hat{\Delta}_{\bxi}}_1
    \end{equation}
    Since $\normn{\hat{\Delta}_{\bxi}}_1 = \cO_{\bbP} (s_{\bxi} \sqrt{\log d / n})$,
    \begin{equation*}
        \hat{\Delta}_{\bxi}^T \nabla_{\bxi \bxi}^2 \ell (\btheta^*, \hat{\beeta}) \hat{\Delta}_{\bxi} (1 - C s_{\bxi} \sqrt{\log d / n}) \lesssim H (\hat{\bxi}, \bxi^*, \hat{\beeta}).
    \end{equation*}
    Finally, since 
    $$H (\hat{\bxi}, \bxi^*, \hat{\beeta}) \le \frac{\lambda_{\bxi}}{2} (3 \normn{\hat{\Delta}_{\bxi, {\cS_{\bxi}}}}_1 - \normn{\hat{\Delta}_{\bxi, {\cS_{\bxi}}^c}}_1) \lesssim \lambda_{\bxi} \normn{\hat{\Delta}_{\bxi, {\cS_{\bxi}}}}_1 = \cO_{\bbP} (s_{\bxi} \log d / n),$$ 
    we conclude that 
    $\hat{\Delta}_{\bxi}^T \nabla_{\bxi \bxi}^2 \ell (\btheta^*, \hat{\beeta}) \hat{\Delta}_{\bxi} = \cO_{\bbP} (s_{\bxi} \log d / n)$.
\end{proof}

\begin{lem} \label{lem:w.conv}
    Under Assumptions~\ref{asp:nuca}-\ref{asp:conv.rate},
    when $\lambda_{\bw_j} \simeq \sqrt{\log d / n}$, we have
    \begin{gather}
        \normn{\hat{\bw}_j - \bw_j^*}_1 = \cO_{\bbP} ((s_{\bxi} \vee s_{\bw_j}) \sqrt{\log d / n}), \label{equ:l1conv.w} \\
        (\hat{\bv}_j - \bv_j^*)^T \nabla_{\bxi \bxi}^2 \ell (\btheta^*, \hat{\beeta}) (\hat{\bv}_j - \bv_j^*) = \cO_{\bbP} ((s_{\bxi} \vee s_{\bw_j}) \log d / n), \label{equ:l1conv.wHess} \\
        (\hat{\bv}_j - \bv_j^*)^T \nabla_{\bxi \bxi}^2 \ell (\hat{\btheta}, \hat{\beeta}) (\hat{\bv}_j - \bv_j^*) = \cO_{\bbP} ((s_{\bxi} \vee s_{\bw_j}) \log d / n). \label{equ:l1conv.wHess.hat} 
    \end{gather}
\end{lem}

\begin{proof}
    Let $\hat{\Delta}_{\bw_j} := \hat{\bw}_j - \bw_j^*$ and $\hat{\Delta}_{\bv_j} := \hat{\bv}_j - \bv_j^*$.
    Since $\normn{\bw_j^*}_1 \geq \norm{\hat{\bw}_j}_1$, we have
    \[ \sum_{j \in \cS_{\bw_j}} \absn{w_j^*} 
    \geq \sum_{j \in \cS_{\bw_j}} \absn{\hat{w}_j} + \sum_{j \in \cS_{\bw_j}^c} \absn{\hat{w}_j} 
    \geq \sum_{j \in \cS_{\bw_j}} \absn{w_j^*} - \sum_{j \in \cS_{\bw_j}} \absn{\hat{\Delta}_{\bw_j, j}} + \sum_{j \in \cS_{\bw_j}^c} \absn{\hat{w}_j}.
    \]
    Hence 
    $$\sum_{j \in \cS_{\bw_j}} \absn{\hat{\Delta}_{\bw_j, j}} \geq \sum_{j \in \cS_{\bw_j}^c} \absn{\hat{w}_j} = \sum_{j \in \cS_{\bw_j}^c} \absn{\hat{\Delta}_{\bw_j, j}}, $$ that is, $\normn{\hat{\Delta}_{\bw_j, \cS_{\bw_j}^c}}_1 \leq \normn{\hat{\Delta}_{\bw_j, \cS_{\bw_j}}}_1$.
    Therefore, $\normn{\hat{\Delta}_{\bw_j}}_1 \leq 2 \normn{\hat{\Delta}_{\bw_j, \cS_{\bw_j}}}_1$.
    Consider the following function
    \begin{align*}
        & \hat{\Delta}_{\bw_j}^T \nabla^2_{\bnu_j \bnu_j} \ell (\hat{\btheta}, \hat{\beeta}) \hat{\Delta}_{\bw_j} \\
        = & \brck{\nabla^2_{\theta_j \bnu_j} \ell (\hat{\btheta}, \hat{\beeta}) - \bw_j^{*T} \nabla^2_{\bnu_j \bnu_j} \ell (\hat{\btheta}, \hat{\beeta})} \hat{\Delta}_{\bw_j} - \brck{\nabla^2_{\theta_j \bnu_j} \ell (\hat{\btheta}, \hat{\beeta}) - \hat{\bw}_j^T \nabla^2_{\bnu_j \bnu_j} \ell (\hat{\btheta}, \hat{\beeta})} \hat{\Delta}_{\bw_j} \\
        =: & I_1 + I_2.
    \end{align*}
    According to the definition of $\hat{\Delta}_{\bw_j}$,
    \[ I_2 \leq \normn{\nabla^2_{\theta_j \bnu_j} \ell (\hat{\btheta}, \hat{\beeta}) - \hat{\bw}_j^T \nabla^2_{\bnu_j \bnu_j} \ell (\hat{\btheta}, \hat{\beeta})}_{\infty} \normn{\hat{\Delta}_{\bw_j}}_1 \le \lambda_{\bw_j} \normn{\hat{\Delta}_{\bw_j}}_1. \]
    Note that for $I_1$, 
    \begin{align*}
        I_1
        = \bv_j^{*T} \nabla_{\bxi \bnu_j}^2 \ell (\btheta^*, \hat{\beeta}) \hat{\Delta}_{\bw_j} 
        + \bv_j^{*T} [\nabla_{\bxi \bnu_j}^2 \ell (\hat{\btheta}, \hat{\beeta}) - \nabla_{\bxi \bnu_j}^2 \ell (\btheta^*, \hat{\beeta})] \hat{\Delta}_{\bw_j} 
        =: I_{11} + I_{12}.
    \end{align*}
    Since 
    $$\bbE [\nabla^2_{\bnu_j \bxi} \ell (\btheta^*, \bar{\beeta}) \bv_j^{*}] = \Ib^*_{\bnu_j \theta_j} - \Ib^*_{\bnu_j \bnu_j} \bw_j^* = 0,$$
    we can use the similar proof of Lemma~\ref{lem:concen.gradHess} to show that
    \[ \abs{I_{11}} \le \normn{\bv_j^{*T} \nabla_{\bxi \bnu_j}^2 \ell (\btheta^*, \hat{\beeta})}_{\infty} \normn{\hat{\Delta}_{\bw_j}}_1 \lesssim \sqrt{\log d / n} \normn{\hat{\Delta}_{\bw_j}}_1. \]
    For $I_{22}$, use similar trick as in (\ref{equ:L.Taylor.1dim}) and we have
    \begin{equation*}
        \absn{I_{12}} 
        \lesssim \hat{\Delta}_{\bxi}^T \nabla_{\bxi \bnu_j}^2 \ell (\btheta^*, \hat{\beeta}) \hat{\Delta}_{\bw_j} 
        = \hat{\Delta}_{\bxi}^T \nabla_{\bxi \bxi}^2 \ell (\btheta^*, \hat{\beeta}) \hat{\Delta}_{\bv_j} 
    \end{equation*}
    By Assumption~\ref{asp:V.sc}, $\ell (\btheta^*, \hat{\beeta})$ is positive semidefinite when $n$ is large enough.
    By the Cauchy-Schwartz inequality,
    \begin{equation*}
        \begin{split}
            \absn{I_{12}} 
            & \le \brck{ \hat{\Delta}_{\bxi}^T \nabla_{\bxi \bxi}^2 \ell (\btheta^*, \hat{\beeta}) \hat{\Delta}_{\bxi} }^{1/2} 
            \brck{ \hat{\Delta}_{\bv_j}^T \nabla_{\bxi \bxi}^2 \ell (\btheta^*, \hat{\beeta}) \hat{\Delta}_{\bv_j} }^{1/2} \\
            & \lesssim \sqrt{s_{\bxi} \log d / n} \brck{ \hat{\Delta}_{\bw_j}^T \nabla_{\bnu_j \bnu_j}^2 \ell (\btheta^*, \hat{\beeta}) \hat{\Delta}_{\bw_j} }^{1/2}.
        \end{split}
    \end{equation*}
    Combine the bounds for $I_{11}, I_{12}, I_{2}$,
    \[ \hat{\Delta}_{\bw_j}^T \nabla^2_{\bnu_j \bnu_j} \ell (\hat{\btheta}, \hat{\beeta}) \hat{\Delta}_{\bw_j} 
    \lesssim \sqrt{\log d / n} \normn{\hat{\Delta}_{\bw_j}}_1 + \sqrt{s_{\bxi} \log d / n} \brck{ \hat{\Delta}_{\bw_j}^T \nabla_{\bnu_j \bnu_j}^2 \ell (\btheta^*, \hat{\beeta}) \hat{\Delta}_{\bw_j} }^{1/2}. \]
    Further notice that
    \begin{equation*}
        \abs{\hat{\Delta}_{\bw_j}^T \brck{\nabla^2_{\bnu_j \bnu_j} \ell (\hat{\btheta}, \hat{\beeta}) - \nabla_{\bnu_j \bnu_j}^2 \ell (\btheta^*, \hat{\beeta})} \hat{\Delta}_{\bw_j}}
        \lesssim s_{\bxi} \sqrt{\log d / n} \abs{\hat{\Delta}_{\bw_j}^T \brck{\nabla_{\bnu_j \bnu_j}^2 \ell (\btheta^*, \hat{\beeta})} \hat{\Delta}_{\bw_j}} 
    \end{equation*}
    using similar trick as in (\ref{equ:L.Taylor.1dim}) and $\normn{\hat{\Delta}_{\bxi}}_1 = \cO_{\bbP} (s_{\bxi} \sqrt{\log d / n})$.
    Therefore,
    \begin{equation} \label{equ:w.conv.theta.star}
        \hat{\Delta}_{\bw_j}^T \nabla_{\bnu_j \bnu_j}^2 \ell (\btheta^*, \hat{\beeta}) \hat{\Delta}_{\bw_j} (1 - C s_{\bxi} \sqrt{\log d / n}) 
        \lesssim \hat{\Delta}_{\bw_j}^T \nabla_{\bnu_j \bnu_j}^2 \ell (\hat{\btheta}, \hat{\beeta}) \hat{\Delta}_{\bw_j}
    \end{equation}
    for some constant $C > 0$.
    Consequently, 
    \begin{equation} \label{equ:w.conv.inequ}
        \hat{\Delta}_{\bw_j}^T \nabla^2_{\bnu_j \bnu_j} \ell (\hat{\btheta}, \hat{\beeta}) \hat{\Delta}_{\bw_j} 
        \lesssim \sqrt{\log d / n} \normn{\hat{\Delta}_{\bw_j}}_1 + \sqrt{s_{\bxi} \log d / n} \brck{ \hat{\Delta}_{\bw_j}^T \nabla_{\bnu_j \bnu_j}^2 \ell (\hat{\btheta}, \hat{\beeta}) \hat{\Delta}_{\bw_j} }^{1/2}.
    \end{equation}

    Consider two cases.
    If $\brck{ \hat{\Delta}_{\bw_j}^T \nabla_{\bnu_j \bnu_j}^2 \ell (\hat{\btheta}, \hat{\beeta}) \hat{\Delta}_{\bw_j} }^{1/2} \lesssim \sqrt{s_{\bxi} \log d / n}$,
    according to Lemma~\ref{lem:L.operationnorm.bound}, 
    $$s_{\bw_j}^{-1/2} \normn{\hat{\Delta}_{\bw_j, \cS_{\bw_j}}}_1 \lesssim [\hat{\Delta}_{\bw_j}^T \nabla^2_{\bnu_j \bnu_j} \ell (\hat{\btheta}, \hat{\beeta}) \hat{\Delta}_{\bw_j}]^{1/2}.$$
    Combine the upper and lower bounds of $\hat{\Delta}_{\bw_j}^T \nabla^2_{\bnu_j \bnu_j} \ell (\hat{\btheta}, \hat{\beeta}) \hat{\Delta}_{\bw_j}$, 
    we have $\normn{\hat{\Delta}_{\bw_j, \cS_{\bw_j}}}_1 \lesssim \sqrt{s_{\bw_j} s_{\bxi} \log d / n}$.
    Therefore, we have 
    \[ \normn{\hat{\Delta}_{\bw_j}}_1 \leq 2 \normn{\hat{\Delta}_{\bw_j, \cS_{\bw_j}}}_1 \lesssim (s_{\bw_j} \vee s_{\bxi}) \sqrt{\log d / n}. \]
    If $\hat{\Delta}_{\bw_j}^T \nabla_{\bnu_j \bnu_j}^2 \ell (\hat{\btheta}, \hat{\beeta}) \hat{\Delta}_{\bw_j} \gtrsim \sqrt{s_{\bxi} \log d / n}$,
    (\ref{equ:w.conv.inequ}) is equivalent to 
    \begin{equation*}
        \brck{ \hat{\Delta}_{\bw_j}^T \nabla^2_{\bnu_j \bnu_j} \ell (\hat{\btheta}, \hat{\beeta}) \hat{\Delta}_{\bw_j} }^{1/2}
        \brce{\brck{ \hat{\Delta}_{\bw_j}^T \nabla^2_{\bnu_j \bnu_j} \ell (\hat{\btheta}, \hat{\beeta}) \hat{\Delta}_{\bw_j} }^{1/2} - \sqrt{s_{\bxi} \log d / n}}
        \lesssim \sqrt{\log d / n} \normn{\hat{\Delta}_{\bw_j}}_1.
    \end{equation*}
    Since $\normn{\hat{\Delta}_{\bw_j}}_1^2 
    \le 4 \normn{\hat{\Delta}_{\bw_j, \cS_{\bw_j}}}_1^2 
    \le 4 s_{\bw_j} \normn{\hat{\Delta}_{\bw_j, \cS_{\bw_j}}}_2^2
    \lesssim s_{\bw_j} \hat{\Delta}_{\bw_j}^T \nabla^2_{\bnu_j \bnu_j} \ell (\hat{\btheta}, \hat{\beeta}) \hat{\Delta}_{\bw_j}$,
    \begin{equation*}
        \brck{ \hat{\Delta}_{\bw_j}^T \nabla^2_{\bnu_j \bnu_j} \ell (\hat{\btheta}, \hat{\beeta}) \hat{\Delta}_{\bw_j} }^{1/2} - \sqrt{s_{\bxi} \log d / n}
        \lesssim \sqrt{s_{\bw_j} \log d / n}.
    \end{equation*}
    Hence, we have
    \begin{equation*}
        \brck{ \hat{\Delta}_{\bw_j}^T \nabla^2_{\bnu_j \bnu_j} \ell (\hat{\btheta}, \hat{\beeta}) \hat{\Delta}_{\bw_j} }^{1/2} 
        \lesssim \sqrt{(s_{\bxi} + s_{\bw_j}) \log d / n}.
    \end{equation*}
    Using the same arguments as in the first case, 
    \[ \normn{\hat{\Delta}_{\bw_j}}_1 \lesssim (s_{\bw_j} + s_{\bxi}) \sqrt{\log d / n}. \]
    In both cases, we can conclude that $ \normn{\hat{\Delta}_{\bw_j}}_1 \lesssim (s_{\bw_j} \vee s_{\bxi}) \sqrt{\log d / n}$ 
    and that 
    \[ \hat{\Delta}_{\bw_j}^T \nabla^2_{\bnu_j \bnu_j} \ell (\hat{\btheta}, \hat{\beeta}) \hat{\Delta}_{\bw_j} 
    \lesssim (s_{\bxi} \vee s_{\bw_j}) \log d / n. \]
    Finally, according to (\ref{equ:w.conv.theta.star}), we have
    \begin{equation*}
        \hat{\Delta}_{\bw_j}^T \nabla^2_{\bnu_j \bnu_j} \ell (\btheta^*, \hat{\beeta}) \hat{\Delta}_{\bw_j} 
    \lesssim (s_{\bxi} \vee s_{\bw_j}) \log d / n.
    \end{equation*}
\end{proof}

\begin{lem} \label{lem:L.operationnorm.bound}
    Denote
    \[ \kappa_D (s_{\bw_j}) = \min \brce{\frac{s_{\bw_j}^{1/2} [\bw_j^T \nabla^2_{\bnu_j \bnu_j} \ell (\hat{\btheta}, \hat{\beeta}) \bw_j]^{1/2}}{\norm{\bw_{j, \cS_{\bw_j}}}_1}: \bw_j \in \bbR^{d-1} \backslash \{0\}, \normn{\bw_{j, \cS_{\bw_j}^c}}_1 \le \xi \normn{\bw_{j, \cS_{\bw_j}}}_1}, \]
    where $\xi$ is a positive constant.
    Under Assumptions~\ref{asp:nuca}-\ref{asp:V.sc}, $\kappa_D (s_{\bw_j}) \ge \kappa / \sqrt{2}$ with probability tending to one.
\end{lem}

\begin{proof}
    Since 
    $\normn{\bv_{j, \cS_{\bw_j}}}_1 \le s_{\bw_j}^{1/2} \normn{\bv_{j, \cS_{\bw_j}}}_2 \le s_{\bw_j}^{1/2} \normn{\bv_j}_2$,
    we have
    \[ \kappa_D^2 (s_{\bw_j}) \ge \min \brce{\frac{\bw_j^T \nabla^2_{\bnu_j \bnu_j} \ell (\hat{\btheta}, \hat{\beeta}) \bw_j}{\norm{\bw_j}_2^2}: \bw_j \in \bbR^{d-1} \backslash \{0\}, \normn{\bw_{j, \cS_{\bw_j}^c}}_1 \le \xi \normn{\bw_{j, \cS_{\bw_j}}}_1}. \]
    Note that
    \[ \frac{\bw_j^T \nabla^2_{\bnu_j \bnu_j} \ell (\hat{\btheta}, \hat{\beeta}) \bw_j}{\norm{\bw_j}_2^2} 
    = \frac{\bw_j^T \nabla^2_{\bnu_j \bnu_j} \ell (\btheta^*, \hat{\beeta}) \bw_j}{\norm{\bw_j}_2^2}
    + \frac{\bw_j^T [\nabla^2_{\bnu_j \bnu_j} \ell (\hat{\btheta}, \hat{\beeta}) - \nabla^2_{\bnu_j \bnu_j} \ell (\btheta^*, \hat{\beeta})] \bw_j}{\norm{\bw_j}_2^2} \]
    and 
    \[ \bw_j^T [\nabla^2_{\bnu_j \bnu_j} \ell (\hat{\btheta}, \hat{\beeta}) - \nabla^2_{\bnu_j \bnu_j} \ell (\btheta^*, \hat{\beeta})] \bw_j = \cO_{\bbP} (s_{\bxi} \sqrt{\log d / n}) = o_{\bbP} (1) \]
    by similar proof as (\ref{equ:Taylor.third}).
    Therefore, with probability tending to one,
    \begin{equation*}
    \begin{split}
        & \frac{\bw_j^T \nabla^2_{\bnu_j \bnu_j} \ell (\hat{\btheta}, \hat{\beeta}) \bw_j}{\norm{\bw_j}_2^2} 
        \ge \frac{3}{4} \frac{\bw_j^T \nabla^2_{\bnu_j \bnu_j} \ell (\btheta^*, \hat{\beeta}) \bw_j}{\norm{\bw_j}_2^2} \\
        = & \frac{3}{4} \frac{\bw_j^T \Ib_{\bnu_j \bnu_j} \bw_j}{\norm{\bw_j}_2^2}
        + \frac{3}{4} \frac{\bw_j^T [\nabla^2_{\bnu_j \bnu_j} \ell (\btheta^*, \hat{\beeta}) - \Ib_{\bnu_j \bnu_j}] \bw_j}{\norm{\bw_j}_2^2} \\
        \ge & \frac{3}{4} \lambda_{\min} (\Ib_{\bnu_j \bnu_j} )
        - \frac{3}{4} \abs{\frac{\bw_j^T [\nabla^2_{\bnu_j \bnu_j} \ell (\btheta^*, \hat{\beeta}) - \Ib_{\bnu_j \bnu_j}] \bw_j}{\norm{\bw_j}_2^2}} \\
        \ge & \frac{3}{4} \brck{ \kappa^2 - \frac{\normn{\bw_j}_1^2 \normn{\nabla^2_{\bnu_j \bnu_j} \ell (\btheta^*, \hat{\beeta}) - \Ib_{\bnu_j \bnu_j}}_{\infty}}{\norm{\bw_j}_2^2}}
    \end{split}
    \end{equation*}
    In addition, from 
    $$\normn{\bw_j}_1^2 \le (\xi + 1)^2 \normn{\bw_{j, \cS_{\bw_j}}}_1^2 \le s_{\bw_j} (\xi + 1)^2 \normn{\bw_j}_2^2$$
    we get
    \[ \frac{\bw_j^T \nabla^2_{\bnu_j \bnu_j} \ell (\hat{\btheta}, \hat{\beeta}) \bw_j}{\norm{\bw_j}_2^2} \ge \frac{3}{4} \brck{\kappa^2 - s_{\bw_j} (\xi + 1)^2 \normn{\nabla^2_{\bnu_j \bnu_j} \ell (\btheta^*, \hat{\beeta}) - \Ib_{\bnu_j \bnu_j}}_{\infty}}. \]
    Similar to the proof of Lemma~\ref{lem:concen.gradHess}, we can obtain 
    $$\normn{\nabla^2_{\bnu_j \bnu_j} \ell (\btheta^*, \hat{\beeta}) - \Ib_{\bnu_j \bnu_j}}_{\infty} = \cO_{\bbP} (\sqrt{\log d / n}).$$
    Hence we have $\normn{\nabla^2_{\bnu_j \bnu_j} \ell (\btheta^*, \hat{\beeta}) - \Ib_{\bnu_j \bnu_j}}_{\infty} = o_{\bbP} (1)$ by Assumption~\ref{asp:conv.rate}.
    When $n$ is large enough, 
    $$\normn{\nabla^2_{\bnu_j \bnu_j} \ell (\btheta^*, \hat{\beeta}) - \Ib_{\bnu_j \bnu_j}}_{\infty} \le \kappa^2 / [3 (\xi + 1)^2].$$
    Therefore, $\kappa_D (s_{\bw_j}) \ge \kappa / \sqrt{2}$ with probability tending to one.
\end{proof}

\begin{lem}[Local smoothness conditions on the loss function] \label{lem:smooth}
    Under Assumptions~\ref{asp:nuca}-\ref{asp:positivity.MSTP} and~\ref{asp:conv.rate}, we have
    \begin{gather}
        \bv_j^{*T} [\nabla_{\bxi} \ell(\hat{\btheta}, \hat{\beeta}) - \nabla_{\bxi} \ell(\btheta^*, \hat{\beeta}) - \nabla_{\bxi \bxi}^2 \ell(\btheta^*, \hat{\beeta}) (\hat{\btheta} - \btheta^*)] = \cO_{\bbP} ((s_{\bxi} \vee s_{\bw_j}) \log d / n), \label{equ:smooth.gradient} \\
        (\hat{\bv}_j - \bv_j^*)^T [\nabla_{\bxi} \ell(\hat{\btheta}, \hat{\beeta}) - \nabla_{\bxi} \ell(\btheta^*, \hat{\beeta})] = \cO_{\bbP} ((s_{\bxi} \vee s_{\bw_j}) \log d / n). \label{equ:smooth.v.gradient}
    \end{gather}
    The same results hold for $\hat{\btheta}_0 = (\theta_0, 0, \hat{\bnu}_{j}^T)^T$, where $\theta_j$ is the parameter we are interested in.
\end{lem}

\begin{proof}
    Using the similar proof as (\ref{equ:L.Taylor.1dim}), 
    \begin{equation*}
    \begin{split}
        & \abs{\bv_j^{*T} [\nabla_{\bxi} \ell(\hat{\btheta}, \hat{\beeta}) - \nabla_{\bxi} \ell(\btheta^*, \hat{\beeta}) - \nabla_{\bxi \bxi}^2 \ell(\btheta^*, \hat{\beeta}) (\hat{\btheta} - \btheta^*)]} \\
        = & \abs{\bv_j^{*T} [\nabla_{\bxi \bxi}^2 \ell (\tilde{q} \hat{\btheta} + (1 - \tilde{q}) \btheta^*, \hat{\beeta}) \hat{\Delta}_{\bxi} - \nabla_{\bxi \bxi}^2 \ell(\btheta^*, \hat{\beeta}) \hat{\Delta}_{\bxi}]} \\
        \lesssim & \normn{\bv_j^{*}}_1 \normn{\hat{\Delta}_{\bxi}^T \nabla_{\bxi \bxi}^2 \ell(\btheta^*, \hat{\beeta}) \hat{\Delta}_{\bxi} }_{\infty}
        \lesssim (s_{\bxi} \vee s_{\bw_j}) \log d / n
    \end{split}
    \end{equation*}
    for some $\tilde{q} \in [0,1]$ by (\ref{equ:l1conv.thetaHess}).
    For (\ref{equ:smooth.v.gradient}), we have
    \begin{equation*}
    \begin{split}
        & \abs{(\hat{\bv}_j - \bv_j^*)^T [\nabla_{\bxi} \ell(\hat{\btheta}, \hat{\beeta}) - \nabla_{\bxi} \ell(\btheta^*, \hat{\beeta})]} \\
        = & \abs{\hat{\Delta}_{\bv_j} \nabla_{\bxi \bxi}^2 \ell (\tilde{q} \hat{\btheta} + (1 - \tilde{q}) \btheta^*, \hat{\beeta}) \hat{\Delta}_{\bxi}} \\
        \le & \abs{\hat{\Delta}_{\bv_j} \nabla_{\bxi \bxi}^2 \ell (\btheta^*, \hat{\beeta}) \hat{\Delta}_{\bxi}}
        + \abs{\hat{\Delta}_{\bv_j} [\nabla_{\bxi \bxi}^2 \ell (\tilde{q} \hat{\btheta} + (1 - \tilde{q}) \btheta^*, \hat{\beeta}) - \nabla_{\bxi \bxi}^2 \ell (\btheta^*, \hat{\beeta})] \hat{\Delta}_{\bxi}}
    \end{split}
    \end{equation*}
    by Taylor's expansion.
    For the first term on the right-hand side of the inequality,
    the Cauchy-Schwartz inequality yields
    \begin{equation*}
    \begin{split}
        & \abs{\hat{\Delta}_{\bv_j} \nabla_{\bxi \bxi}^2 \ell (\btheta^*, \hat{\beeta}) \hat{\Delta}_{\bxi}} \\
        \le & \abs{\hat{\Delta}_{\bv_j} \nabla_{\bxi \bxi}^2 \ell (\btheta^*, \hat{\beeta}) \hat{\Delta}_{\bv_j}}^{1/2}
        \abs{\hat{\Delta}_{\bxi} \nabla_{\bxi \bxi}^2 \ell (\btheta^*, \hat{\beeta}) \hat{\Delta}_{\bxi}}^{1/2} \\
        \lesssim & \sqrt{s_{\bxi} \log d / n} \sqrt{(s_{\bxi} \vee s_{\bw_j}) \log d / n}
        = \cO_{\bbP} ((s_{\bxi} \vee s_{\bw_j}) \log d / n)
    \end{split}
    \end{equation*}
    by (\ref{equ:l1conv.thetaHess}) and (\ref{equ:l1conv.wHess}) and the positive semidefiniteness of $\ell (\btheta^*, \hat{\beeta})$.
    For the second term on the right-hand side, 
    similar proof as (\ref{equ:L.Taylor.1dim}) yields
    \begin{equation*}
    \begin{split}
        & \abs{\hat{\Delta}_{\bv_j} [\nabla_{\bxi \bxi}^2 \ell (\tilde{q} \hat{\btheta} + (1 - \tilde{q}) \btheta^*, \hat{\beeta}) - \nabla_{\bxi \bxi}^2 \ell (\btheta^*, \hat{\beeta})] \hat{\Delta}_{\bxi}} \\
        \lesssim & \normn{\hat{\Delta}_{\bv_j}}_1 \normn{ \hat{\Delta}_{\bxi}^T \nabla_{\bxi \bxi}^2 \ell(\btheta^*, \hat{\beeta}) \hat{\Delta}_{\bxi} }_{\infty} \\
        \lesssim & (s_{\bxi} \vee s_{\bw_j}) \sqrt{\log d / n}) ((s_{\bxi} \vee s_{\bw_j}) \log d / n)
    \end{split}
    \end{equation*}
    by (\ref{equ:l1conv.thetaHess}) and (\ref{equ:l1conv.w}).
\end{proof}

Finally, we give the proof of Theorem~\ref{thm:main}.

\begin{proof}
    For the assumptions needed in \citet[Theorem 3.2]{ning2017general},
    Lemmas~\ref{lem:theta.conv} and~\ref{lem:w.conv} satisfy Assumption 1,
    Lemma~\ref{lem:concen.gradHess} satisfies Assumption 2,
    Lemma~\ref{lem:smooth} satisfies Assumption 3,
    and Lemma~\ref{lem:CLT} satisfies Assumption 4.
    Now we only need to verify that $\hat{I}_{\theta_j | \bnu_j} - I^*_{\theta_j | \bnu_j} = o_{\bbP} (1)$.

    First note that
    \[ \hat{I}_{\theta_j | \bnu_j} = \nabla^2_{\theta_j \theta_j} \ell (\hat{\btheta}, \hat{\beeta}) - \hat{\bw}_j^T \nabla^2_{\bnu_j \theta_j} \ell (\hat{\btheta}, \hat{\beeta}) \]
    and
    \[ I^*_{\theta_j | \bnu_j} = I^*_{\theta_j \theta_j} - \bw_j^{*T} \Ib^*_{\bnu_j \theta_j}, \qquad \text{where } \bw_j^* = \Ib^{*-1}_{\bnu_j \bnu_j} \Ib^*_{\bnu_j \theta_j}. \]
    For the second part of $\hat{I}_{\theta_j | \bnu_j}$ and $I^*_{\theta_j | \bnu_j}$,
    \begin{equation*}
    \begin{split}
        & \hat{\bw}_j^T \nabla^2_{\bnu_j \theta_j} \ell (\hat{\btheta}, \hat{\beeta}) - \bw_j^{*T} \Ib^*_{\bnu_j \theta_j} \\
        \le & \abs{(\hat{\bw}_j - \bw_j^{*})^T \nabla^2_{\bnu_j \theta_j} \ell (\hat{\btheta}, \hat{\beeta})}
        + \abs{\bw_j^{*T} [\nabla^2_{\bnu_j \theta_j} \ell (\hat{\btheta}, \hat{\beeta}) - \nabla^2_{\bnu_j \theta_j} \ell (\btheta^*, \hat{\beeta})]} \\
        & + \abs{\bw_j^{*T} [\nabla^2_{\bnu_j \theta_j} \ell (\btheta^*, \hat{\beeta}) - \Ib^*_{\bnu_j \theta_j}]} \\
        =: & J_1 + J_2 + J_3.
    \end{split}
    \end{equation*}
    With $\be = [1, 0, \dots, 0]$,
    \begin{equation*}
        J_1 
        = \abs{\hat{\Delta}_{\bv_j}^T \nabla^2_{\bxi \bxi} \ell (\hat{\btheta}, \hat{\beeta}) \be} 
        \le \norm{\hat{\Delta}_{\bv_j}^T \nabla^2_{\bxi \bxi} \ell (\hat{\btheta}, \hat{\beeta}) \hat{\Delta}_{\bv_j}}_2 
        \norm{\be^T \nabla^2_{\bxi \bxi} \ell (\hat{\btheta}, \hat{\beeta}) \be}
        \lesssim \sqrt{(s_{\bxi} \vee s_{\bw_j}) \log d / n}
    \end{equation*}
    by the Cauchy-Schwartz inequality, the bound (\ref{equ:l1conv.wHess.hat}) and the boundedness of the variables.
    Using similar arguments as in (\ref{equ:L.Taylor.1dim}),
    \[ J_2 \lesssim \abs{\bv_j^{*T} \nabla^2_{\bxi \bxi} \ell (\btheta^*, \hat{\beeta}) (\hat{\btheta} - \btheta^*)}. \]
    Similarly, with the Cauchy-Schwartz inequality, the bound (\ref{equ:l1conv.thetaHess}) and the boundedness of the variables,
    \begin{equation*}
        J_2 
        \le \norm{(\hat{\btheta} - \btheta^*)^T \nabla^2_{\bxi \bxi} \ell (\hat{\btheta}, \hat{\beeta}) (\hat{\btheta} - \btheta^*)}_2 
        \lesssim \sqrt{s_{\bxi} \log d / n}
    \end{equation*}
    Using similar arguments as the proof of (\ref{equ:concen.hess}), we have that 
    $$J_3 \lesssim \sqrt{\log d / n}.$$
    In addition, similar arguments also imply that 
    $$\absn{\nabla^2_{\theta_j \theta_j} \ell (\hat{\btheta}, \hat{\beeta}) - I^*_{\theta_j \theta_j}} \lesssim \sqrt{\log d / n}.$$
    Therefore, $\absn{ \hat{I}_{\theta_j | \bnu_j} - I^*_{\theta_j | \bnu_j}} \lesssim \sqrt{(s_{\bxi} \vee s_{\bw_j}) \log d / n} = o_{\bbP} (1)$ and thus $\sqrt{n} (\tilde{\theta}_j - \theta_j^*) I^{*}_{\theta_j | \bnu_j} \sqrt{\sigma_j^*} \Rightarrow N(0, 1)$.
\end{proof}

\vskip 0.2in
\bibliography{bibfile}

\begin{thebibliography}{}

\bibitem[Antos et~al., 2007]{antos2007fitted}
Antos, A., Szepesv{\'a}ri, C., and Munos, R. (2007).
\newblock Fitted q-iteration in continuous action-space mdps.
\newblock {\em Advances in neural information processing systems}, 20.

\bibitem[Bang and Robins, 2005]{bang2005doubly}
Bang, H. and Robins, J.~M. (2005).
\newblock Doubly robust estimation in missing data and causal inference models.
\newblock {\em Biometrics}, 61(4):962--973.

\bibitem[Boruvka et~al., 2018]{boruvka2018assessing}
Boruvka, A., Almirall, D., Witkiewitz, K., and Murphy, S.~A. (2018).
\newblock Assessing time-varying causal effect moderation in mobile health.
\newblock {\em Journal of the American Statistical Association}, 113(523):1112--1121.

\bibitem[Chernozhukov et~al., 2018a]{chernozhukov2018double}
Chernozhukov, V., Chetverikov, D., Demirer, M., Duflo, E., Hansen, C., Newey, W., and Robins, J. (2018a).
\newblock Double/debiased machine learning for treatment and structural parameters.
\newblock {\em The Econometrics Journal}, 21(1):C1--C68.

\bibitem[Chernozhukov et~al., 2018b]{chernozhukov2018plug}
Chernozhukov, V., Nekipelov, D., Semenova, V., and Syrgkanis, V. (2018b).
\newblock Plug-in regularized estimation of high dimensional parameters in nonlinear semiparametric models.
\newblock Technical report, Centre for Microdata Methods and Practice, Institute for Fiscal Studies.

\bibitem[Fan and Li, 2001]{fan2001variable}
Fan, J. and Li, R. (2001).
\newblock Variable selection via nonconcave penalized likelihood and its oracle properties.
\newblock {\em Journal of the American statistical Association}, 96(456):1348--1360.

\bibitem[Ferguson, 2017]{ferguson2017course}
Ferguson, T.~S. (2017).
\newblock {\em A Course in Large Sample Theory}.
\newblock Routledge.

\bibitem[Gao et~al., 2022]{gao2022non}
Gao, D., Liu, Y., and Zeng, D. (2022).
\newblock Non-asymptotic properties of individualized treatment rules from sequentially rule-adaptive trials.
\newblock {\em Journal of Machine Learning Research}, 23(250):1--42.

\bibitem[Hastie et~al., 2017]{hastie2017elements}
Hastie, T., Tibshirani, R., and Friedman, J. (2017).
\newblock {\em The elements of statistical learning: data mining, inference, and prediction}.
\newblock Springer.

\bibitem[Hu and Wager, 2023]{hu2023off}
Hu, Y. and Wager, S. (2023).
\newblock Off-policy evaluation in partially observed markov decision processes under sequential ignorability.
\newblock {\em The Annals of Statistics}, 51(4):1561--1585.

\bibitem[Jeng et~al., 2018]{jeng2018high}
Jeng, X.~J., Lu, W., and Peng, H. (2018).
\newblock High-dimensional inference for personalized treatment decision.
\newblock {\em Electronic Journal of Statistics}, 12(1):2074.

\bibitem[Jiang and Li, 2016]{jiang2016doubly}
Jiang, N. and Li, L. (2016).
\newblock Doubly robust off-policy value evaluation for reinforcement learning.
\newblock In {\em International Conference on Machine Learning}, pages 652--661. PMLR.

\bibitem[Kallus and Uehara, 2020]{kallus2020double}
Kallus, N. and Uehara, M. (2020).
\newblock Double reinforcement learning for efficient off-policy evaluation in markov decision processes.
\newblock {\em Journal of Machine Learning Research}, 21(167).

\bibitem[Kallus and Uehara, 2022]{kallus2022efficiently}
Kallus, N. and Uehara, M. (2022).
\newblock Efficiently breaking the curse of horizon in off-policy evaluation with double reinforcement learning.
\newblock {\em Operations Research}, 70(6):3282--3302.

\bibitem[Kosorok and Laber, 2019]{kosorok2019precision}
Kosorok, M.~R. and Laber, E.~B. (2019).
\newblock Precision medicine.
\newblock {\em Annual review of statistics and its application}, 6(1):263--286.

\bibitem[Lax and Terrell, 2014]{lax2014calculus}
Lax, P.~D. and Terrell, M.~S. (2014).
\newblock {\em Calculus with applications}, volume~4.
\newblock Springer.

\bibitem[Liang et~al., 2022]{liang2022estimation}
Liang, M., Choi, Y.-G., Ning, Y., Smith, M.~A., and Zhao, Y.-Q. (2022).
\newblock Estimation and inference on high-dimensional individualized treatment rule in observational data using split-and-pooled de-correlated score.
\newblock {\em Journal of Machine Learning Research}, 23(262):1--65.

\bibitem[Liao et~al., 2021]{liao2021off}
Liao, P., Klasnja, P., and Murphy, S. (2021).
\newblock Off-policy estimation of long-term average outcomes with applications to mobile health.
\newblock {\em Journal of the American Statistical Association}, 116(533):382--391.

\bibitem[Liao et~al., 2022]{liao2022batch}
Liao, P., Qi, Z., Wan, R., Klasnja, P., and Murphy, S.~A. (2022).
\newblock Batch policy learning in average reward markov decision processes.
\newblock {\em The Annals of Statistics}, 50(6):3364--3387.

\bibitem[Liu et~al., 2018a]{liu2018breaking}
Liu, Q., Li, L., Tang, Z., and Zhou, D. (2018a).
\newblock Breaking the curse of horizon: Infinite-horizon off-policy estimation.
\newblock In {\em Advances in Neural Information Processing Systems}, pages 5356--5366.

\bibitem[Liu et~al., 2018b]{liu2018augmented}
Liu, Y., Wang, Y., Kosorok, M.~R., Zhao, Y., and Zeng, D. (2018b).
\newblock Augmented outcome-weighted learning for estimating optimal dynamic treatment regimens.
\newblock {\em Statistics in Medicine}, 37(26):3776--3788.

\bibitem[Luckett et~al., 2020]{luckett2020estimating}
Luckett, D.~J., Laber, E.~B., Kahkoska, A.~R., Maahs, D.~M., Mayer-Davis, E., and Kosorok, M.~R. (2020).
\newblock Estimating dynamic treatment regimes in mobile health using v-learning.
\newblock {\em Journal of the American Statistical Association}, 115(530):692--706.

\bibitem[Luedtke and Van Der~Laan, 2016]{luedtke2016statistical}
Luedtke, A.~R. and Van Der~Laan, M.~J. (2016).
\newblock Statistical inference for the mean outcome under a possibly non-unique optimal treatment strategy.
\newblock {\em Annals of Statistics}, 44(2):713.

\bibitem[Ma et~al., 2022]{ma2022learning}
Ma, H., Zeng, D., and Liu, Y. (2022).
\newblock Learning individualized treatment rules with many treatments: A supervised clustering approach using adaptive fusion.
\newblock {\em Advances in Neural Information Processing Systems}, 35:15956--15969.

\bibitem[Ma et~al., 2023]{ma2023learning}
Ma, H., Zeng, D., and Liu, Y. (2023).
\newblock Learning optimal group-structured individualized treatment rules with many treatments.
\newblock {\em Journal of Machine Learning Research}, 24(102):1--48.

\bibitem[Marling and Bunescu, 2020]{marling2020ohiot1dm}
Marling, C. and Bunescu, R. (2020).
\newblock The ohiot1dm dataset for blood glucose level prediction: Update 2020.
\newblock In {\em CEUR workshop proceedings}, volume 2675, page~71. NIH Public Access.

\bibitem[Moodie et~al., 2012]{moodie2012q}
Moodie, E.~E., Chakraborty, B., and Kramer, M.~S. (2012).
\newblock Q-learning for estimating optimal dynamic treatment rules from observational data.
\newblock {\em Canadian Journal of Statistics}, 40(4):629--645.

\bibitem[Murphy, 2003]{murphy2003optimal}
Murphy, S.~A. (2003).
\newblock Optimal dynamic treatment regimes.
\newblock {\em Journal of the Royal Statistical Society: Series B (Statistical Methodology)}, 65(2):331--355.

\bibitem[Murphy, 2005]{murphy2005generalization}
Murphy, S.~A. (2005).
\newblock A generalization error for q-learning.
\newblock {\em Journal of Machine Learning Research}, 6.

\bibitem[Nie et~al., 2021]{nie2021learning}
Nie, X., Brunskill, E., and Wager, S. (2021).
\newblock Learning when-to-treat policies.
\newblock {\em Journal of the American Statistical Association}, 116(533):392--409.

\bibitem[Ning and Liu, 2017]{ning2017general}
Ning, Y. and Liu, H. (2017).
\newblock A general theory of hypothesis tests and confidence regions for sparse high dimensional models.
\newblock {\em Annals of Statistics}, 45(1):158--195.

\bibitem[Precup, 2000]{precup2000temporal}
Precup, D. (2000).
\newblock {\em Temporal abstraction in reinforcement learning}.
\newblock PhD thesis, University of Massachusetts Amherst.

\bibitem[Qi et~al., 2020]{qi2020multi}
Qi, Z., Liu, D., Fu, H., and Liu, Y. (2020).
\newblock Multi-armed angle-based direct learning for estimating optimal individualized treatment rules with various outcomes.
\newblock {\em Journal of the American Statistical Association}, 115(530):678--691.

\bibitem[Shi et~al., 2018]{shi2018high}
Shi, C., Fan, A., Song, R., and Lu, W. (2018).
\newblock High-dimensional a-learning for optimal dynamic treatment regimes.
\newblock {\em Annals of Statistics}, 46(3):925.

\bibitem[Shi et~al., 2020a]{shi2020breaking}
Shi, C., Lu, W., and Song, R. (2020a).
\newblock Breaking the curse of nonregularity with subagging: inference of the mean outcome under optimal treatment regimes.
\newblock {\em Journal of Machine Learning Research}, 21.

\bibitem[Shi et~al., 2021]{shi2021deeply}
Shi, C., Wan, R., Chernozhukov, V., and Song, R. (2021).
\newblock Deeply-debiased off-policy interval estimation.
\newblock In {\em International Conference on Machine Learning}, pages 9580--9591.

\bibitem[Shi et~al., 2020b]{shi2020does}
Shi, C., Wan, R., Song, R., Lu, W., and Leng, L. (2020b).
\newblock Does the markov decision process fit the data: Testing for the markov property in sequential decision making.
\newblock In {\em International Conference on Machine Learning}, pages 8807--8817. PMLR.

\bibitem[Shi et~al., 2022]{shi2022statistical}
Shi, C., Zhang, S., Lu, W., Song, R., et~al. (2022).
\newblock Statistical inference of the value function for reinforcement learning in infinite-horizon settings.
\newblock {\em Journal of the Royal Statistical Society Series B}, 84(3):765--793.

\bibitem[Sra et~al., 2011]{sra2011optimization}
Sra, S., Nowozin, S., and Wright, S.~J. (2011).
\newblock {\em Optimization for machine learning}.
\newblock MIT Press.

\bibitem[Sutton and Barto, 2018]{sutton2018reinforcement}
Sutton, R.~S. and Barto, A.~G. (2018).
\newblock {\em Reinforcement Learning: An Introduction}.
\newblock MIT Press.

\bibitem[Thomas and Brunskill, 2016]{thomas2016data}
Thomas, P. and Brunskill, E. (2016).
\newblock Data-efficient off-policy policy evaluation for reinforcement learning.
\newblock In {\em International Conference on Machine Learning}, pages 2139--2148. PMLR.

\bibitem[Thomas, 2015]{thomas2015safe}
Thomas, P.~S. (2015).
\newblock {\em Safe reinforcement learning}.
\newblock PhD thesis, University of Massachusetts Amherst.

\bibitem[van~der Vaart and Wellner, 1996]{van1996weak}
van~der Vaart, A. and Wellner, J.~A. (1996).
\newblock {\em Weak Convergence and Empirical Processes: With Applications to Statistics}.
\newblock Springer Science \& Business Media.

\bibitem[Zhang et~al., 2013]{zhang2013robust}
Zhang, B., Tsiatis, A.~A., Laber, E.~B., and Davidian, M. (2013).
\newblock Robust estimation of optimal dynamic treatment regimes for sequential treatment decisions.
\newblock {\em Biometrika}, 100(3):681--694.

\bibitem[Zhang, 2010]{zhang2010nearly}
Zhang, C.-H. (2010).
\newblock Nearly unbiased variable selection under minimax concave penalty.
\newblock {\em The Annals of Statistics}, 38(2):894--942.

\bibitem[Zhang and Zhang, 2014]{zhang2014confidence}
Zhang, C.-H. and Zhang, S.~S. (2014).
\newblock Confidence intervals for low dimensional parameters in high dimensional linear models.
\newblock {\em Journal of the Royal Statistical Society: Series B (Statistical Methodology)}, 76(1):217--242.

\bibitem[Zhao et~al., 2011]{zhao2011reinforcement}
Zhao, Y., Zeng, D., Socinski, M.~A., and Kosorok, M.~R. (2011).
\newblock Reinforcement learning strategies for clinical trials in nonsmall cell lung cancer.
\newblock {\em Biometrics}, 67(4):1422--1433.

\bibitem[Zhao et~al., 2015]{zhao2015new}
Zhao, Y.-Q., Zeng, D., Laber, E.~B., and Kosorok, M.~R. (2015).
\newblock New statistical learning methods for estimating optimal dynamic treatment regimes.
\newblock {\em Journal of the American Statistical Association}, 110(510):583--598.

\bibitem[Zhu et~al., 2019]{zhu2019proper}
Zhu, W., Zeng, D., and Song, R. (2019).
\newblock Proper inference for value function in high-dimensional q-learning for dynamic treatment regimes.
\newblock {\em Journal of the American Statistical Association}, 114(527):1404--1417.

\end{thebibliography}

\end{document}